\documentclass[10pt,journal,compsoc]{IEEEtran}
\ifCLASSOPTIONcompsoc
  \usepackage[nocompress]{cite}
\else
  \usepackage{cite}
\fi

\usepackage[utf8]{inputenc} 
\usepackage[T1]{fontenc}    
\usepackage{hyperref}       
\usepackage{url}            
\usepackage{booktabs}       
\usepackage{amsfonts}       
\usepackage{nicefrac}       
\usepackage{microtype}      
\usepackage{xcolor}         
\usepackage{bm}
\usepackage{natbib}
\setcitestyle{numbers,square}
\usepackage{subfigure}

\usepackage{threeparttable}
\usepackage{amsmath}
\usepackage{amsthm}
\usepackage{amssymb}
\usepackage{amsfonts}
\usepackage{amsopn}
\usepackage{verbatim}
\usepackage[noend]{algpseudocode}
\usepackage{algorithmicx, algorithm}
\usepackage{color}
\usepackage{enumitem}
\usepackage{stfloats}
\usepackage{float}
\usepackage{multirow}
\usepackage{siunitx}
\usepackage{graphicx}
\usepackage{mathrsfs}
\usepackage{wrapfig}
\usepackage{thmtools,thm-restate}

\newtheorem{theorem}{Theorem}
\newtheorem{example}{Example}
\newtheorem{corollary}{Corollary}[theorem]

\newtheorem{lemma}{Lemma}
\newtheorem{definition}{Definition}

\newtheorem{remark}{Remark}

\newcommand{\U}{\mathbf{U}}
\newcommand{\E}{\mathbf{E}}

\newcommand{\x}{\mathbf{x}}

\newcommand{\RR}{\mathbb{R}}
\newcommand{\Hk}{\widehat{H}_k}

\def\x{\bm{x}}

\def\a{\bm{a}}
\def\b{\bm{b}}
\newtheorem*{mytheorem}{Theorem 8}
\newtheorem*{myremark}{Remark 1}

\begin{document}
\title{Hilbert Curve Projection Distance\\
for Distribution Comparison}

\author{Tao~Li,   
        Cheng~Meng,    
        Hongteng~Xu,~\IEEEmembership{Member,~IEEE},
        Jun Yu
\IEEEcompsocitemizethanks{\IEEEcompsocthanksitem Tao Li is with the Institute of Statistics and Big Data, Renmin University of
China.\protect\\
E-mail: 2019000153@ruc.edu.cn

\IEEEcompsocthanksitem Cheng Meng is with the Center for Applied Statistics, Institute of Statistics and Big Data, Renmin University of
China.\protect\\
E-mail: chengmeng@ruc.edu.cn

\IEEEcompsocthanksitem Hongteng Xu is the corresponding author. He is with the Gaoling School of Artificial Intelligence,
Renmin University of China and Beijing Key Laboratory of Big Data
Management and Analysis Methods.\protect\\
E-mail: hongtengxu@ruc.edu.cn

\IEEEcompsocthanksitem Jun Yu is with the School of Mathematics and Statistics, Beijing Institute of Technology.\protect\\
E-mail: yujunbeta@bit.edu.cn

}
}


\IEEEtitleabstractindextext{%
\begin{abstract}
Distribution comparison plays a central role in many machine learning tasks like data classification and generative modeling. 
In this study, we propose a novel metric, called \textit{Hilbert curve projection (HCP) distance}, to measure the distance between two probability distributions with low complexity.
In particular, we first project two high-dimensional probability distributions using Hilbert curve to obtain a coupling between them, and then calculate the transport distance between these two distributions in the original space, according to the coupling.
We show that HCP distance is a proper metric and is well-defined for probability measures with bounded supports. 
Furthermore, we demonstrate that the modified empirical HCP distance with the $L_p$ cost in the $d$-dimensional space converges to its population counterpart at a rate of no more than $O(n^{-1/2\max\{d,p\}})$. 
To suppress the curse-of-dimensionality, we also develop two variants of the HCP distance using (learnable) subspace projections. 
Experiments on both synthetic and real-world data show that our HCP distance works as an effective surrogate of the Wasserstein distance with low complexity and overcomes the drawbacks of the sliced Wasserstein distance.
The code of this work is at \href{https://github.com/sherlockLitao/HCP}{https://github.com/sherlockLitao/HCP}.
\end{abstract}

}

\maketitle

\IEEEdisplaynontitleabstractindextext

\IEEEpeerreviewmaketitle

\IEEEraisesectionheading{\section{Introduction}\label{sec:introduction}}
\IEEEPARstart{M}{easuring} the distance between two probability distributions is significant for many machine learning tasks, e.g., data classification~\cite{kusner2015word,huang2016supervised,rakotomamonjy2018distance}, generative modeling~\cite{goodfellow2014generative,kingma2013auto}, among others. 
Among the commonly-used distance measures for probability distributions, classic $f$-divergence based metrics, e.g., the Kullback-Leibler (KL) divergence and the total variation (TV) distance, do not work well when the probability distributions have disjoint supports~\cite{arjovsky2017wasserstein}, while the kernel-based methods like the maximum mean discrepancy (MMD)~\cite{gretton2012kernel} require sophisticated kernel selection.
Recently, the Wasserstein distance~\cite{villani2009optimal} has attracted wide attention in the machine learning community because of its advantages on overcoming these limitations, and it has shown great potential in many challenging learning problems~\cite{tolstikhin2018wasserstein,arjovsky2017wasserstein}.

\begin{figure*}[t]
    \centering
    \subfigure[]{
    \includegraphics[height=3.8cm]{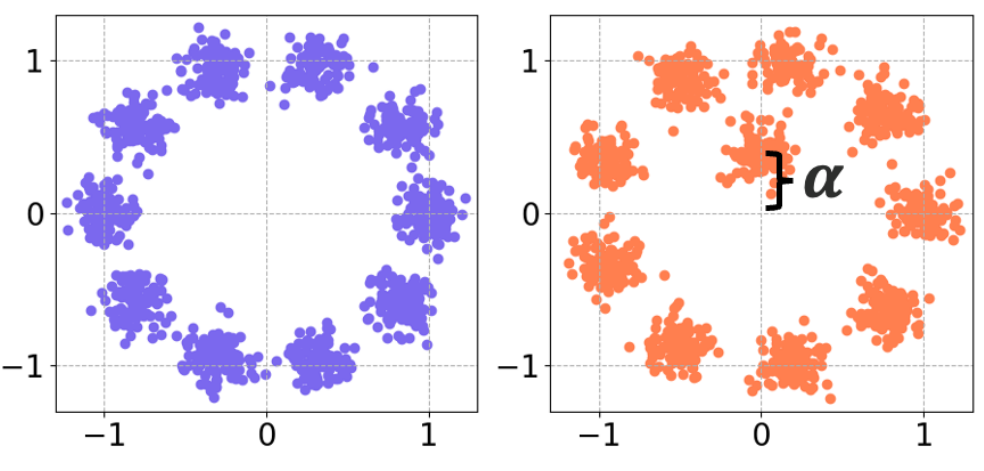}\label{fig:fig1a}
    }
    \subfigure[]{
    \includegraphics[height=3.8cm]{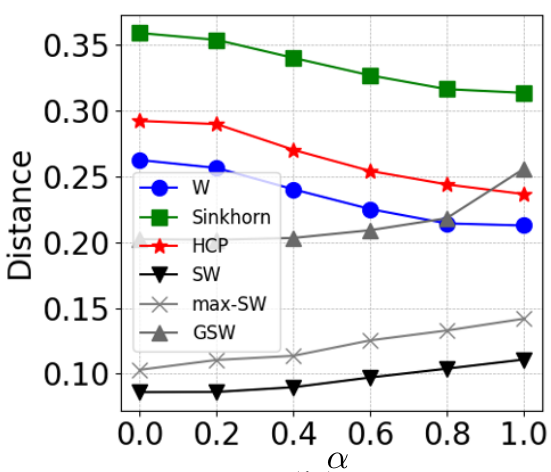}\label{fig:fig1b}
    }
    \subfigure[]{
    \includegraphics[height=3.8cm]{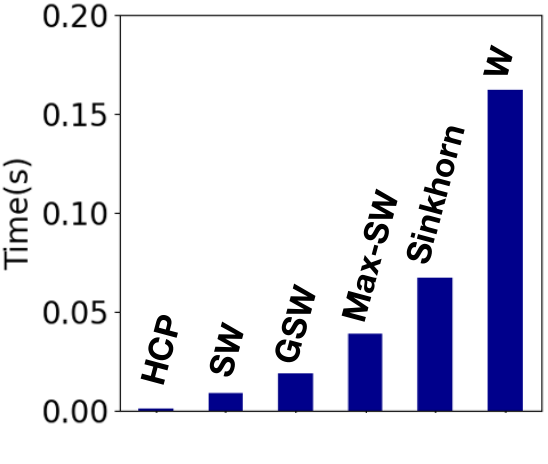}\label{fig:fig1c}
    }
    \caption{(a) The samples of source and target distributions.
    (b) Illustrations of various distances with the increase of $\alpha$.
    (c) Comparisons for various distances on their runtime.
    The proposed HCP distance provides an effective and efficient surrogate of the Wasserstein distance, which performs similarly and has low computational complexity.}
    \label{fig:fig1}
\end{figure*}

Given the samples of the two distributions, the computation of Wasserstein distance corresponds to solving either differential equations~\cite{brenier1997homogenized,benamou2002monge} or linear programming problems~\cite{rubner1997earth,pele2009fast}.
To alleviate the computational burden, the Sinkhorn distance~\cite{cuturi2013sinkhorn} imposes an entropic regularizer on the Wasserstein distance and leverages the Sinkhorn-scaling algorithm accordingly. 
The work in~\cite{arjovsky2017wasserstein} considers the Kantorovich duality of Wasserstein distance and converts the problem to a ``max-min'' game. 
Besides these two approximation methods, more surrogates of the Wasserstein distance have been proposed in recent years, e.g., the sliced Wasserstein (SW) distance~\cite{bonneel2015sliced}, the generalized sliced Wasserstein (GSW) distance~\cite{kolouri2019generalized}, the tree-structured Wasserstein (TSW) distance~\cite{le2019tree}, and so on. 
Despite the computational efficiency, these surrogates may fail to provide effective approximations for the Wasserstein distance. 
Take the two Gaussian mixture distributions in Fig.~\ref{fig:fig1a} as an example. 
We keep the source distribution (in purple) unchanged while shifting the central Gaussian component of the target distribution (in orange) vertically with an offset $\alpha\in [0, 1]$. 
For the various distances defined between the two distributions, Fig.~\ref{fig:fig1b} shows their changes with respect to $\alpha$. 
Existing methods often lead to coarse approximations of the Wasserstein distance, whose tendencies w.r.t. $\alpha$ can even be opposite to the Wasserstein distance.
This phenomenon indicates that replacing the Wasserstein distance with these surrogates may lead to sub-optimal, even undesired, results in some learning tasks. 

In this study, we propose a novel metric for distribution comparison, called Hilbert curve projection (HCP) distance.
In principle, our HCP distance first projects two probability distributions along the Hilbert curve~\cite{bader2012space} of the sample space and then calculates the coupling based on the projected distributions.
Such a Hilbert curve projection works better than linear projections on preserving the structure of the data distribution since the Hilbert curve enjoys the locality-preserving property, i.e., the locality between data points in the high-dimensional space being preserved in the projected one-dimensional space~\cite{abel1990comparative,moon2001analysis}.
Our HCP distance provides a new surrogate of the Wasserstein distance, with both efficiency and effectiveness --- it performs similarly as the Wasserstein distance does and spends less time than other methods, as shown in Fig.~\ref{fig:fig1b} and~\ref{fig:fig1c}, respectively. 

We provide in-depth analysis of the HCP distance, demonstrating that it is a well-defined metric for probability measures with bounded supports.
Given $n$ samples in $d$-dimensional space, the computational complexity for calculating empirical HCP distance is approximately linear to $n$.
In addition, the modified empirical HCP distance with the $L_p$ cost converges to its population counterpart at a rate of no more than $O(n^{-1/2\max\{d,p\}})$.
Furthermore, to mitigate the curse-of-dimensionality, we develop two variants of the HCP distance using (learnable) subspace projections.
We test the HCP distance and its variants on various machine learning tasks, including data classification and generative modeling, and compare them with state-of-the-art methods. 
Empirical results support the superior performance of the proposed metrics in both synthetic and real-data settings.

\section{Related Work and Preliminaries} \subsection{Wasserstein distance and sliced Wasserstein distance}
Let $\mathscr{P}_p(\RR^d)$ be the set of Borel probability measures in $\mathbb{R}^d$ with finite $p$-th moment.
Consider two probability measures $\mu,\nu \in \mathscr{P}_p(\RR^d)$ with corresponding probability density functions $f_{\mu}, f_{\nu}$.
The $p$-Wasserstein distance~\cite{villani2009optimal} between $\mu$ and $\nu$ is defined as 
\begin{eqnarray}\label{WP}
\begin{aligned}
\mathrm{W}_p(\mu,\nu)&=\Bigl(\inf_{\gamma \in \Gamma(\mu,\nu)}\int_{\RR^d\times\RR^d}\|x-y\|_p^p\mathrm{d}\gamma(x,y) \Bigr)^{1/p}\\
&\xrightarrow{\text{1D}~\mu,\nu}\Bigl(\int_0^1\|F_{\mu}^{-1}(z)- F_{\nu}^{-1}(z)\|_p^p\mathrm{d}z \Bigr)^{1/p},
\end{aligned}
\end{eqnarray}
where $\|\cdot\|_p$ is the $L_p$ norm and $\Gamma(\mu,\nu)$ is the set of all couplings (or called transportation plans):
$\Gamma(\mu,\nu)=\{\gamma \in \mathscr{P}_p(\RR^d\times\RR^d) \quad s.t. ~\forall \text{ Borel set } A,B \subset \RR^d,
\gamma(A \times \RR^d)=\mu(A), \gamma(\RR^d \times B)=\nu(B) \}$.

Though it is difficult to calculate Wasserstein distance in general, according to~\eqref{WP}, for one-dimensional probability measures $\mu$ and $\nu$, the Wasserstein distance has a closed-form, where $F_\mu(x)=\mu((-\infty,x])=\int_{-\infty}^xf_\mu(x)\mathrm{d}x$ is the cumulative distribution function (CDF) for $f_\mu$, and similarly, $F_\nu$ is the CDF for $f_\nu$. 
This fact motivates the design of the sliced Wasserstein (SW) distance~\cite{bonneel2015sliced} (and its variants~\cite{deshpande2019max,kolouri2018sliced}), which projects $d$-dimensional probability measures to 1D space and computes the 1D Wasserstein distance accordingly.
Let $\mathbb{S}_{d, q}=\left\{\E \in \mathbb{R}^{d \times q}: \E^{\top} \E=\mathbf{I}_{q}\right\}$ ($q<d$) be the set of orthogonal matrices and $P_\E(\x)=\E^{\top} \x$ be the linear transformation for $\x \in \mathbb{R}^{d}$.
Denote ${P_\E}_{\#} \mu$ as the pushforward of $\mu$ by $P_{\E}$, which corresponds to the distribution of the projected samples.
For all $\mu, \nu \in \mathscr{P}_{p}(\RR^d)$, the $p$-sliced Wasserstein distance between them is given by
\begin{equation}\label{SW}
   \mathrm{SW}_{p}(\mu, \nu)=\Bigl(\int_{\E\in\mathbb{S}_{d, 1}} \mathrm{W}_{p}^{p}\left({P_\E}_{\#} \mu, {P_\E}_{\#} \nu\right) \mathrm{d} \sigma(\E)\Bigr)^{1 / p},
\end{equation}
where $\sigma$ is the uniform distribution on $\mathbb{S}_{d, 1}$. 
However, as aforementioned, the SW distance often fails to approximate the Wasserstein distance because its linear projections break the structure of the original distributions. 
Additionally, the random projections introduce unnecessary randomness when computing the distance.

\subsection{Other optimal transport distances}
Based on the Wasserstein distance and the SW distance mentioned above, many optimal transport-based distances have been proposed in recent years, which can be roughly categorized into two classes.
The first class considers approximating the Wasserstein distance by alternative optimization methods.
Typically, the Sinkhorn distance in~\cite{cuturi2013sinkhorn} imposes an entropic regularizer on the Wasserstein distance.
Following this framework, many variants have been proposed to accelerate the computation~\cite{altschuler2017near,altschuler2019massively,dvurechensky2017adaptive,thibault2021overrelaxed,li2023importance,liao2022fast}. 
Besides the Sinkhorn-based algorithm, other methods, such as primal-dual method~\cite{guo2020fast,dvurechensky2018computational}, stochastic gradient descent~\cite{genevay2016stochastic}, proximal point method~\cite{xie2020fast}, Bregman alternating direction method of multipliers (Bregman ADMM)~\cite{wang2014bregman,ye2017fast,xu2022representing}, and so on, have drawn great attention.
However, the computational cost of these methods is at least $O(n^2)$, which may not be applicable to large-scale data.

The second class follows the strategy of the SW distance, finding surrogates of the Wasserstein distance by various projection methods.
To improve the efficiency of the SW distance, Max-sliced Wasserstein (Max-SW)~\cite{deshpande2019max}, distributional sliced Wasserstein~\cite{nguyen2020distributional} and orthogonal sliced Wasserstein~\cite{rowland2019orthogonal} have been proposed. 
Recently, the projection robust and integral projection robust Wasserstein distance consider projecting on the subspaces of higher dimensions~\cite{paty2019subspace,lin2020projection,lin2021projection}.
Beyond using linear projections, generalized sliced Wasserstein (GSW)~\cite{kolouri2019generalized}, convolutional sliced Wasserstein~\cite{nguyen2022revisiting}, and amortized sliced Wasserstein~\cite{nguyen2022amortized} have been proposed, which capture the complicated structure of data distributions by nonlinear projections. 
In addition to above methods, the tree sliced Wasserstein (TSW)~\cite{le2019tree} generates random tree metrics for data points and then computes Wasserstein distance on tree metrics.
TSW computes distance on given tree metrics and thus, it is more suitable for classification task compared with generative model. 
Note that, these projection-based distances may fail to provide effective surrogates for the Wasserstein distance, as shown in Fig.~\ref{fig:fig1}, and searching for effective projections will bring additional computational cost.

\subsection{Applications of optimal transport distances}
Optimal transport distances have recently drawn great attention in various machine-learning tasks.
Wasserstein distance and its variants serve as the loss functions for generative modeling, such as Wasserstein generative adversarial networks (WGANs)~\cite{arjovsky2017wasserstein,genevay2018learning,deshpande2019max,wu2019sliced} and Wasserstein autoencoders (WAEs)~\cite{tolstikhin2018wasserstein,kolouri2018sliced,xu2020learning}.
In classification tasks, optimal transport distances measure the discrepancy between set-level data~\cite{frogner2015learning}, leading to discriminative models for various data, such as texts~\cite{kusner2015word,huang2016supervised}, point clouds~\cite{rakotomamonjy2018distance}, and graphs~\cite{togninalli2019wasserstein}. 
Besides generative modeling and classification, optimal transport distances are also applied to other problems, such as data clustering~\cite{ye2017fast,li2023efficient}, dimension reduction~\cite{meng2020sufficient,flamary2018wasserstein}, and domain adaptation~\cite{courty2017joint}.
These optimal transport-based methods have shown the potential for various practical applications, e.g., graph matching and partitioning~\cite{xu2019gromov,xu2019scalable}, color transfer~\cite{rabin2014adaptive,meng2019large}, document analysis~\cite{yurochkin2019hierarchical}, and so on.

\begin{figure*}[t]
    \centering
    \subfigure[$\{\Hk\}_{k=1}^{3}$]{
    \includegraphics[height=4.5cm]{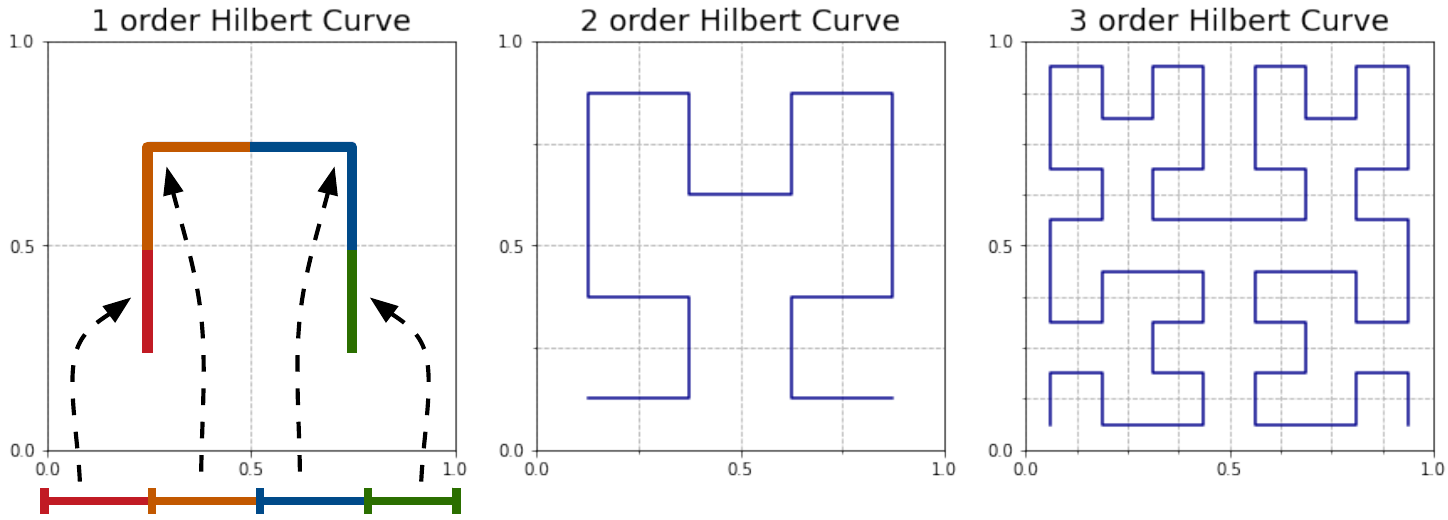}\label{fig:hci-a}
    }
    \hspace{0.3cm}
    \subfigure[HCP v.s. Linear proj.]{
    \includegraphics[height=4.5cm]{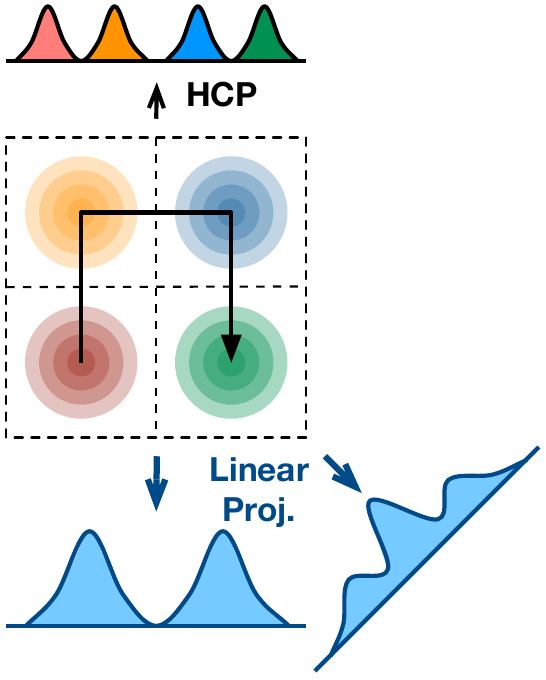}\label{fig:hci-b}
    }
    \caption{(a) The $k$-order Hilbert curve, with $k=1,2,3$, in 2D space. (b) The comparison between Hilbert curve projection (HCP) and linear projections.}
    \label{fig:hci}
\end{figure*}

\section{Proposed Method}
\subsection{Hilbert curve and its locality-preserving property}
Our work is based on the well-known Hilbert curve~\cite{bader2012space}. 
Mathematically, for a $d$-dimensional ($d\geq 2$) unit hyper-cube, i.e., $[0,1]^d$, the $k$-order Hilbert space-filling curve, denoted as $\Hk$, partitions $[0,1]$ and $[0,1]^d$ into $(2^k)^d$ intervals and blocks, respectively, and constructs a bijection between them. 
Taking the $\{\Hk\}_{k=1}^{3}$ in 2D space as examples, Fig.~\ref{fig:hci-a} illustrates how the intervals in $[0,1]$ are constructed and mapped to the blocks in $[0,1]^2$. 
The Hilbert curve is defined as the limit of a sequence of $k$-order Hilbert space-filling curves, i.e., $H(x) = \lim_{k\to\infty}\Hk(x)$ with $x\in [0, 1]$.
It provides a well-defined surjection $H:[0,1]\rightarrow[0,1]^d$ and is able to cover the entire hyper-cube~\cite{bader2012space}.
Note that, although the Hilbert curve $H$ is not a bijection, most of the data points in $[0,1]^d$ are still invertible --- it is known that the set $\mathcal{A}$, which includes the points in $[0, 1]^d$ such that these points have more than one pre-image in $[0, 1]$, has measure zero~\cite{he2016extensible}.
Actually, for any point in $\mathcal{A}$, there are finite pre-images in $[0,1]$.
Hence, there is a bijection between $[0,1]^d$ and $\big\{\min\{H^{-1}(x)\} : x \in [0,1]^d\big\}$.
We denote $\mathcal{N}=[0,1]\backslash\big\{\min\{H^{-1}(x)\} : x \in [0,1]^d\big\}$ as the negligible set.

We are interested in the Hilbert curve because it enjoys the so-called \textbf{locality-preserving property}~\cite{zumbusch2012parallel,he2016extensible}: For any $x, y \in[0,1]$, one has $$\|H(x)-H(y)\|_2 \leq 2 \sqrt{d+3}|x-y|^{1 / d}.$$ 
Such an inequality indicates the advantage of the Hilbert curve over linear projections.
In particular, if two points are far from each other in a high-dimensional space, their pre-images with respect to the Hilbert curve will also be far from each other. 
Fig.~\ref{fig:hci-b} further illustrates this property through a toy example. 
Specifically, for a 2D distribution with four modals, while linear projections tend to wrongly merge some modals, the projection along the Hilbert curve can distinguish the modals successfully.
This property motivates us to propose the Hilbert curve projection distance shown below.

\subsection{Hilbert curve projection distance}
\textbf{Hilbert curve for probability measure:} 
In this study, we focus on probability measures with bounded supports. 
This condition has been widely used in the optimal transport literature to simplify the theoretical analysis ~\citep{fatras2020learning,piccoli2014generalized,weed2019sharp,xi2022distributional}.
Let $\mathscr{P}_\infty(\RR^d)$ be the set of Borel probability measures in $\mathbb{R}^d$ with bounded supports.
Denote the support of a probability measure $\mu\in\mathscr{P}_\infty(\RR^d)$ as $\Omega_{\mu}$. 
Let $\widetilde{\Omega}_\mu=\prod_{i=1}^d[a_i,b_i]$ be the smallest hyper-rectangle covering $\Omega_\mu$. 
For each $\mu$, we can define a Hilbert curve $H_\mu :[0,1]\rightarrow\widetilde\Omega_\mu$ as $H_\mu(t)=(b-a)\odot H(t)+a$ where 
$\odot$ is the Hadamard product and $a,b$ are vectors with $i$-th dimension being $a_i,b_i$ respectively. 
\footnote{
In the case when $a_i=b_i$, one can utilize the following two strategies without affecting the theoretical properties.
The first strategy is to let $b_i=a_i+1$.
This may cause redundant computational costs in this dimension. 
The second strategy is removing this dimension, performing Hilbert curve in the $\mathbb{R}^{d-1}$ and complementing this dimension for the final Hilbert curve.}

Denote $\mathcal{K}=\{\frac{m_1}{2^{m_2}}:m_1,m_2\in\mathbb{N}, m_1\leq 2^{m_2}\}$ as a dense set in $[0,1]$.   
According to~\cite{zumbusch2012parallel,he2016extensible}, $H_\mu([0,t])$ is a Borel measurable set for any $t\in\mathcal{K}$ and 
$H_\mu([0,t])$ is a Lebesgue measurable set for any $t\in[0,1]$.
This motivates us to define a cumulative distribution function along the Hilbert curve (denoted as $g_\mu:[0,1]\rightarrow [0,1]$) and the corresponding inverse cumulative distribution function ($g_{\mu}^{-1}$), respectively:
\begin{eqnarray}\label{eq:cdf}
\begin{aligned}
    &g_\mu(t)=\sideset{}{_{s\in\mathcal{K},~s\geq t}}\inf \mu\Bigl(H_\mu([0,s])\Bigr),\\
    &g_\mu^{-1}(t)=\sideset{}{_{s\in[0,1],~g_\mu(s)>t}}\inf s. 
\end{aligned}
\end{eqnarray}
Accordingly, the formal definition of our Hilbert curve projection distance is as follows.
\begin{definition}[Hilbert Curve Projection Distance]\label{def:HCP}
Let $\mathscr{P}_\infty(\RR^d)$ be the set of Borel probability measures in $\mathbb{R}^d$ with bounded supports.
Denote the supports of two probability measures $\mu,\nu\in\mathscr{P}_\infty(\RR^d)$ as $\Omega_\mu$ and $\Omega_\nu$, respectively.
Denote $\mathcal{K}=\{\frac{m_1}{2^{m_2}}:m_1,m_2\in\mathbb{N}, m_1\leq 2^{m_2}\}$ as a dense set in $[0,1]$.
Let $H_\mu :[0,1]\rightarrow\widetilde\Omega_\mu$, where $\widetilde\Omega_\mu$ is the smallest hyper-rectangle that covers $\Omega_\mu$, $g_\mu(t)=\sideset{}{_{s\in\mathcal{K},~s\geq t}}\inf \mu\Bigl(H_\mu([0,s])\Bigr)$, and $g_\mu^{-1}(t)=\sideset{}{_{s\in[0,1],~g_\mu(s)>t}}\inf s$ (with $H_\nu$, $g_\nu$ and $g_\nu^{-1}$ defined in the same way). 
For $p\in\mathbb{Z}_{+}$, the $p$-order Hilbert curve projection distance is defined as
\begin{eqnarray}\label{eq:def-hcp}
\begin{aligned}
    \text{HCP}_p(\mu, \nu) = \Bigl(\int_0^1{\|H_\mu(g_\mu^{-1}(t))-H_\nu(g_\nu^{-1}(t))\|}_p^p\mathrm{d}t\Bigr)^{\frac{1}{p}}.
\end{aligned}
\end{eqnarray}
\end{definition}

\begin{remark}
    The assumption for bounded support is commonly used in optimal transport literature \citep{fatras2020learning,piccoli2014generalized,weed2019sharp,xi2022distributional} and is essential for technical proof.
    For unbounded cases, one possible remedy is to use a bounded measurable bijective mapping $f$, such as element-wise $\tan^{-1}(\cdot)$, to transform the original measures $\mu$ and $\nu$.
    We then could get the transport plan between $f_\# \mu$ and $f_\# \nu$ based on the Hilbert curve projections where $f_\# \mu$ is the pushforward of $\mu$ by $f$, and compute the distance in the original unbounded space.
\end{remark}

According to the definition, the principle of our HCP distance is projecting high-dimensional distributions along their Hilbert curves to obtain an efficient and effective coupling between them, and then calculating the corresponding HCP distance between two distributions in the original space according to the coupling.
The following theoretical results show that our HCP distance is a proper metric, and it is an upper bound of the $p$-Wasserstein distance. 

\begin{theorem}\label{thm:HCP}
$\text{HCP}_p$ is a well-defined metric in $\mathscr{P}_\infty(\RR^d)$, and $\text{W}_p(\mu,\nu)\leq \text{HCP}_p(\mu,\nu)$,  $\forall \mu, \nu\in \mathscr{P}_\infty(\RR^d)$.
\end{theorem}
Given two random variables, i.e., $Z_1\sim \mu$ and $Z_2\sim \nu$, we denote $\text{HCP}(\mu,\nu)$ as $\text{HCP}(Z_1,Z_2)$.
Clearly, HCP distance has the following properties which are also valid for Wasserstein distance~\cite{panaretos2019statistical}.
\begin{enumerate}
    \item For any $z\in\mathbb{R}^d$, $\text{HCP}_p(Z_1+z,Z_1)=\|z\|_p$.
    \item For any $a\in\mathbb{R}$, $\text{HCP}_p(aZ_1,aZ_2)=|a|\text{HCP}_p(Z_1,Z_2)$.
    \item For any $z\in\mathbb{R}^d$, $\text{HCP}_p(Z_1+z,Z_2+z)=\text{HCP}_p(Z_1,Z_2)$.
    \item For any $z\in\mathbb{R}^d$, $\text{HCP}_2^2(Z_1+z,Z_2)=\text{HCP}_2^2(Z_1,Z_2)+\|z+\mathbb{E}Z_1-\mathbb{E}Z_2\|_2^2-\|\mathbb{E}Z_1-\mathbb{E}Z_2\|_2^2$.
\end{enumerate}
Here, ``$Z_1+z$'' means impose a translation $z$ on the random variable $Z_1$, and ``$aZ_1$'' means scaling the random variable $Z_1$.

\subsection{Topological properties of the HCP distance}
As shown in Theorem~\ref{thm:HCP}, HCP distance induces a stronger topology compared to Wasserstein distance because $\text{W}_p(\mu,\nu)\leq \text{HCP}_p(\mu,\nu)$.
This means that the sequence of probability measures, i.e., $\{\mu_n\}$, always converges in Wasserstein distance when $n\rightarrow\infty$ if it converges in HCP distance, i.e., $\text{HCP}(\mu_n,\mu)\to 0 \Rightarrow \text{W}(\mu_n,\mu)\to 0$.

Additionally, we compare our HCP distance with the total variation (TV) distance on their induced topology and propose the following Theorem:
\begin{theorem}\label{thm:topo}
    Let $\widetilde\Omega_\mu$ be the smallest hyper-rectangle that covers the support of the probability measure $\mu\in\mathscr{P}_\infty(\RR^d)$.
    When $\{\mu_n\}$ converges to $\mu$ in the total variation distance and $\widetilde{\Omega}_{\mu_n}=\widetilde{\Omega}_\mu$ for all $n$'s, we have
    $\text{TV}(\mu_n,\mu)\to 0 \Rightarrow \text{HCP}(\mu_n,\mu)\to 0$.
\end{theorem}

Note that, our HCP distance is not equivalent to the Wasserstein distance or the TV distance because
\begin{eqnarray*}
\begin{aligned}
    &\text{W}(\mu_n,\mu)\to 0 \nRightarrow \text{HCP}(\mu_n,\mu)\to 0,\\
    &\text{HCP}(\mu_n,\mu)\to 0 \nRightarrow \text{TV}(\mu_n,\mu)\to 0.
\end{aligned}
\end{eqnarray*}
The following two examples verify the above claims, respectively. 
\begin{example}\label{exam1}
Consider two probability distribution 
$\mu_\theta=\frac{1}{4}(\delta_{(0,0)} + \delta_{(1,1)} + \delta_{(\frac{1}{2}-\theta,\frac{1}{4})} + \delta_{(\frac{1}{2}-\theta,\frac{3}{4})})
$ and $\nu_\theta=\frac{1}{4}(\delta_{(0,0)} + \delta_{(1,1)} + \delta_{(\frac{1}{2}+\theta,\frac{1}{4})} + \delta_{(\frac{1}{2}+\theta,\frac{3}{4})})
$ where $\delta$ is the Dirac measure.
Then, when $0<\theta<0.5$, we have $\text{W}_2(\mu_\theta,\nu_\theta)=|\sqrt{2}\theta|$.
However, when $\theta \neq 0$, $\text{HCP}_2(\mu_\theta,\nu_\theta)=\sqrt{2\theta^2+1/8}$. 
\end{example}
\begin{example}\label{exam2}
Let $Z\sim \text{Unif}[0, 1]$ be samples of the uniform distribution on the unit interval.
Let $\mu_0$ be the probability distribution of $(0, Z)\in \mathbb{R}^2$.
Let $\mu_\theta$ be the family of probability distributions
parametrized with $\theta$ corresponding to $(\theta, Z) \in \mathbb{R}^2$. 
Then $\text{HCP}_p(\mu_0,\mu_\theta)=\text{W}_p(\mu_0,\mu_\theta)=|\theta|$.
However, when $\theta \neq 0$, $\text{TV}(\mu_0,\mu_\theta)=1$.
\end{example}

In summary, we can find that HCP metricizes a topology stronger than the weak topology induced by the Wasserstein distance. 
Additionally, as shown in Example~\ref{exam2} (which is also used in~\cite{arjovsky2017wasserstein}), our HCP distance can perform as well as the Wasserstein distance does when comparing the probability measures with disjoint supports.

\subsection{Numerical implementation}
Let $\Delta^{n}$ be the $n$-Simplex.
Given the samples of two probability measures, i.e., $X=\{x_i\}_{i=1}^{n}\sim \mu$ and $Y=\{y_j\}_{j=1}^{m}\sim \nu$, whose empirical distributions are $\a\in\Delta^{n-1}$ and $\b\in\Delta^{m-1}$, respectively, we use a $k$-order Hilbert curve to calculate the empirical HCP distance between the two sample sets.
Let $\widetilde\Omega_X$ and $\widetilde\Omega_Y$ be the smallest hyper-rectangles that cover these two sample sets, respectively.
We define two $k$-order Hilbert curves, i.e., $\Hk^X:[0,1]\rightarrow \widetilde\Omega_X$ and $\Hk^Y:[0,1]\rightarrow \widetilde\Omega_Y$.
Here, $\Hk^X$ partitions both $[0,1]$ and $\widetilde\Omega_X$ into $2^{kd}$ blocks, denoted by $\{c'_{j,X}\}_{j=1}^{2^{dk}}$ and $\{c_{j,X}\}_{j=1}^{2^{dk}}$, respectively, and construct a bijection between these blocks.
For any data point $x\in\widetilde\Omega_X$, we assign $x$ to its corresponding block $c_{j,X}$ in $\widetilde\Omega_X$, $j\in\{1,\ldots,2^{kd}\}$, then map $x$ to the center of the block $c_{j,X}'={(\Hk^X)}^{-1}(c_{j,X})$.
Therefore, all the samples belonging to the same block are mapped to the same point in $[0,1]$.
Based on $\Hk^Y$, we map $\{y_j\}_{j=1}^{m}$ to $[0,1]$ in the same way.
The mapped points along with their probability densities are then used to calculate the optimal coupling matrix $\mathbf{P}\in\RR^{n\times m}$ using the closed-form formulation of the 1D optimal transport problem. 
In particular, we first sort the mapped points, then calculate $\mathbf{P}$ using the North-West corner rule with $O(n+m)$ operations~\cite{peyre2019computational}. 
Note that there are at most $m+n$ nonzero elements in $\mathbf{P}$.
Let $\mathcal{S}:= \{(i,j)|P_{ij}\neq 0\}$ be the index set.
Finally, the empirical HCP distance can then be calculated by
$(\sum_{(i,j)\in\mathcal{S}}\|x_i-y_j\|_p^p P_{ij})^{1/p}$.

\begin{figure*}[t]
    \centering
    \includegraphics[width=0.95\textwidth]{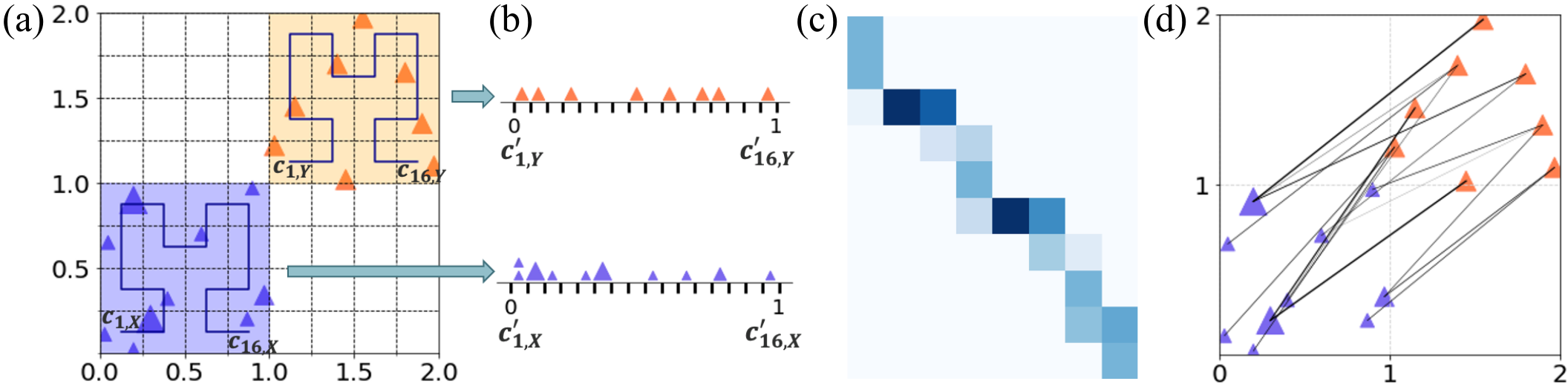}
    \caption{An illustration of Algorithm~\ref{alg:hcp} when $d=k=2$. 
    (a) The source (purple) and target (orange) data points, with corresponding hyper-rectangles and $k$-order Hilbert curves.
    (b) The projected points along the Hilbert curves.
    (c) The coupling matrix calculated by the projected points. 
    (d) The HCP distance calculates the distance between the original samples based on the coupling matrix.}
    \label{fig:alg1}
\end{figure*}

\begin{algorithm}[t]
\caption{Computation of HCP distance}\label{alg:hcp}
\begin{algorithmic}[1]
\State {\bf Input:} $(\{x_i\}_{i=1}^{n},\a)$, $(\{y_j\}_{j=1}^{m},\b)$, $k$
\State Map  $\{x_i\}_{i=1}^{n}$ to $\{x_i'\}_{i=1}^{n}$, $\{y_j\}_{j=1}^{m}$ to $\{y_j'\}_{j=1}^{m}$, through ${(\Hk^X)}^{-1}$ and ${(\Hk^Y)}^{-1}$\hfill \textcolor{red}{$O((n+m)dk)$}
\State Calculate the optimal transport plan $\mathbf{P}$ between $(\{x_i'\}_{i=1}^{n},\a)$ and $(\{y_j'\}_{j=1}^{m},\b)$ using sorting and the North-West corner rule. 
Let $\mathcal{S}:= \{(i,j)|P_{ij}\neq 0\}$ \hfill \textcolor{red}{$O(n\log(n)+m\log(m))$}
\State {\bf Output:} $\mathbf{P}$, $\text{HCP}_p=(\sum_{(i,j)\in\mathcal{S}}\|x_i-y_j\|_p^p P_{ij})^{1/p}$
\end{algorithmic}
\end{algorithm}

The above pipeline is illustrated in Fig.~\ref{fig:alg1} and summarized in Algorithm~\ref{alg:hcp}, respectively.
As suggested by~\cite{bader2012space}, we select $k$ that of the order $O(\log(n))$ in practice.
Empirical results in the following experimental section show that the performance of Algorithm~\ref{alg:hcp} is not sensitive to $k$.

Essentially, the empirical HCP distance is to compute the distance between two Hilbert rank-based sorted samples.\footnote{The Hilbert rank is defined as follows: We say $x_1$ ranks in front of $x_2$, that is to say, $\min\{H^{-1}(x_1)\}<\min\{H^{-1}(x_2)\}$.}
Note that, there are two main routines for Hilbert sort.
The first gets Hilbert indices by projecting points in high dimension to the Hilbert curve and then sorts these indices based on the Hilbert rank~\cite{butz1969convergence,butz1971alternative,zumbusch2012parallel,bader2012space,skilling2004programming}.
The second idea is recursively sorting points without using Hilbert indices, e.g., the work in~\cite{tanaka2001study,imamura2016fast,bader2012space} and the C++ library CGAL \citep{fabri2009cgal}.
Though we take the first routine here, codes based on these two algorithms are both provided.

\textbf{Computational cost.}
The complexity of computing the $k$-order Hilbert index for $n$ points in $d$-dimensional space is $O(ndk)$~\cite{hamilton2008compact,tanaka2001study,imamura2016fast}.
As shown in Algorithm~\ref{alg:hcp}, solving the optimal transport problem in Step~3 requires $O(n\log n+m\log m)$ time.
When $m=O(n)$ and $k=O(\log(n))$, the overall computational complexity of HCP distance is at the order of $O(n\log(n)d)$.

\textbf{Comparison with existing methods.}
The proposed HCP distance enjoys several critical advantages over the Wasserstein and SW distance. 
\begin{itemize}
    \item Firstly, HCP can provide a decent transport plan between the input probability measures as a byproduct while SW could not.
    The key reason is that Hilbert curve is invertible almost everywhere.
    Linear projections in SW and nonlinear projections in GSW do not satisfy this property.
    Such a coupling matrix is essential for effective generative modeling, as will be seen in Section~\ref{sec:exp}.
    \item Secondly, we compute the distance in the original space rather than in the projected one-dimensional space.
    Hilbert curve only plays a role in achieving a transport plan.
    We don't apply any transformation on data points when computing HCP distance.
    However, SW involves transforming data points by linear projections and then computing Wasserstein distance using these transformed data points.
    Fig.~\ref{fig:fig1} provides an intuitive example to show the difference between these two strategies.
    The reason why SW and its variants lead to an opposite trend compared with the Wasserstein distance is that SW computes Wasserstein distance using linear transformed data points, and such linear transformation may break the structure of the original distributions.
    We refer to the Experiment Section for a more intuitive discussion.
    \item Last but not least, HCP computes faster than SW distance in practice.
    This is because calculating SW distance requires projection and sorting multiple times, while calculating HCP distance requires only once.
    Additionally, beyond the Hilbert curve-based discrepancy in~\cite{bernton2019approximate}, our HCP distance can deal with the samples with different sizes and weights with theoretical guarantees.
\end{itemize}
In summary, compared to Wasserstein distance, HCP has an approximately linear computational complexity, and thus is applicable to large-scale datasets. 
Compared to SW distances, HCP distance performs more similarly to the Wasserstein distance.

\subsection{Statistical convergence of empirical HCP distance}
Let $\{x_i\}_{i=1}^n\sim \mu$, whose empirical measure is defined by $\mu_n=\frac{1}{n}\sum_{i=1}^n\delta_{x_i}$.
Directly studying the statistical convergence of $\text{HCP}_p(\mu,\mu_n)$ is challenging because of the randomness of the bounded supports --- the smallest hyper-rectangle covering the support of the probability measure $\mu_n$, i.e., $\widetilde\Omega_{\mu_n}$ can be various w.r.t. sample size $n$, which leads to different Hilbert curves, and accordingly, we could not easily analyze the convergence rate without any other strict conditions on the support's boundary. 

To eliminate the influence of the randomness, we consider an indirect strategy, studying a modified empirical Hilbert curve projection distance instead. 
Specifically, following the definitions in~\eqref{eq:cdf}, we first define the cumulative distribution function and its inverse for the empirical measure $\mu_n$, whose Hilbert curve, however, is based on the original probability measure $\mu$:
\begin{eqnarray}\label{eq:cdf2}
\begin{aligned}
    &\hat{g}_{\mu_n}(t)=\sideset{}{_{s\in\mathcal{K},~s\geq t}}\inf\mu_n\Bigl(H_\mu([0,s])\Bigr),\\
    &\hat{g}_{\mu_n}^{-1}(t)=\sideset{}{_{s\in[0,1],~\hat{g}_{\mu_n}(s)>t}}\inf s.
\end{aligned}
\end{eqnarray}
Accordingly, we define the modified empirical Hilbert curve projection distance as:
\begin{eqnarray}\label{eq:mod-hcp1}
\begin{aligned}
    \overline{\text{HCP}}_p(\mu, \mu_n) = \Bigl(\int_0^1{\|H_\mu(g_\mu^{-1}(t))-H_{\mu}(\hat{g}_{\mu_n}^{-1}(t))\|}_p^p\mathrm{d}t\Bigr)^{\frac{1}{p}}.
\end{aligned}
\end{eqnarray}
The only difference between ${\text{HCP}}_p(\mu, \mu_n)$ and $\overline{\text{HCP}}_p(\mu, \mu_n)$ is that the latter replaces the $H_{\mu_n}$ defined on $\widetilde\Omega_{\mu_n}$ with the $H_\mu$ defined on $\widetilde\Omega_{\mu}$. 
Note that, such a modified HCP distance is hard to implement in practice because both $\widetilde\Omega_{\mu}$ and $H_\mu$ are unknown in general. 
However, compared to the original HCP distance, the modified HCP distance is much easier to analyze because the Hilbert curve $H_\mu$ it used is deterministic and irrelevant to the sample. 
We demonstrate that the modified empirical HCP distance converges to its population counterpart almost surely. 
The following theorem provides an upper bound for the convergence rate. 
\begin{theorem}\label{thm:conv1}
Let $\{x_i\}_{i=1}^n$ be an i.i.d. sample that is generated from the probability measure $\mu\in\mathscr{P}_\infty(\RR^d)$.
The empirical measure is defined by $\mu_n=\frac{1}{n}\sum_{i=1}^n\delta_{x_i}$.
Then, we have almost surely
\begin{eqnarray*}
\begin{aligned}
    \overline{\text{HCP}}_p(\mu, \mu_n)\to 0,~~\text{and}~~\mathbb{E}\overline{\text{HCP}}_p(\mu, \mu_n)\lesssim O(n^{-\frac{1}{2\max\{p,d\}}}).
\end{aligned}
\end{eqnarray*}
\end{theorem}

Directly from Theorem~\ref{thm:conv1}, we can conclude the following theoretical results.
\begin{corollary}\label{coro:conv1}
Assume that probability measures $\mu,\nu\in\mathscr{P}_\infty(\RR^d)$. 
Let $\{x_i\}_{i=1}^n$ and $\{y_i\}_{i=1}^n$ be two i.i.d. samples, which are generated from probability measures $\mu$ and $\nu$, respectively.
Let $\{x_{{(i)}^*}\}_{i=1}^n$ and $\{y_{{(i)}^*}\}_{i=1}^n$ be the sorted samples along the Hilbert curves $H_\mu$ and $H_\nu$, respectively.
Then, we have almost surely
\begin{eqnarray*}
\begin{aligned}
\overline{\text{HCP}}_p(\mu_n, \nu_n)= \Bigl(\frac{1}{n}\sum_{i=1}^n \|x_{{(i)}^*}-y_{{(i)}^*})\|_p^p\Bigr)^{\frac{1}{p}}\to \text{HCP}_p(\mu, \nu),
\end{aligned}
\end{eqnarray*}
where $\mu_n$ and $\nu_n$ are the empirical version of $\mu$ and $\nu$, respectively.
Furthermore, we have
\begin{eqnarray*}
\begin{aligned}
|\mathbb{E}\overline{\text{HCP}}_p(\mu_n, \nu_n)- \text{HCP}_p(\mu, \nu)|\lesssim O(n^{-\frac{1}{2\max\{p,d\}}}).
\end{aligned}
\end{eqnarray*}
\end{corollary}
Corollary~\ref{coro:conv1} tells us the modified empirical Hilbert curve distance is to compute the
distance between two Hilbert rank-based sorted samples.
Moreover, when the samples are with
different numbers, we have
\begin{corollary}\label{rmk:coro1}
Assume that probability measures $\mu,\nu\in\mathscr{P}_\infty(\RR^d)$. 
Let $\{x_i\}_{i=1}^n$ and $\{y_j\}_{j=1}^m$ be two i.i.d. samples, which are generated from probability measures $\mu$ and $\nu$, respectively.
Let $\{x_{{(i)}^*}\}_{i=1}^n$ and $\{y_{{(j)}^*}\}_{j=1}^m$ be the sorted samples along the Hilbert curves $H_\mu$ and $H_\nu$, respectively.
Then, we have
\begin{eqnarray*}
\begin{aligned}
\overline{\text{HCP}}_p(\mu_n, \nu_m)= \Bigl(\pi_{ij}\sum_{i=1}^n\sum_{j=1}^m \|x_{{(i)}^*}-y_{{(j)}^*})\|_p^p\Bigr)^{\frac{1}{p}},
\end{aligned}
\end{eqnarray*}
where $\mu_n,\nu_m$ are the empirical version of $\mu,\nu$, respectively and, $\pi_{ij}$ is the optimal transport plan between $\sum_{i=1}^n \delta_i/n$ and $\sum_{j=1}^m \delta_j/m$ with Euclidean distance cost.
Furthermore, we have
\begin{eqnarray*}
\begin{aligned}
|\mathbb{E}\overline{\text{HCP}}_p(\mu_n, \nu_m)- \text{HCP}_p(\mu, \nu)|
\lesssim O(\min\{n,m\}^{-\frac{1}{2\max\{p,d\}}}).
\end{aligned}
\end{eqnarray*}
\end{corollary}

Additionally, from Theorem~\ref{thm:conv1}, we know that convergence rate of modified empirical HCP distance has an upper bound $O(n^{-1/2p}+n^{-1/2d})$, which is slightly slower than the convergence rate of Wasserstein distance (i.e., $O(n^{-1/2p}+n^{-1/d})$ provided by~\cite{panaretos2019statistical}). 
In particular, given a probability measure $\mu$ and its empirical version $\mu_n$, we have
\begin{eqnarray*}
\begin{aligned}
    W_p(\mu,\mu_n) \leq \overline{\text{HCP}}_p(\mu, \mu_n).
\end{aligned}
\end{eqnarray*}
Furthermore, the following corollary indicates that under some mild conditions, the modified HCP distance can have the same convergence rate as Wasserstein distance does. 
\begin{corollary}
Assume that probability measure $\mu\in\mathcal{P}_\infty(\RR^d)$.
If there exist two Borel measurable sets $A,B\subset\RR^d$ such that $\mu(A)>0, \mu(B)>0, \mu(A\cup B)=1$ and $dist(A, B)=\inf_{x\in A,y\in B}\|x-y\|_2>0$, then when $p\geq d$, we have
\begin{eqnarray*}
\begin{aligned}
    \mathbb{E}\overline{\text{HCP}}_p(\mu, \mu_n)
=O(n^{-\frac{1}{2p}}),~~\text{and}~~ \mathbb{E}W_p(\mu_n,\mu)=O(n^{-\frac{1}{2p}}),
\end{aligned}
\end{eqnarray*}
where $\mu_n$ is the empirical version of $\mu$.
\label{rmk:conv1}
\end{corollary}

The above theoretical results of the modified HCP distance provide us with important insights into the convergence of our HCP distance --- with the increase of the sample size $n$, the difference between $\widetilde\Omega_{\mu_n}$ and $\widetilde\Omega_{\mu}$ may not be too large in probability. 
Accordingly, the convergence of our HCP distance should be similar to that of the modified HCP distance in probability as well. 
Under some special cases, we can easily analyze the convergence of our HCP distance.
For example, for nondegenerate discrete measures with finite supports, $\mathbb{E}{\text{HCP}}_p(\mu, \mu_n)$
is of the order $O(n^{-1/2p})$, independently of the dimension, which is the same as the Wasserstein distance~\cite{panaretos2019statistical}. 
\begin{corollary}\label{thm:conv2}
Assume that probability measure $\mu$ is a non-degenerate discrete probability measure with $K$ supports $\{s_i\}_{i=1}^K$, that is, $\mu=\sum_{i=1}^K p_i\delta_{s_i}$ and $\bm{p}=\{p_i\}_{i=1}^{K}\in\Delta^{K-1}$.
Then, we have
\begin{eqnarray*}
\begin{aligned}
    \mathbb{E}{\text{HCP}}_p(\mu, \mu_n)
=O(n^{-\frac{1}{2p}}),
\end{aligned}
\end{eqnarray*}
where $\mu_n$ is the empirical version of $\mu$.
\end{corollary}

\subsection{Other space-filling curves} 
The proposed distance can be implemented based on other space-filling curves as well.
For example, the Peano and Sierpinski space-filling curves also satisfy the H{\"o}lder inequality with exponent $1/d$.
The Z-order space-filling curve, which is differentiable almost everywhere, also satisfies the H{\"o}lder inequality but with exponent $1/(d\log_23)$~\cite{zumbusch2012parallel,he2016extensible}. 
However, compared to the Hilbert curve, the Peano curve and Sierpinski curve are difficult to implement through algorithms. 
The convergence rate of the distance based on the Z-order curve is $O(n^{-\frac{1}{2\max\{d\log_23,p\}}})$, which is slower than that based on the Hilbert curve. 
In sum, we mainly focus on the Hilbert curve in this study.

\section{Variants of the Hilbert Curve Projection Distance}
The theoretical results in the previous section indicate that analogous to the Wasserstein distance, our HCP distance may suffer from the curse-of-dimensionality as well. 
Motivated by the projection-robust Wasserstein distance \citep{lin2020projection,lin2021projection}, we propose two variants of the HCP distance to alleviate this limitation.

\subsection{Integral projection robust Hilbert curve projection distance} 
We first propose the integral projection robust Hilbert curve projection (IPRHCP) distance that combines the idea of HCP distance and random projections.
\begin{definition}
\label{def:IPRHCP}
Suppose that probability measures $\mu,\nu\in\mathscr{P}_\infty(\RR^d)$. 
The $p$-order $q$-dimensional integral projection robust Hilbert curve projection distance is defined as 
\begin{eqnarray}\label{eqa:IPRHCP}
\begin{aligned}
    &\text{IPRHCP}_{p,q}(\mu, \nu)\\
    &=\Bigl(\int_{\E\in\mathbb{S}_{d, q}} \text{HCP}_{p}^{p}\left({P_\E}_{\#} \mu, {P_\E}_{\#} \nu\right) \mathrm{d} \sigma(\E)\Bigr)^{\frac{1}{p}},
\end{aligned}
\end{eqnarray}
where $\sigma$ is the uniform distribution on ${\mathbb{S}_{d, q}}$.
\end{definition}
Next, we demonstrate that IPRHCP distance is a valid distance metric and reveal the relations between IPRHCP distance and other metrics, including the $p$-order SW distance~\cite{bonneel2015sliced} and the $p$-order $q$-dimensional integral projection robust Wasserstein distance~\cite{lin2021projection}, denoted as $\text{IPRW}_{p,q}$. 
\begin{theorem}
$\text{IPRHCP}_{p,q}$ is a well-defined metric in $\mathscr{P}_\infty(\RR^d)$, and we have $\text{IPRW}_{p,q}\left(\mu, \nu\right) \leq \text{IPRHCP}_{p,q}\left(\mu, \nu\right)$, $\forall \mu,\nu \in \mathscr{P}_\infty(\RR^d)$.
\label{thm:IPRHCP}
\end{theorem}
In practice, the expectation in~\eqref{eqa:IPRHCP} can be approximated using a Monte Carlo scheme: 
We first randomly and uniformly draw several matrices from the set of orthogonal matrices $\mathbb{S}_{d,q}$.
We then project the distributions to subspace $\E$ and compute the HCP distance between the projected samples.
Finally, we replace the expectation on the right hand side of~\eqref{eqa:IPRHCP} with a finite-sample average.

\begin{theorem}\label{thm:iprhcp_ineq}
Given two probability measures $\mu,\nu \in \mathscr{P}_\infty(\RR^d)$, we have
$\text{SW}_{p}^{p}\left(\mu, \nu\right) \leq \alpha_{q,p} \text{IPRHCP}_{p,q}^{p}\left(\mu, \nu\right)$, where $\alpha_{q, p}= \int_{\mathbb{S}_{q,1}}\|\theta\|_{p}^{p} d \theta/q \leq 1$.
As a special case, when $p=2$, one has $\alpha_{q, 2}=1/q$ and $\text{SW}_{2}\left(\mu, \nu\right) \leq \text{IPRHCP}_{2,q}\left(\mu, \nu\right)/\sqrt{q}$.
\end{theorem}
\begin{corollary}\label{rmk:iprhcp_ineq}
If we replace $\mathbb{S}_{d, q}$ in~\eqref{eqa:IPRHCP} with matrix set $\left\{\E \in \mathbb{R}^{d \times q}: \E^{\top} \E=\mathbf{J}_{q}\right\}$ where $\mathbf{J}_{q}$ is a $q\times q$ all-ones matrix, we have $\text{IPRHCP}_{p,q}(\mu,\nu)=q^{1/p}\text{SW}_p(\mu,\nu)$, $\forall \mu,\nu \in \mathscr{P}_\infty(\RR^d)$.
\end{corollary}

IPRHCP shares a similar sense to SW.
As shown in Theorem~\ref{thm:iprhcp_ineq} and Corollary~\ref{rmk:iprhcp_ineq}, we provided some inequalities and equalities between IPRHCP and SW to illustrate their relationship.
We provide the following theorem to show IPRHCP overcomes curse-of-dimensionality.

\begin{theorem}\label{thm:iprhcp}
Suppose that probability measures $\mu,\nu\in\mathscr{P}_\infty(\RR^d)$.
Let $\{x_i\}_{i=1}^n$ and $\{y_i\}_{i=1}^n$ be two i.i.d. samples, which are generated from probability measures $\mu$ and $\nu$, respectively.
Let $\{x_{\E,{(i)}^*}\}_{i=1}^n$ and $\{y_{\E,{(i)}^*}\}_{i=1}^n$ be the sorted samples of $\{\E^Tx_i\}_{i=1}^n$ and $\{\E^Ty_i\}_{i=1}^n$ along the Hilbert curves $H_{{P_\E}_{\#} \mu}$ and $H_{{P_\E}_{\#} \nu}$, respectively. 
Based on the definition of $\overline{\text{HCP}}(\mu_n, \nu_n)$, we can define
\begin{eqnarray*}
\begin{aligned}
    &\overline{\text{IPRHCP}}_{p,q}(\mu_n, \nu_n)\\
    =&\Bigl(\int_{\E\in\mathbb{S}_{d, q}} \overline{\text{HCP}}_{p}^{p}\left({P_\E}_{\#} \mu_n, {P_\E}_{\#} \nu_n\right) \mathrm{d}\sigma(\E)\Bigr)^{\frac{1}{p}}\\
    =&\Bigl(\int_{\E\in\mathbb{S}_{d, q}}\frac{1}{n}\sum_{i=1}^n \|x_{\E,{(i)}^*}-y_{\E,{(i)}^*})\|_p^p\mathrm{d}\sigma(\E)\Bigr)^{\frac{1}{p}},
\end{aligned}
\end{eqnarray*}
where $\mu_n$ and $\nu_n$ are the empirical version of $\mu$ and $\nu$, respectively.
Then, we have
\begin{eqnarray*}
\begin{aligned}
|\mathbb{E}\overline{\text{IPRHCP}}_{p,q}(\mu_n, \nu_n)-\text{IPRHCP}_{p,q}(\mu, \nu)|\lesssim O(n^{-\frac{1}{2\max\{p,q\}}}).
\end{aligned}
\end{eqnarray*}
\end{theorem}

\subsection{Projection robust Hilbert curve projection distance}
The IPRHCP distance considers the integration of the HCP distances defined in all $q$-dimensional subspaces.
When assuming the two distributions differ only on one low-dimensional subspace, as the projection robust Wasserstein (PRW) distance~\cite{lin2020projection} does, we can avoid the integration and just consider the maximal possible HCP distance among all projections, which leads to the proposed projection robust Hilbert curve projection (PRHCP) distance.
\begin{definition}
\label{def:PRHCP}
Suppose that probability measures $\mu,\nu\in\mathscr{P}_\infty(\RR^d)$.
The $p$-order $q$-dimensional projection robust Hilbert curve projection distance is defined as 
\begin{equation}\label{PRHCP}
\text{PRHCP}_{p,q}(\mu, \nu)=\sideset{}{_{\E\in\mathbb{S}_{d, q}}}\sup \text{HCP}_{p}\left({P_\E}_{\#} \mu, {P_\E}_{\#} \nu\right).
\end{equation}
\end{definition}

The PRHCP distance is also a valid distance.
\begin{theorem}
    $\text{PRHCP}_{p,q}(\mu,\nu)$ is a well-defined metric in $\mathscr{P}_\infty(\RR^d)$, and we have $\text{PRW}_{p,q}\left(\mu, \nu\right) \leq \text{PRHCP}_{p,q}\left(\mu, \nu\right)$, $\forall\mu,\nu \in \mathscr{P}_\infty(\RR^d)$.
\label{theorem:PRHCP}
\end{theorem}

In practice, given the samples of the probability measures, i.e., the sample matrices $\mathbf{X}=[x_i^\top]\in\RR^{n\times d}$ and $\mathbf{Y}=[y_j^\top]\in\RR^{m\times d}$, we consider an EM-like optimization scheme to calculate the empirically PRHCP distance, i.e., we optimize the transport plan $\mathbf{P}$ and the $d \times q$ orthogonal matrix $\E$ alternately and iteratively.
Details for calculating PRHCP distance are summarized in Algorithm~\ref{alg:ALG2}.
This algorithm is similar to the one for calculating the subspace robust Wasserstein distance in~\cite{paty2019subspace}, except that the transport plan is calculated by the HCP distance.
As we observed in numerical experiments, Algorithm~\ref{alg:ALG2} performs well for high-dimensional cases and is robust to noise.
Theoretical justification for these observations is left for future work.

\begin{algorithm}[t]
        \caption{Computation of PRHCP distance}
        \label{alg:ALG2}
        \begin{algorithmic}[1]
        \State\textbf{Input:} $(\mathbf{X}=\{x_i\}_{i=1}^{n},\a)$, $(\mathbf{Y}=\{y_j\}_{j=1}^{m},\b)$, $k$, $q$
        \State Initialize $\U=\bm{\Omega}= \mathbf{I}_d$, $t=0$, $\tau=1$
        \State \textbf{While not converge}
        \begin{itemize}
        \item[a)] $\mathbf{P}\leftarrow\mbox{Algorithm~1}[(\{\U^\top x_i\}_{i=1}^{n},\a), (\{\U^\top y_j\}_{j=1}^{m},\b), k]$  \hfill \textcolor{red}{$O((n\log(n)+m\log(m))d)$}
        \item[b)] $\U \in\RR^{d\times q}\leftarrow$ top $q$ singular vectors of the matrix $(\mathbf{X}-\mbox{diag}(\a^{-1})\mathbf{P}\mathbf{Y})$ with weight $\a$ \hfill \textcolor{red}{$O((n+m)d^2)$}
        \item[c)] $\bm{\Omega} \leftarrow (1-\tau)\bm{\Omega}+\tau\U\U^\top$, and then $\U\leftarrow$ top $q$ eigenvectors of $\bm{\Omega}$ \hfill \textcolor{red}{$O(d^2q+d^3)$}
        \item[d)] $t\leftarrow t+1$, $\tau \leftarrow 2/(2+t)$
        \end{itemize}
        \State\textbf{Output:}  The coupling $\mathbf{P}$, and $\text{PRHCP}_{p,q}=(\sum_{(i,j)\in\{(i,j)|P_{ij}\neq 0\}} \|\U^\top x_i-\U^\top y_j\|_p^p P_{ij})^{1/p}$
        \end{algorithmic}
\end{algorithm}

\textbf{Computational cost.}
For brevity, we consider the case that $n=m>d>q$.
Step~3(a) requires $O(n\log(n)d)$ time, as discussed in the last section.
Recall that there are at most $(n+m)$ nonzero elements in $\mathbf{P}$, and thus Step~3(b) requires only $O(n+m)d^2$ time.
The cost for Step~3(c) involves $O(d^2q)$ for $\U\U^{\top}$ and $O(d^3)$ for solving the eigen-decomposition problem, respectively.
Thus, the overall complexity of Algorithm~\ref{alg:ALG2} is $O(n\log(n)dL+nd^2L)$, where $L$ is the number of iterations. 

\textbf{The proofs of above Theorems and their corollaries are given in Appendix.}

\begin{figure}[t]
    \centering
    \subfigure[CPU time of the $k$-order Hilbert curve]{\includegraphics[width=0.95\linewidth]{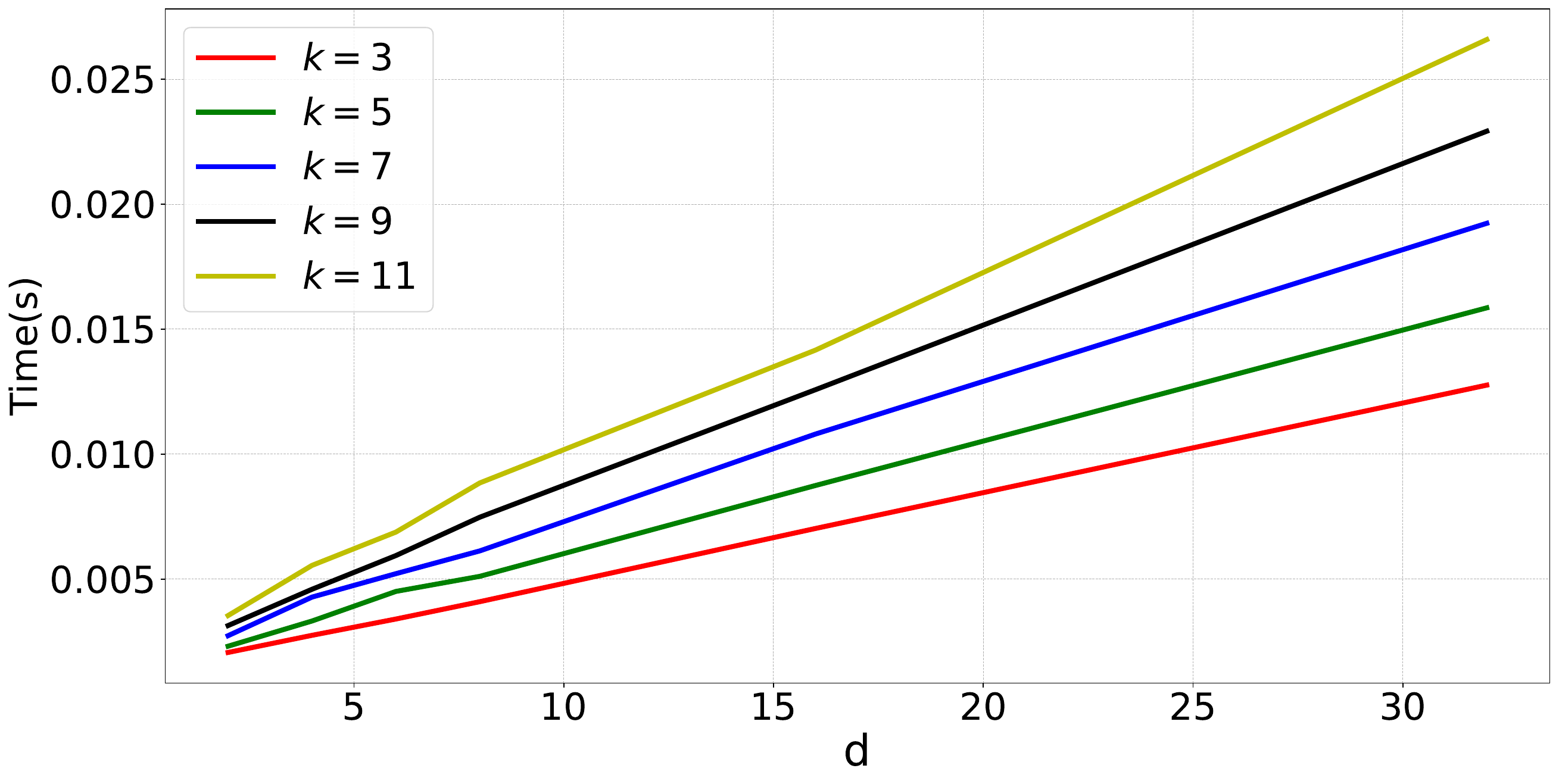}\label{fig:Hk100}
    }
    \subfigure[HCP distance and its CPU time ($d=2$)]{\includegraphics[width=0.95\linewidth]{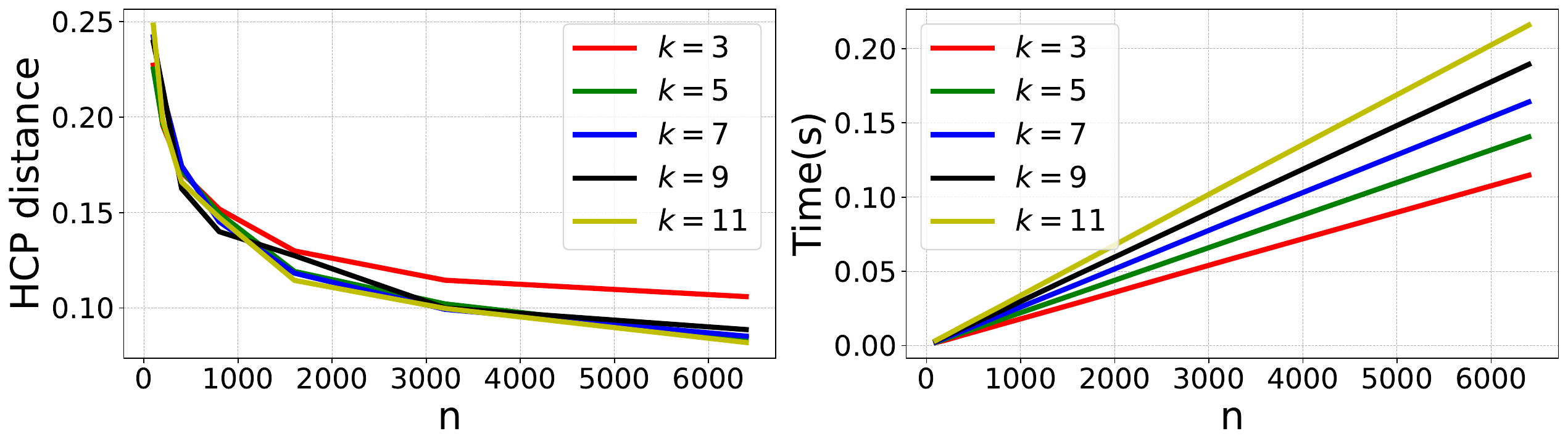}\label{fig:Hk2}
    }
    \subfigure[HCP distance and its CPU time ($d=10$)]{\includegraphics[width=0.95\linewidth]{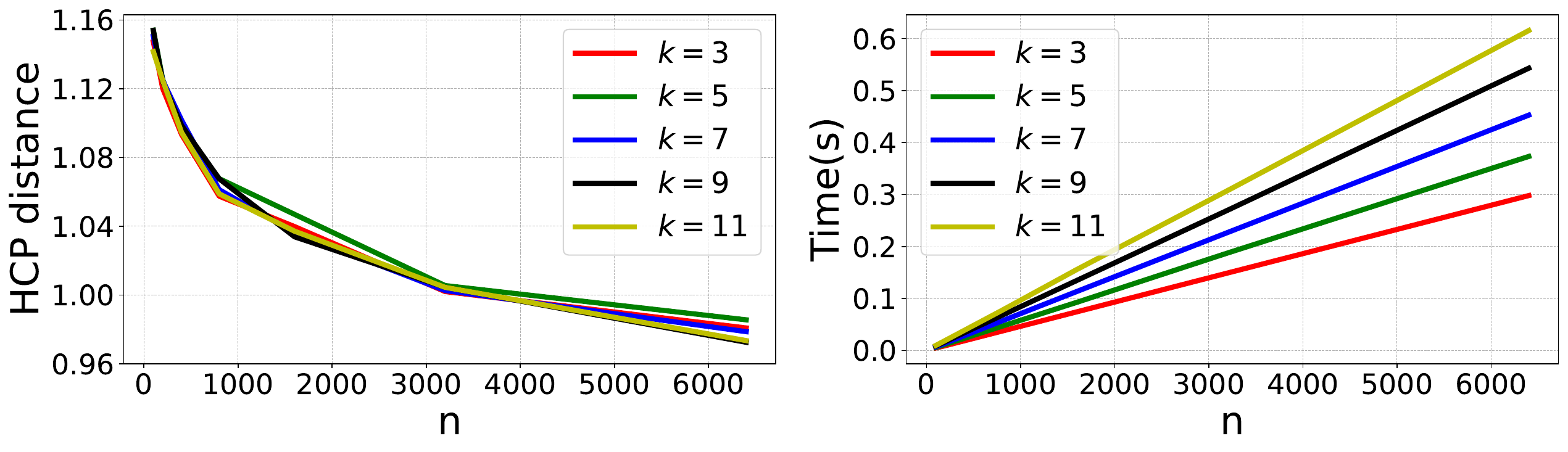}\label{fig:Hk10}
    }
    \caption{(a) CPU time for generating the $k$-order Hilbert curve versus $d$ when $n=100$.
    (b) Left: HCP distance versus $n$ when $d=2$. 
    Right: CPU time for generating the $k$-order Hilbert curve versus $n$ when $d=2$.
    (c) Left: HCP distance versus $n$ when $d=10$. 
    Right: CPU time for generating the $k$-order Hilbert curve versus $n$ when $d=10$.}
\end{figure}

\section{Experiments}\label{sec:exp}
To demonstrate the feasibility and efficiency of our HCP distance and its variants, we conducted extensive numerical experiments and compared them with the main-stream competitors, including maximum mean discrepancy (MMD), Wasserstein distance, Sinkhorn distance~\cite{cuturi2013sinkhorn}, SW distance~\cite{bonneel2015sliced}, max-SW distance~\cite{deshpande2019max}, GSW distance~\cite{kolouri2019generalized}, TSW distance~\cite{le2019tree}, and PRW distance~\cite{lin2020projection}.
For all the distances, we considered the Euclidean cost, i.e., $p=2$.
We use $k$-order Hilbert curves with $k=5\log(n)$.
We set the dimension for the intrinsic space as $q=2$ for PRW, IPRHCP, and PRHCP.
All experiments are implemented by an AMD 3600 CPU and an RTX 1080Ti GPU.
For each experiment, we replicate it 100 times and record the average performance.

\subsection{Analytic experiments on synthetic data}
\subsubsection{Robustness and efficiency analysis} 
The performance of our HCP distance is mainly determined by three factors: (1) the order of the Hilbert curve; (2) the dimension of sample; and (3) the number of samples. 
To demonstrate the robustness and efficiency of our HCP distance, we test it on synthetic data and analyze the influences of the above three factors.

Specifically, we generate two sample sets of size $n$ from the uniform distribution on the unit hypercube $[0,1]^d$ and we calculate the HCP distance between these two sample sets. 
When calculating the HCP distance, the $k$-order Hilbert curve is applied.
The results in Fig.~\ref{fig:Hk100} indicate the computational cost for generating the $k$-order Hilbert curve is linear to $d$. 
Figs.~\ref{fig:Hk2} and~\ref{fig:Hk10} show the average HCP distances and the average CPU time for generating the $k$-order Hilbert curve versus different $n$'s when $d=2$ or $10$, respectively.
From these two figures, we observe that the HCP distance is not sensitive to the choice of $k$, as long as $k$ is not too small (i.e., $k>3$). 
We also observe that the computational cost for generating the $k$-order Hilbert curve is linear to $n$.

\begin{figure}[t]
    \centering
    \includegraphics[width=0.95\linewidth]{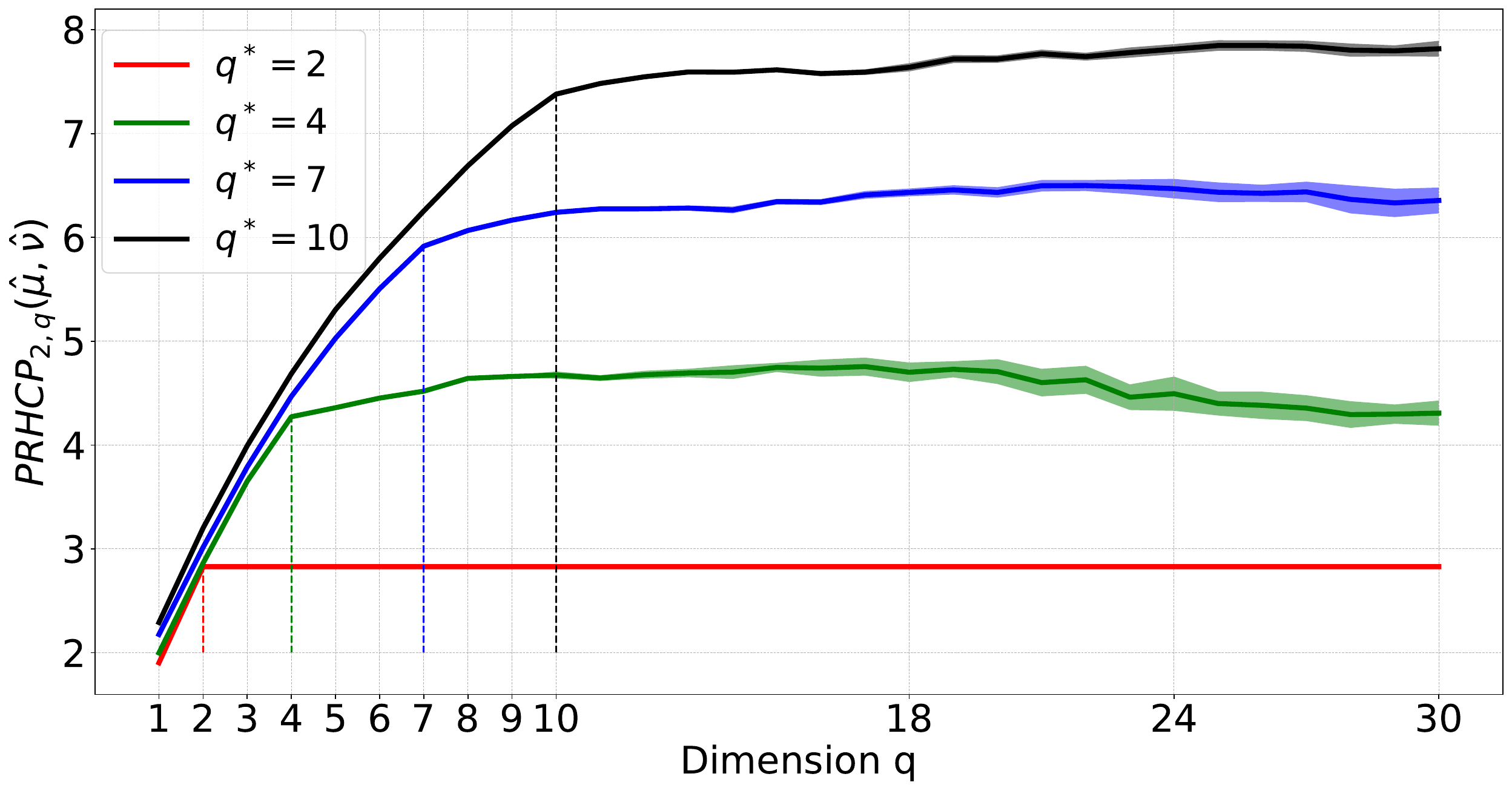}
    \caption{$\text{PRHCP}_{2,q}(\mu_n,\nu_n)$ versus the dimension $q$ for $q^*=2,4,7,10$.}
    \label{fig:prhcp}
\end{figure}

For the variant of our HCP distance, i.e., the PRHCP distance, one more factor should be considered --- the dimension of subspace. 
Ideally, this distance should be robust to the setting of $q$ as long as $q$ is equal to or larger than the dimension of the effective subspaces. 

To demonstrate their robustness to $q$, we follow the settings in~\cite{lin2020projection,paty2019subspace}, considering a uniform distribution $\mu=\mathcal{U}([-1,1])^{d}$ and its pushforward under a map $T$, i.e., $\nu=T_{\#} \mu$.
Here, the map $T(x)=x+2 \operatorname{sign}(x) \odot\left(\sum_{i=1}^{q^*} e_{i}\right)$, where sign is taken elementwise, $q^{*}=2,4,7,10$, and $\left(e_{1}, \ldots, e_{d}\right)$ is the canonical basis of $\mathbb{R}^{d}$. 
Obviously, the map $T$ splits the hypercube into four different hyper-rectangles, and the dimension of the effective subspace equals to $q^{*}$. 

Setting $d=50$ and $n=100$, we calculate the PRHCP distance under different $q$'s.
Fig.~\ref{fig:prhcp} shows the PRHCP distance increases rapidly when $q<q^*$ and tends to be stable and consistent when $q \geq q^{*}$.
Such an observation indicates PRHCP can dig out useful subspace information effectively.

\subsubsection{Comparisons in high-dimensional scenarios}

As shown in Fig.~\ref{fig:fig1}, our HCP distance provides an effective and efficient surrogate of Wasserstein distance for 2D data. 
Here, we further compare various metrics on approximating Wasserstein distance for high-dimensional data. 
In particular, let $\{x_{i}\}_{i=1}^n$ and $\{y_{i}\}_{i=1}^n$ be i.i.d. samples generated from two Gaussian distributions, i.e.,  $\mathcal{N}_{d}\left(\bm{0}_d, \bm{\Sigma}_{X}\right)$ and $\mathcal{N}_{d}\left(\bm{0}_d, \bm{\Sigma}_{Y}\right)$, respectively.
We consider three different settings as follows.
\begin{enumerate}[noitemsep,topsep=0pt]
    \item $\bm{\mu}_{X}=\bm{0}_d,\quad \boldsymbol{\mu}_{Y}=(\theta,\theta,0,\ldots,0)^\top,\quad
    \bm{\Sigma}_{X}=\bm{\Sigma}_{Y}=\bm{I}_d$.
    \item $\bm{\mu}_{X}=\text{diag}(3\bm{I}_2, \bm{I}_{d-2}),\bm{\mu}_{Y}=\text{diag}(\theta\bm{I}_2, \bm{I}_{d-2})$.
    \item $\bm{\mu}_{X}=\text{diag}(3\bm{I}_2, \bm{I}_{d-2}),\bm{\mu}_{Y}=\text{diag}(\theta\bm{I}_2+3\theta\bm{B}_2, \bm{I}_{d-2})$.
\end{enumerate}
where $\bm{I}_d$ and $\bm{B}_d$ are identity and backward identity matrices with size $(d\times d)$, respectively.
In each of the three settings, the distance between the two distributions is controlled by a hyperparameter $\theta$.

\begin{figure*}[t]
    \centering
    \subfigure[]{
    \includegraphics[height=3.8cm]{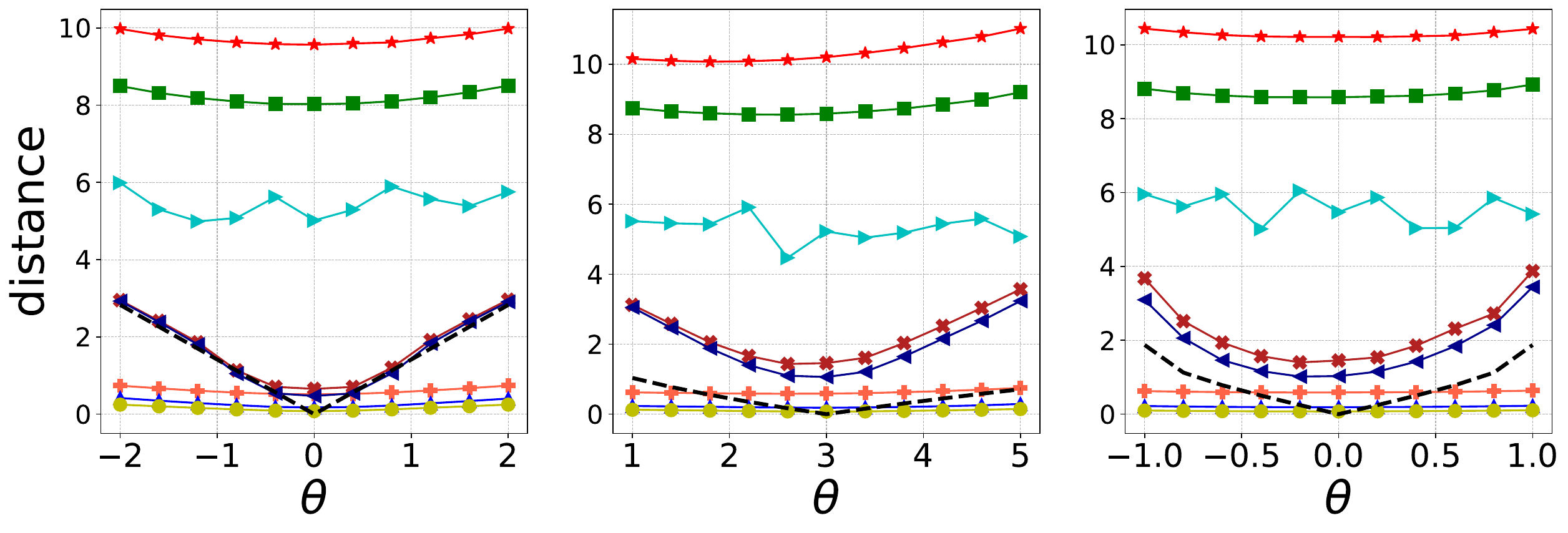}\label{fig:simu1}
    }
    \subfigure[]{
    \includegraphics[height=3.8cm]{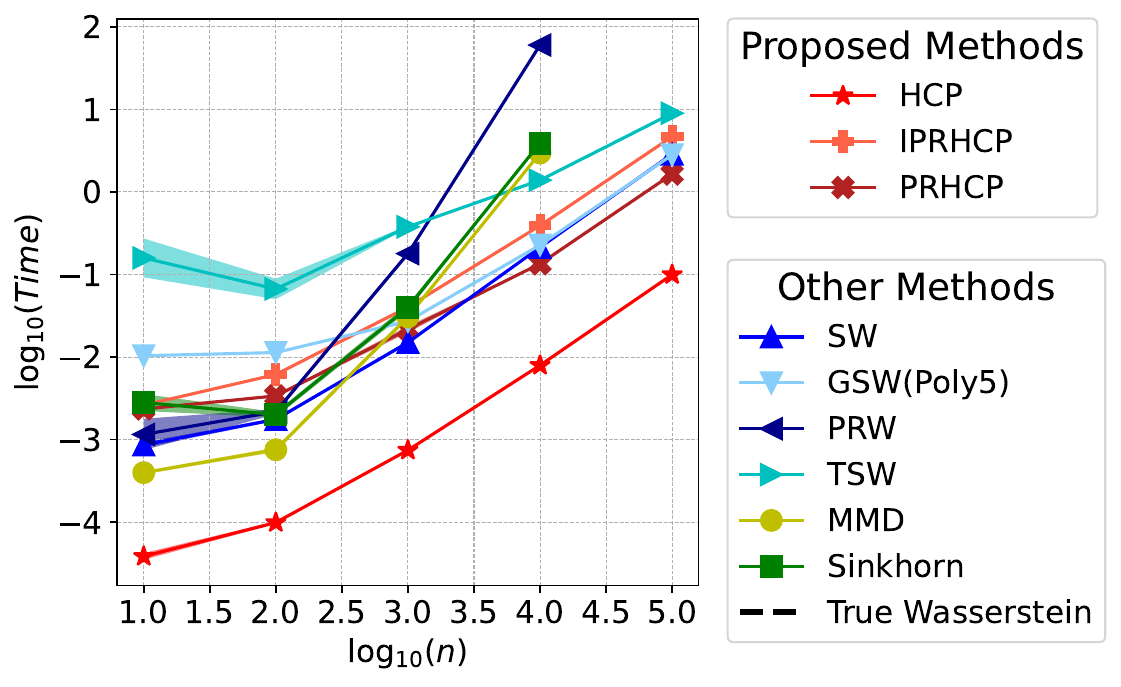}\label{fig:time}
    }
    \caption{The comparison for various metrics. (a) Distances versus different $\theta$. (b) CPU time versus different $n$.}
\end{figure*}

We set $n=200$, $d=50$.  
Given different $\theta$'s, we generate different samples and calculate the distance between the two sample sets under different metrics. 
Fig.~\ref{fig:simu1} shows the averaged distance in 100 trials. 
Taking the true Wasserstein distance between the two Gaussian densities as a benchmark, $\text{W}_2(\mathcal{N}_{d}\left(\bm{0}_d, \bm{\Sigma}_{X}\right),\mathcal{N}_{d}\left(\bm{0}_d, \bm{\Sigma}_{Y}\right))=\text{tr}(\bm{\Sigma}_{X}+\bm{\Sigma}_{Y}-2(\bm{\Sigma}_{X}^{\frac{1}{2}}\bm{\Sigma}_{Y}\bm{\Sigma}_{X}^{\frac{1}{2}})^{\frac{1}{2}})^{\frac{1}{2}}$, we observe that most of the metrics, including our HCP distance, suffer from the curse-of-dimensionality or lack of robustness to noise, i.e., their distances are not sensitive to the parameter $\theta$.
Among these metrics, the PRW distance and our PRHCP distance are the only two that provide reasonable distances --- they perform similarly as the true Wasserstein distance. 
In other words, although the HCP distance suffers from the curse-of-dimensionality, this problem can be mitigated by combining the HCP distance with the subspace projection strategy, leading to the PRHCP distance. 
Besides the comparison on the effectiveness, we also compare the CPU time for different metrics. 
Fig.~\ref{fig:time} shows the CPU time (in seconds) versus different $n$'s.
The time for our methods, including the HCP distance and its variants, is approximately linear to $n$. 
Compared to other metrics, our HCP requires significantly less time than all the competitors, and its two variants are at least comparable to other distances in runtime. 
Especially, our PRHCP distance works as well as the PRW distance does in high-dimensional scenarios, but its runtime is much less than the PRW distance's runtime, which demonstrates its superiority on both effectiveness and efficiency.

\begin{figure}[t]
    \centering
    \includegraphics[width=0.95\linewidth]{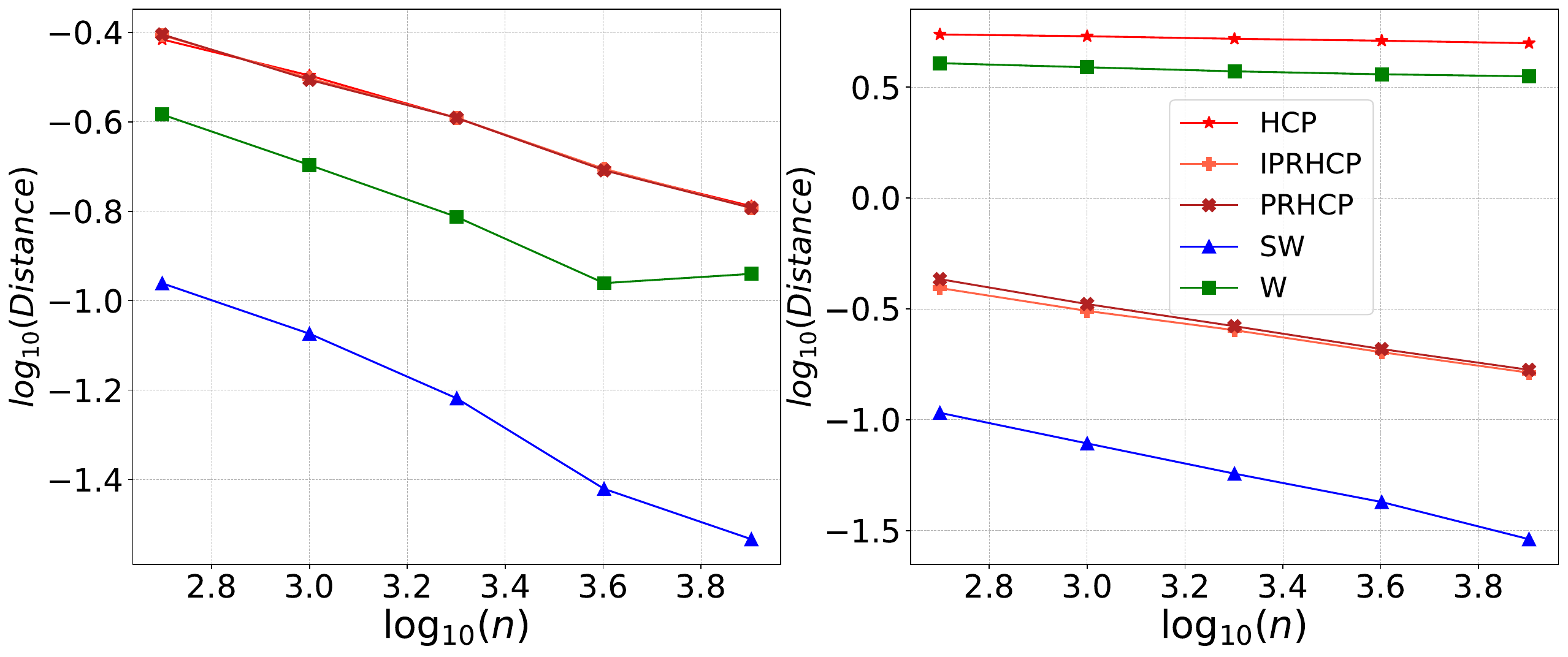}
    \caption{Comparison for sample complexity. 
    Left: $d=2$. Right: $d=20$. 
    Each curve represents a distance versus $n$.}
    \label{fig:iprhcp2}
\end{figure}

Additionally, we consider a synthetic example to demonstrate the empirical sample complexity of the proposed distances.
We generate two samples of size $n$ from the standard $d$-dimensional Gaussian distributions and we calculate the distances between these two samples w.r.t. different distance metrics.
Fig.~\ref{fig:iprhcp2} shows the average distances versus $n$ for $d=2$ and $20$, respectively.
We observe that when $d=20$, the Wasserstein distance and the HCP distance converge slowly as expected, while SW, IPRHCP, and PRHCP converge much faster. 
In the aspect of the empirical sample complexity, the slope of the curves indicates that our HCP distance is comparable to the Wasserstein distance, and our IPRHCP and PRHCP distances are comparable to the SW distance.

\subsection{Approximation of Wasserstein flow}
\subsubsection{Comparison on synthetic data}
\begin{figure*}[t]
    \centering
    \subfigure[]{
    \includegraphics[height=7cm]{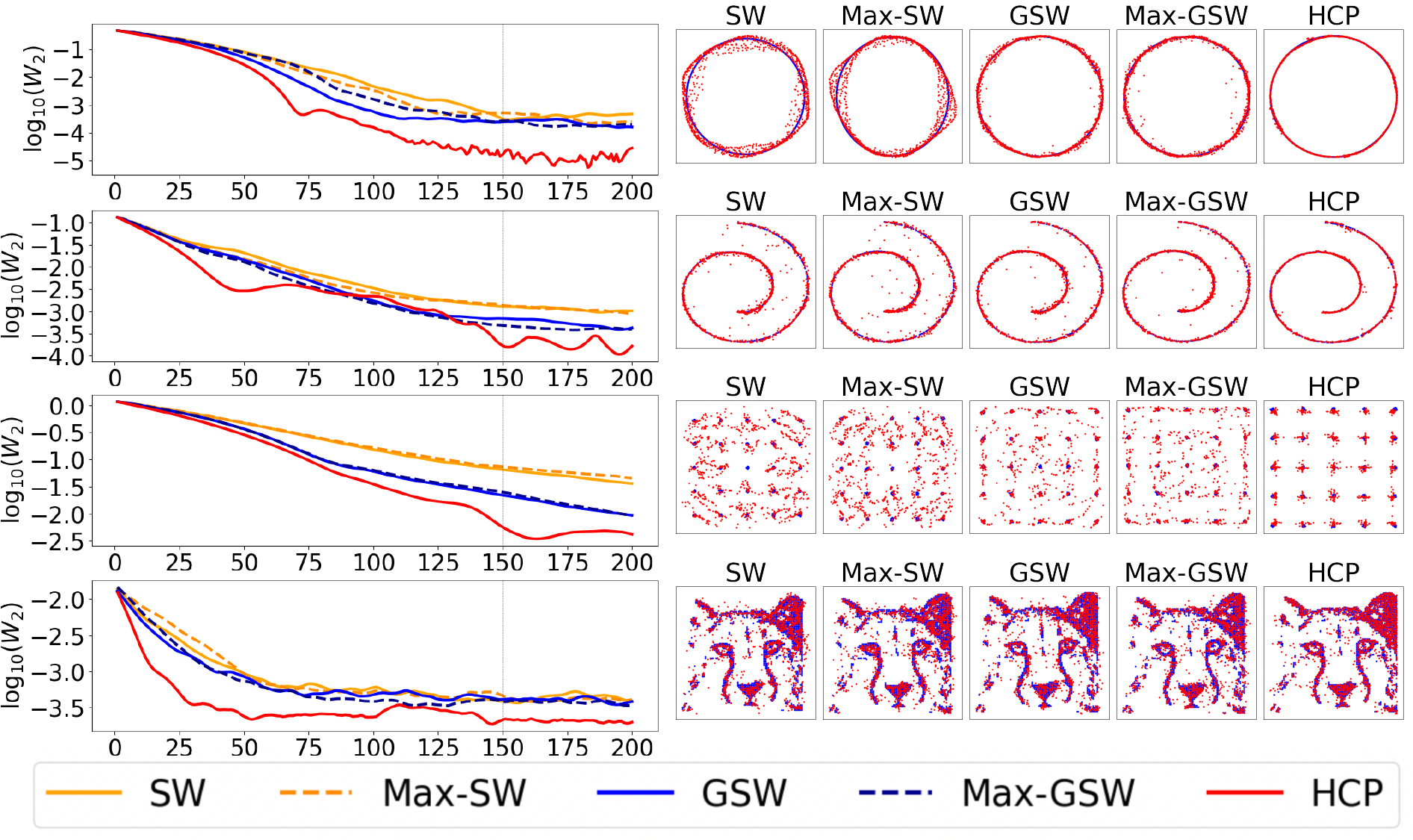}\label{fig:simu2}
    }
    \subfigure[]{
    \includegraphics[height=7cm]{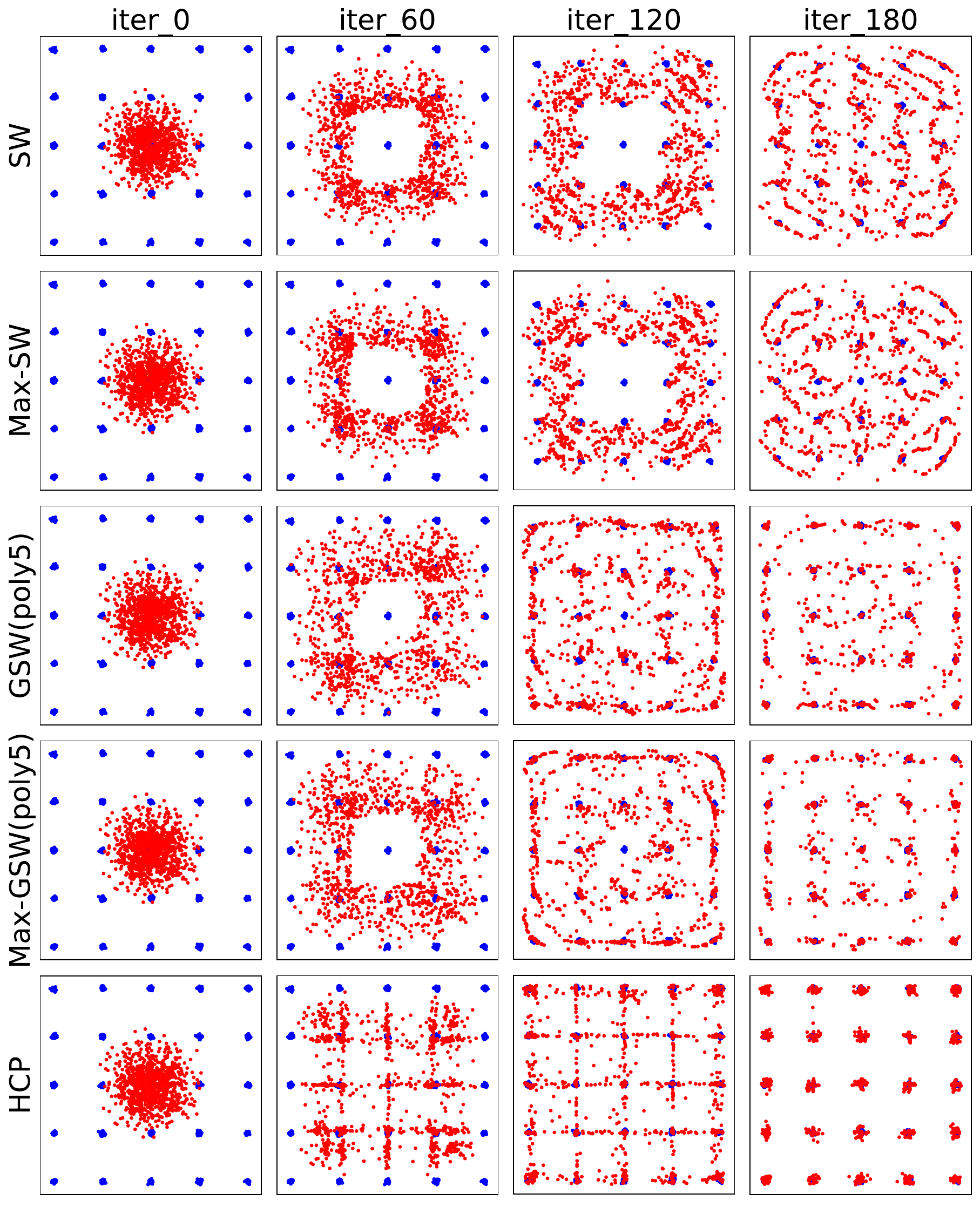}\label{fig:flow_supp}
    }
    \caption{(a) Left: Log 2-Wasserstein distance between the source and target distributions versus the number of iterations $t$. 
    Right: A snapshot when $t=150$.
    (b) Iterations of different distances based flow.}
\end{figure*}

Following the experiment in~\cite{kolouri2019generalized}, we consider the problem $\min_{\mu}\text{W}_2(\mu,\nu)$, where $\nu$ is a fixed target distribution, and $\mu$ is the source distribution initialized as $\mu_{0}=\mathcal{N}(0,1)$ and updated iteratively via $\partial_{t} \mu_{t}=-\nabla \text{W}_2\left(\mu_{t}, \nu\right)$. 
We consider four different distributions for the target $\nu$, i.e., \textit{Circle}, \textit{Swiss Roll}, \textit{25-Gaussian}, and \textit{Puma}, and approximate the Wasserstein distance $\text{W}_2$ by SW, max-SW, GSW, max-GSW, and HCP.
Each method applies one projection per iteration and sets the learning rate to be 0.01.
The experiments are replicated one hundred times, and we record the averaged 2-Wasserstein distance between $\mu_t$ and $\nu$ at each iteration. 
The comparison for the methods on their convergence curves and the snapshots of their learning results when $t=150$ are shown in Fig.~\ref{fig:simu2}.
We can find that applying HCP helps to accelerate the learning process and leads to better results.

Fig.~\ref{fig:simu2} shows that using SW or its variants as the loss function may lead to slow convergence.
Taking the \textit{25-Gaussian} case as an example, we provide an intuitive explanation for this phenomenon. 
In particular, we illustrate the iterations of SW, Max-SW, GSW, Max-GSW and HCP in Fig.~\ref{fig:flow_supp}.
We observe that the flow w.r.t. SW and its variants go through 2 processes: firstly, red points spread out without covering the central Gaussian; secondly, they cover the central Gaussian slowly.  
Such an observation indicates linear projection fails to preserve high-dimensional data structure, especially when the data are multi-modal, and thus resulting in slow convergence.
The proposed HCP distance, on the contrary, utilizes a Hilbert curve to preserve the structure of high-dimensional data and thus leads to faster convergence.

\begin{figure*}[t]
    \centering 
    \subfigure[]{
    \includegraphics[width=0.9\linewidth]{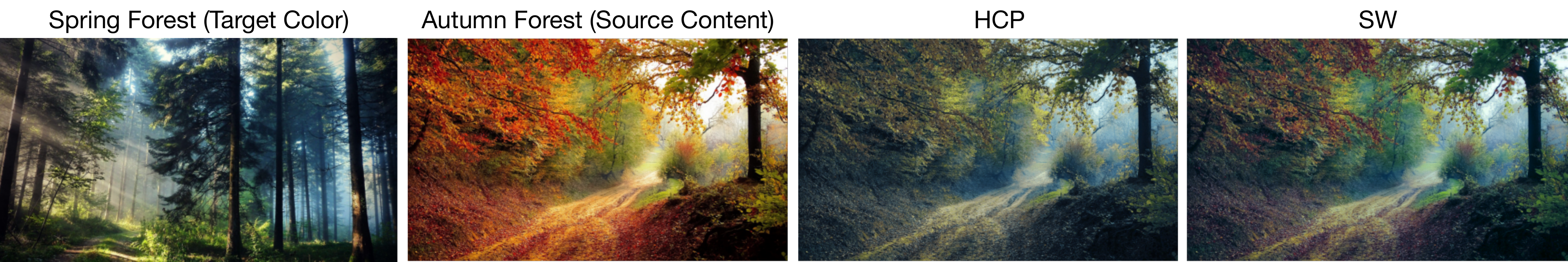}\label{fig:color_transfer1}
    }
    \subfigure[]{
    \includegraphics[width=0.9\linewidth]{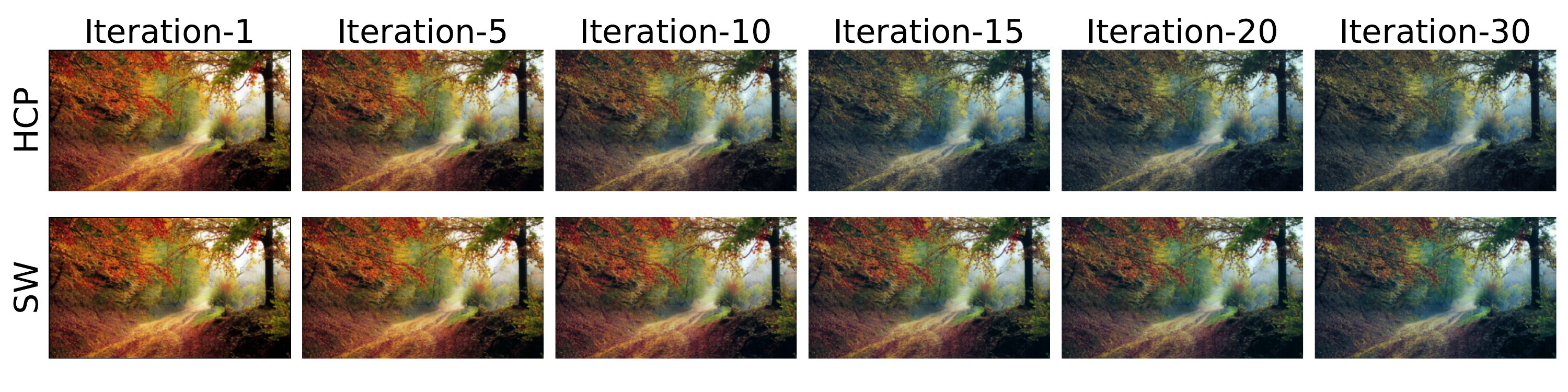}\label{fig:color_transfer2}
    }
    \caption{(a) The images from left to right are  the image with the target color, the image with the source content, and the color transfer results achieved based on our HCP distance and the SW distance, respectively. 
    (b) The first row is iterations of color transfer based on our HCP distance. 
    The second row is iterations of color transfer based on the SW distance.}
    \label{fig:color_transfer}
\end{figure*}

\subsubsection{Color transfer for images}
Besides testing on synthetic data, we consider the real-world color transfer task.
As shown in Fig.~\ref{fig:color_transfer1}, we transfer the color of a \textit{Spring Forest} image to an \textit{Autumn Forest} image. 
Each image is represented as nearly two million pixels in the RGB space ($d=3$). 
Considering the large sample size, we use SW distance and HCP distance to approximate the Wasserstein flow, with the same learning hyperparameters.
The comparison of the methods on their color transfer results and iterations are shown in Fig.~\ref{fig:color_transfer}.
We can find that applying the HCP distance helps to accelerate the learning process.
Quantitatively, it takes 496.7 seconds for the SW-based method and 57.3 seconds for our HCP-based method.

\subsection{Data classification}
\subsubsection{3D point cloud classification}
For low-dimensional data like 3D points, our HCP distance is superior to other distances in their classification tasks.
We consider the ModelNet10 dataset~\cite{wu20153d} that contains around 5,000 CAD objects from 10 categories.
For each category, we randomly sample 50 objects for training and 30 object objects for testing.
Following the work in~\cite{qi2017pointnet}, we randomly sample $n=100,200,500,1000,2000$ points per object to get 3D point cloud data. 
We calculate the pairwise distance between the point clouds w.r.t. different distance metrics and then use the K-NN algorithm ($n_{neighbors}=5$) to evaluate the classification accuracy on the testing set.
We used the RBF kernel for MMD, and we set the number of slices $n_s=10$ for SW, $n_s=10,T=7,\kappa=4$ for TSW. 
Here, $T$ is the predefined deepest level of the tree, $n_s$ is the number of slices and $\kappa$ is the number of clusters. 
Table~\ref{table:2} summarizes the averaged performance of each metric in 10 trials. 
Our HCP outperforms other distances on accuracy and requires the least amount of time.
\begin{table*}[t]
  \caption{Comparisons on 3D point cloud classification}
  \label{table:2}
  \centering
  \small{
  \begin{threeparttable}
  \begin{tabular}{ccccccccccccc}
    \toprule
    &\multicolumn{5}{c}{Accuracy(\%)} &\multicolumn{5}{c}{CPU time(s)} \\
    \cmidrule(lr){2-6}\cmidrule(lr){7-11}
    Method     & n=100  & n=200  & n=500 & n=1000  & n=2000   & n=100  & n=200  & n=500 & n=1000  & n=2000  \\
    \midrule
    HCP &   73.3$_{\pm1.3}$ &  79.2$_{\pm2.8}$ & \textbf{81.8$_{\pm0.2}$}  & \textbf{81.0$_{\pm0.1}$} & \textbf{82.3$_{\pm0.7}$} & \textbf{17.9} & \textbf{34.3} & \textbf{90.4} & \textbf{186.7} & 374.5 \\
    SW &   71.5$_{\pm2.2}$ &  77.0$_{\pm1.0}$ & 79.2$_{\pm0.5}$ & 79.3$_{\pm0.7}$ & 80.5$_{\pm1.0}$ & 123.9 & 202.1 & 410.5 & 830.8 & 1808.7 \\
    TSW &  72.7$_{\pm3.2}$ &  75.0$_{\pm2.5}$ & 77.0$_{\pm2.2}$ & 77.7$_{\pm1.5}$ & 78.0$_{\pm1.0}$ & 120.7 & 137.4 & 165.8 & 217.2 & \textbf{282.1}  \\
    GSW(Poly5) &  68.3$_{\pm2.0}$ &  73.2$_{\pm0.5}$ & 76.0$_{\pm0.7}$ & 76.8$_{\pm1.2}$ & 77.7$_{\pm0.3}$ & 839.7 & 940.6 & 1228.8 & 1782.4 & 2970.2  \\
    MMD &  66.8$_{\pm4.2}$ &  71.8$_{\pm0.2}$ & 74.0$_{\pm0.7}$ & / & / & 153.6 & 385.7 & 1785.2 & / & /  \\
    Sinkhorn &  \textbf{77.0$_{\pm2.7}$} &  \textbf{80.8$_{\pm0.5}$} & 81.5$_{\pm0.8}$ & / & / & 768.7 & 2305.0 & 6184.6 & / & /  \\
    \bottomrule
  \end{tabular}
  \begin{tablenotes}
    \item[*] ``/'' means that we fail to get a result in 10,000 seconds.
    \end{tablenotes}
\end{threeparttable}
  }
\end{table*}

\subsubsection{Document classification}
As a typical high-dimensional data classification problem, document classification can be achieved by comparing the Wasserstein distance between two documents' word embedding sets, as the Word Mover distance~\cite{kusner2015word} does.
Our PRHCP distance provides an efficient surrogate of the Wasserstein distance in this problem, which is demonstrated by the following experiment. 
Following the preprocessing used in~\cite{kusner2015word}, we obtain 3,000 documents belonging to three categories from the TWITTER dataset, in which each document is represented as a set of 300-dimensional word embeddings derived by the pre-trained \textit{word2vec} model~\cite{mikolov2013distributed}.
We randomly split the dataset into 80\% for training and 20\% for testing.
Similar to the above point cloud classification experiment, we use the K-NN algorithm ($n_{neighbors}=10$) based on different metrics and evaluate the averaged learning results in 10 trials.
In this experiment, we set the number of slices $n_s=20$ for SW, $n_s=10,T=7,\kappa=4$ for TSW.
For PRHCP, we first find the 10-dimensional subspace based on the training data by Algorithm~\ref{alg:ALG2} and project testing data to the subspace. 
Table~\ref{table:3} shows that our PRHCP distance outperforms other distances on classification accuracy, and its runtime is comparable to TSW.

\begin{table}
  \caption{Comparisons on document classification}
  \label{table:3}
  \centering
  \small{
  \begin{tabular}{ccc}
    \toprule
    Method     & Accuracy(\%)  & CPU time(s)  \\
    \midrule
    PRHCP &   \textbf{74.6$_{\pm0.9}$} & 669.7\\
    TSW &   71.2$_{\pm0.6}$ &  \textbf{287.7} \\
    Sinkhorn &  70.0$_{\pm0.4}$ &  10106.7  \\
    SW &  68.6$_{\pm1.1}$ &  2789.7 \\
    \bottomrule
  \end{tabular}
  }
\end{table}

\begin{figure*}[t]
    \centering
    \includegraphics[width=0.95\textwidth]{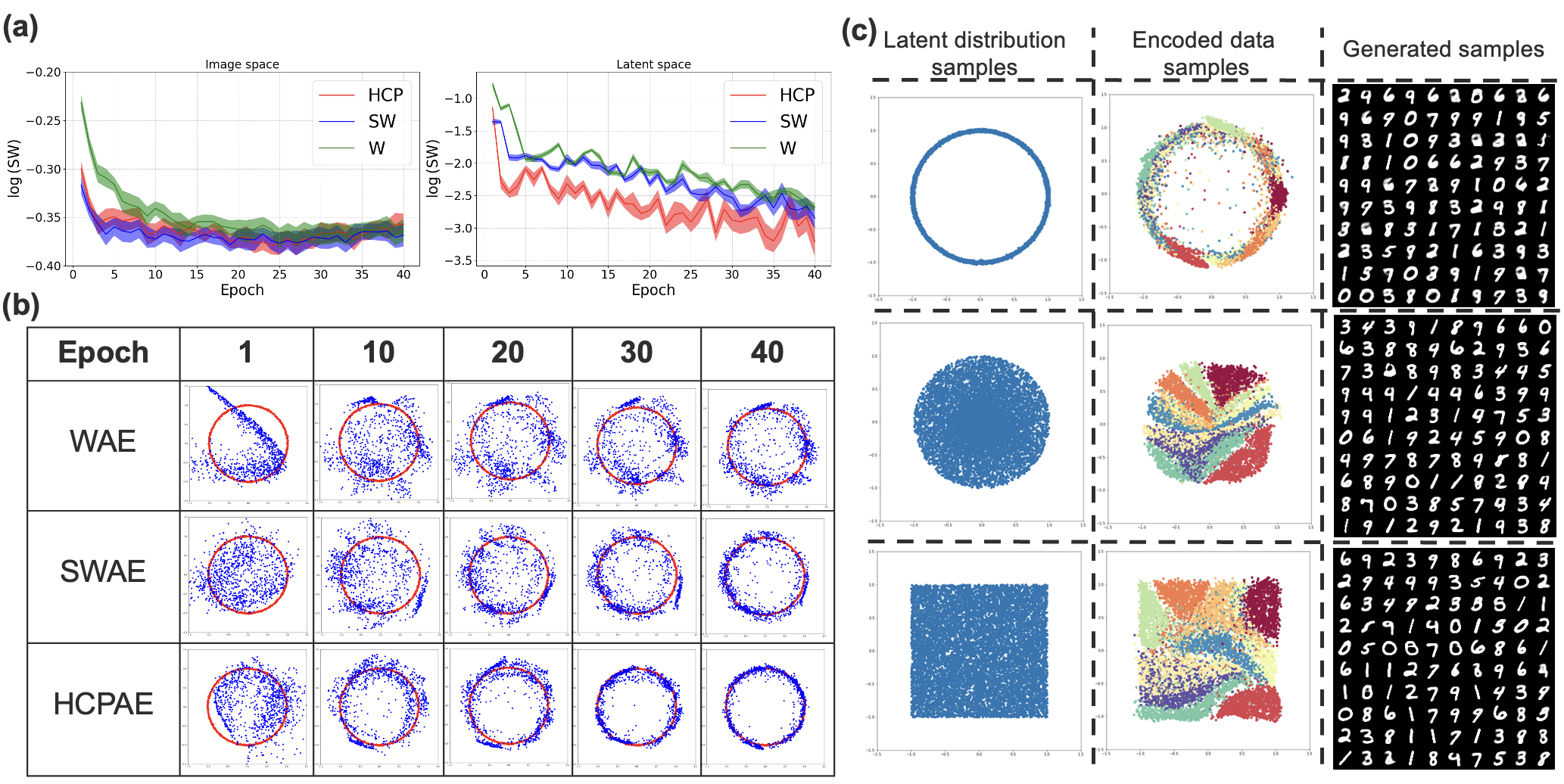}
    \caption{(a): SW distances between the target sample and the encoded testing sample w.r.t. the image space (left) and the latent space (right); (b) Visualization of these two sample in the latent space during training; (c) Visualization of the encoded samples and the generated images.}
    \label{fig:hcp_real11}
\end{figure*}

\subsection{Generative modeling}
The proposed distances help us to design new members of Wasserstein autoencoder (WAE)~\cite{tolstikhin2018wasserstein}. 
In particular, when training autoencoders, we leverage HCP, IPRHCP, and PRHCP to penalize the distance between the latent prior distribution and the expected posterior distribution, which leads to three different generative models, denoted as HCP-AE, IPRHCP-AE, and PRHCP-AE. 
We test these three models in image generation tasks and compare them with the original Wasserstein autoencoder (WAE)~\cite{tolstikhin2018wasserstein} and the well-known sliced Wasserstein autoencoder (SWAE)~\cite{kolouri2018sliced}. 

\subsubsection{HCP-based autoencoders}
We first test the capability of HCP-AE in shaping the low-dimensional latent space of the encoder.
We train an HCP-AE to encode the MNIST dataset~\cite{lecun1998gradient} to a two-dimensional latent space (for the sake of visualization), in which both the autoencoding architecture and the hyperparameter setting are the same as those in~\cite{kolouri2018sliced}.
A simple autoencoder with mirrored classic deep convolutional neural networks with 2D average poolings, Leaky-ReLu activation functions, and upsampling layers in the decoder is used.
The batch size is 500 and the number of projections for SWAE is 40.

To evaluate the performance, we randomly selected a sample of size 1,000 from the encoded test data points (blue points in Fig.~\ref{fig:hcp_real11}(b)) and a random sample from the target prior distribution in the latent space (red points in Fig.~\ref{fig:hcp_real11}(b)).
We observed that our HCP-AE convergences much faster than other methods in the latent space.
Moreover, the SW distances versus the number of epochs w.r.t. the image space and the latent space are shown in Fig.~\ref{fig:hcp_real11}(a). 
We observed that though these three methods perform similarly in the image space, our HCP-AE converges much faster in the latent space.
Fig.~\ref{fig:hcp_real11}(c) visualizes the samples from two different prior distributions in the latent space, the encoded data samples via HCP-AE, and their generated images.
The latent codes indeed obey the prior distributions, which reflects the clustering structure of the digits. 
Accordingly, the learned models are able to generate high quality digit images.

\begin{figure*}[t]
	\centering
	\subfigure[IPRHCP-AE: face generation]{
			\includegraphics[height=4.3cm]{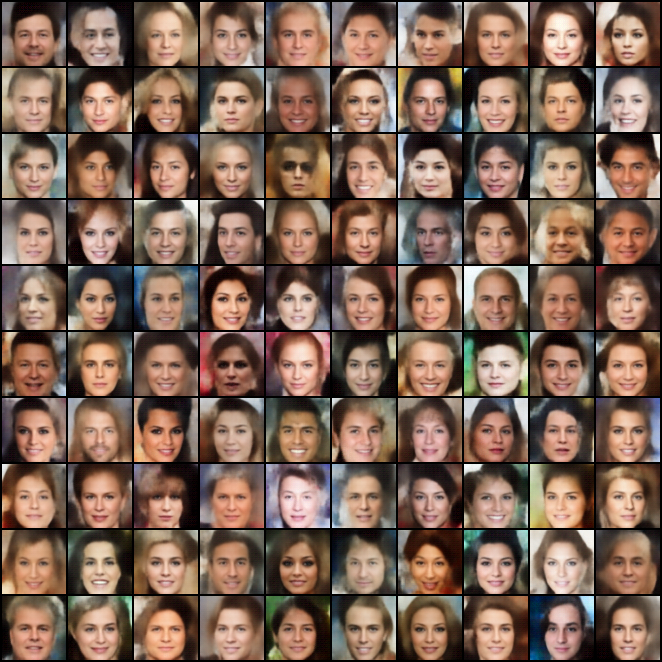}
			\includegraphics[height=4.3cm]{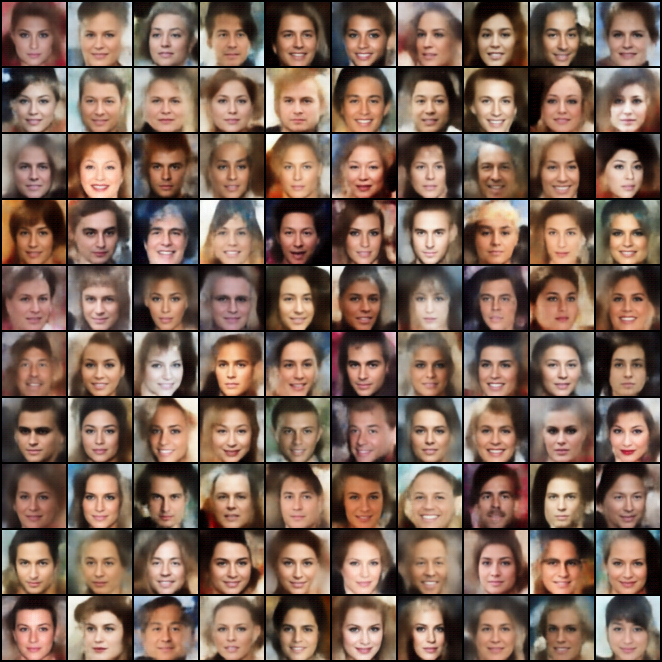}
	}
	\subfigure[IPRHCP-AE: face interpolation]{
			\includegraphics[height=4.3cm]{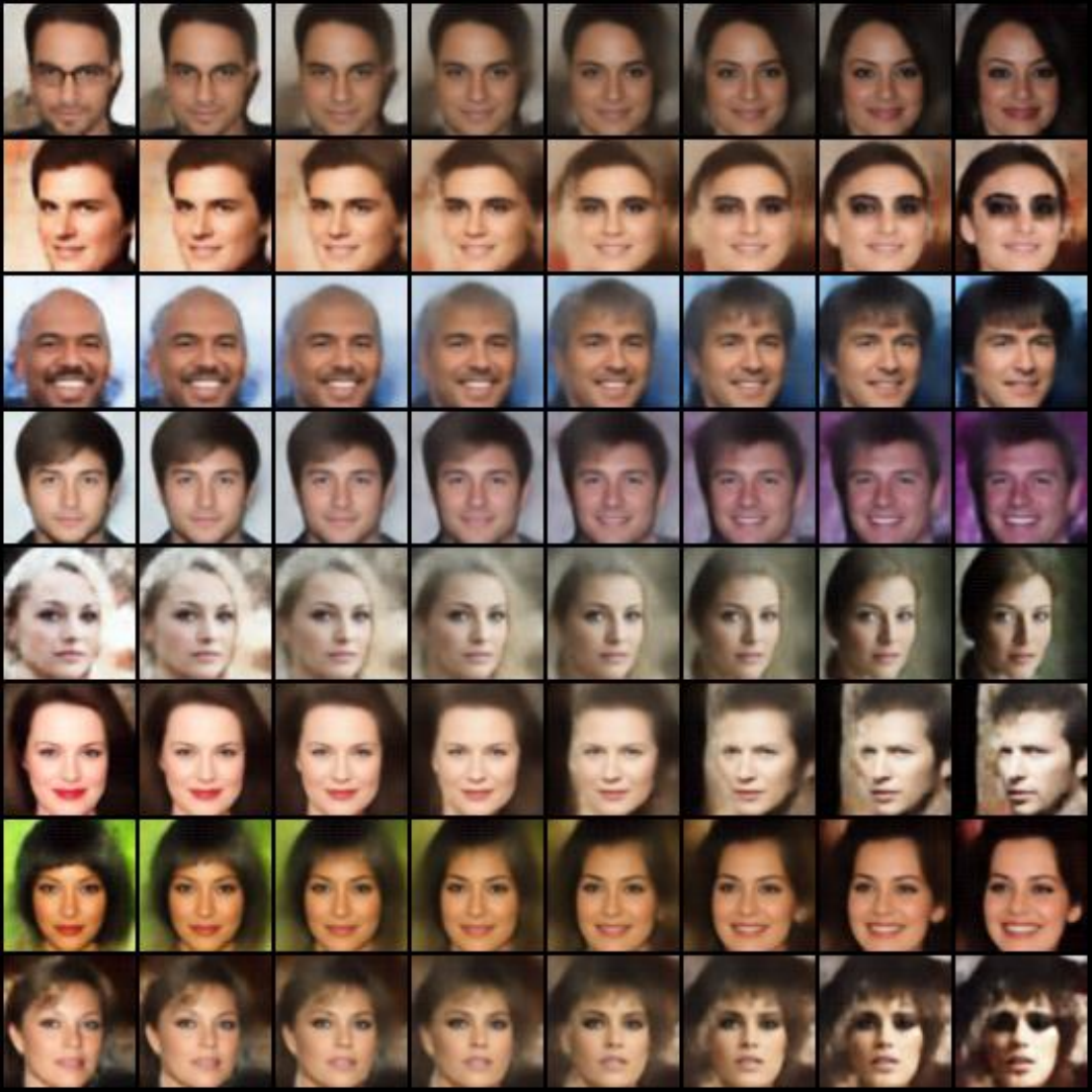}
			\includegraphics[height=4.3cm]{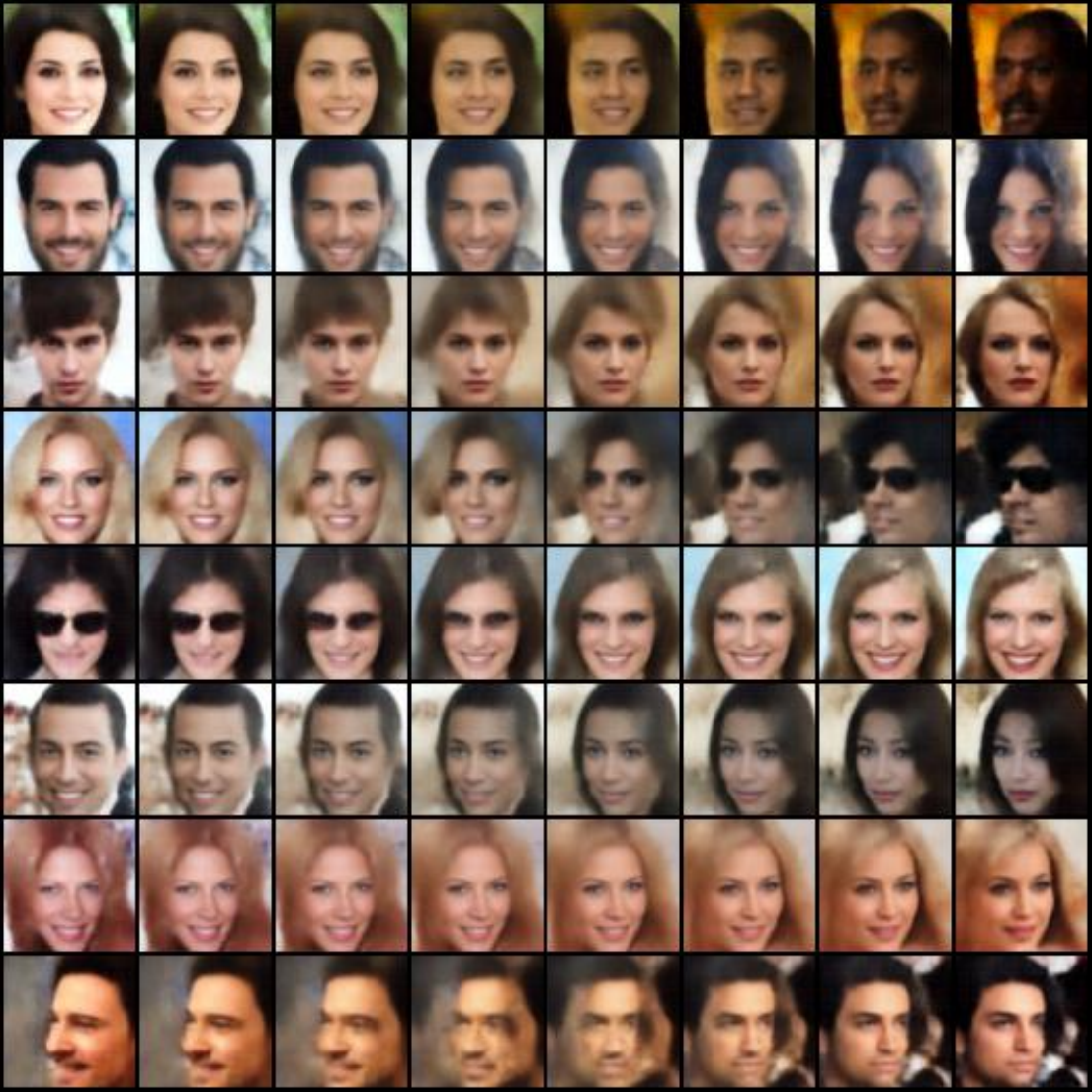}
	}\\
	\subfigure[PRHCP-AE: face generation]{
			\includegraphics[height=4.3cm]{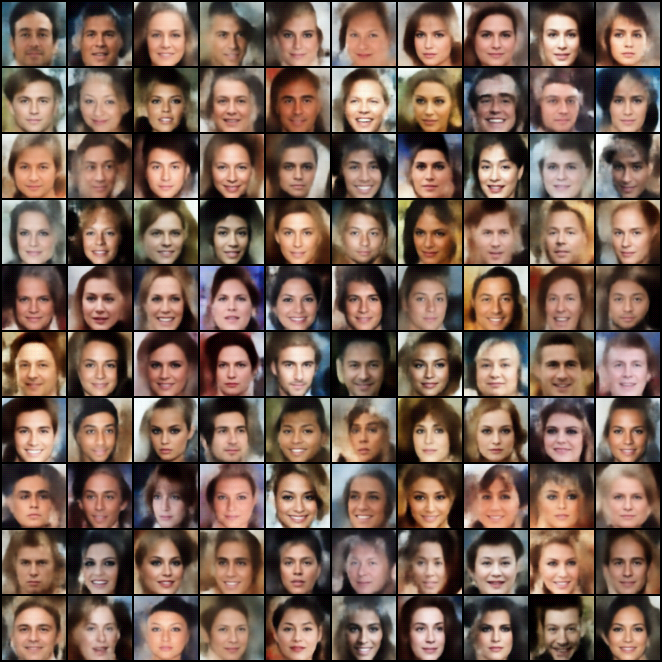}
			\includegraphics[height=4.3cm]{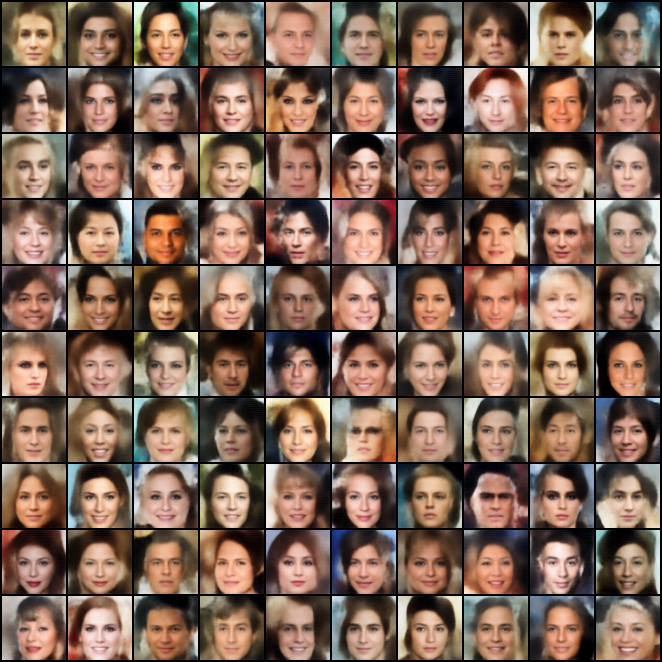}
	}
	\subfigure[PRHCP-AE: face interpolation]{
			\includegraphics[height=4.3cm]{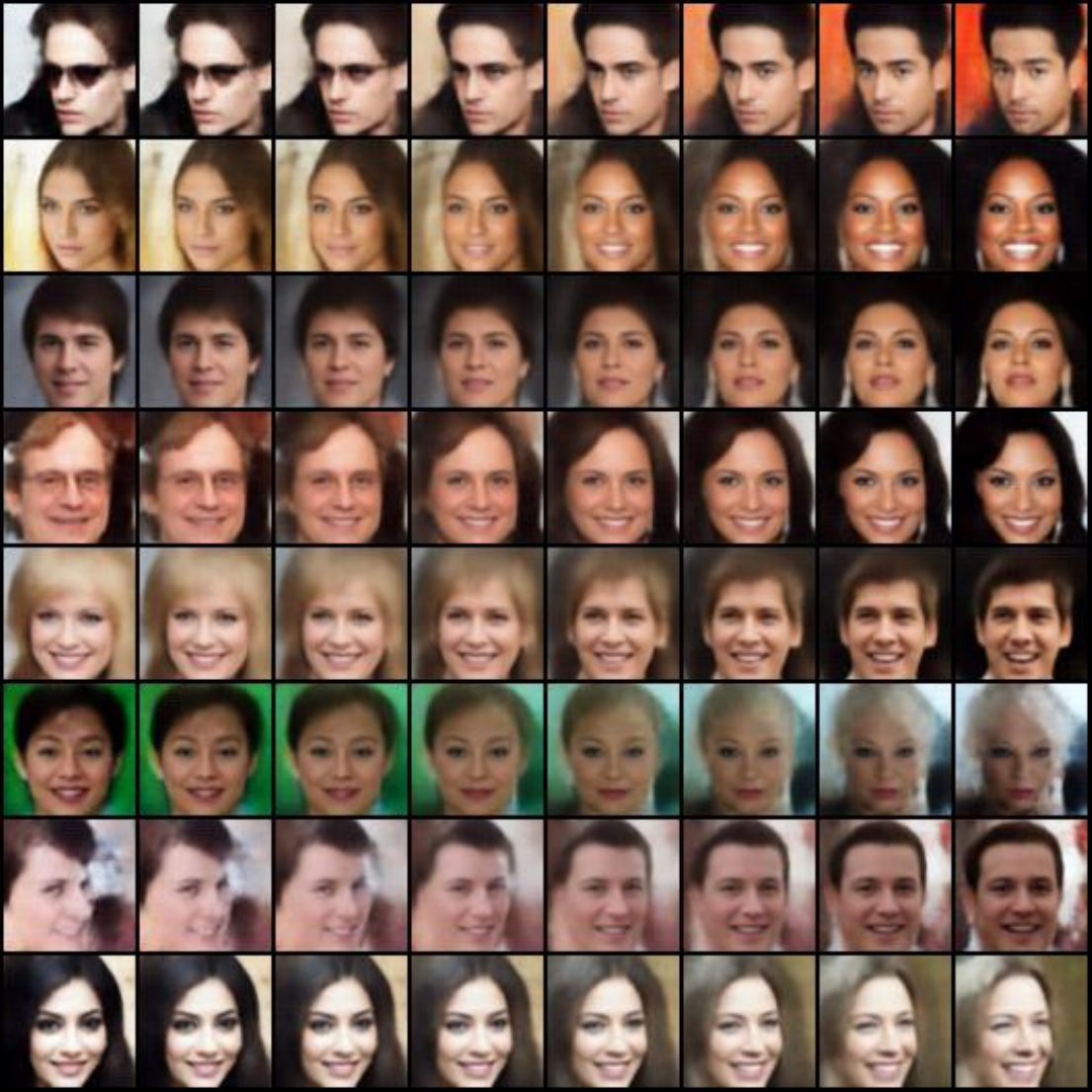}
			\includegraphics[height=4.3cm]{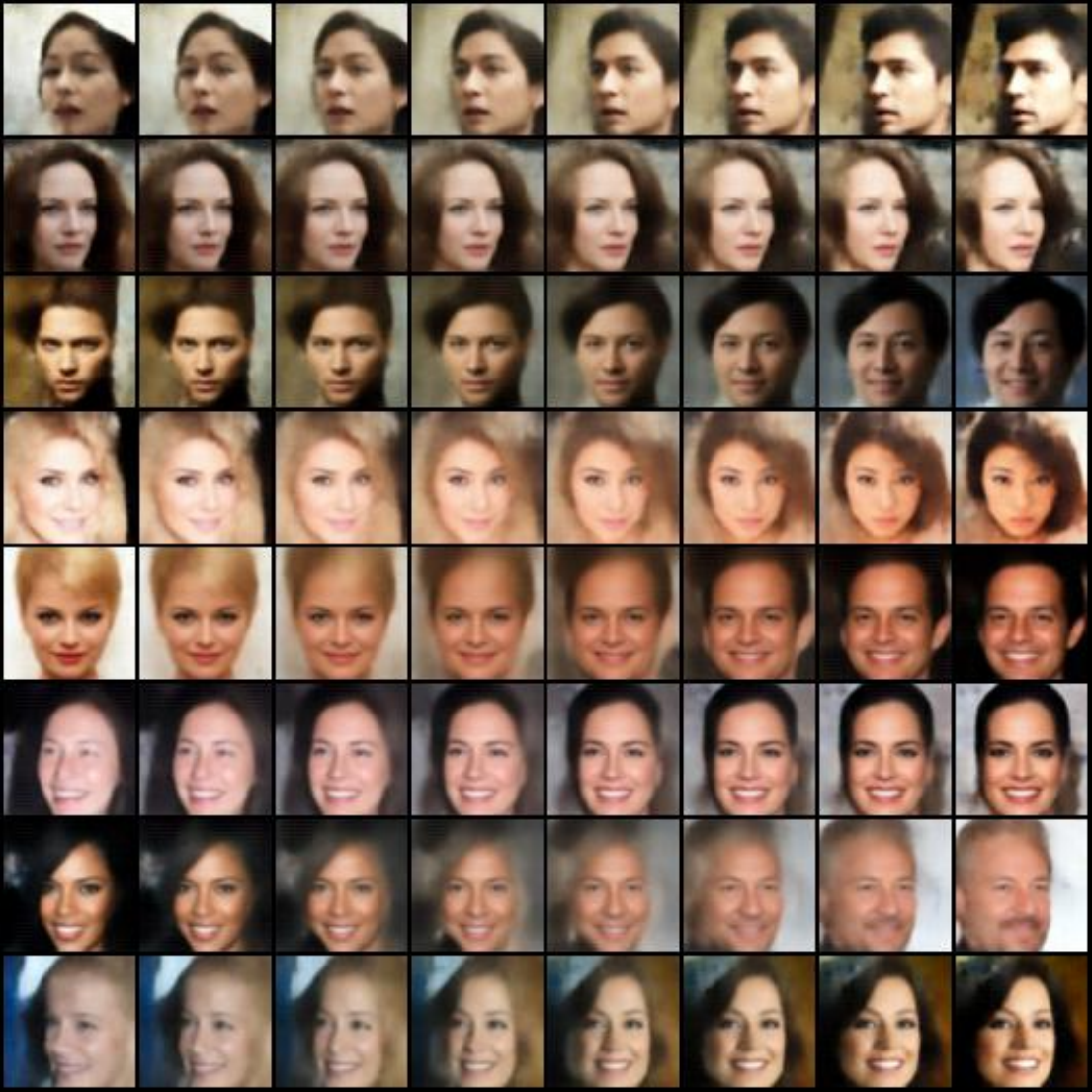}
	}
	\caption{The performance of IPRHCP-AE and PRHCP-AE on face generation and interpolation.}
	\label{fig:intp}
\end{figure*}

\subsubsection{IPRHCP and PRHCP-based autoencoders}
Secondly, we test the feasibility of IPRHCP-AE and PRHCP-AE in the cases with high-dimensional latent space.
For fairness, all the autoencoders have the same DCGAN-style architecture~\cite{radford2015unsupervised} and hyperparameters: the learning rate is 0.001; the optimizer is Adam~\cite{kingma2014adam} with $\beta_1 = 0.9$ and $\beta_2 = 0.999$; the number of epochs is 50; the batch size is 100; the weight of regularizer $\gamma$ is 1; the dimension of latent code is 8 for MNIST and 64 for CelebA; the number of random projections is 50.
All the autoencoders use Euclidean distance as the distance between samples, which means the reconstruction loss is the mean-square error (MSE).
We compare the proposed methods with the baselines on $i$) the reconstruction loss on testing samples;
$ii$) the Fr\'echet Inception Distance (FID)~\cite{heusel2017gans} between 10,000
testing samples and 10,000 randomly generated samples.
Table~\ref{table:1} lists the main
differences between IPRHCP-AE, PRHCP-AE and these baselines.
Among these autoencoders, our IPRHCP-AE and PRHCP-AE are comparable to the considered alternatives on both testing reconstruction loss and FID score.
Some image generation and interpolation results achieved by our methods are shown in Fig.~\ref{fig:intp}.

\begin{table}[t]
  \caption{Comparisons for various methods on learning image generators}\label{table:1}
  \centering
  \begin{small}
  \begin{tabular}{ccccc}
    \toprule
    \multirow{2}{*}{Method} &\multicolumn{2}{c}{\textbf{MNIST}} &\multicolumn{2}{c}{\textbf{CelebA}} \\
    \cmidrule(lr){2-5}
    & Rec. loss  & FID  & Rec. loss  & FID  \\
    \midrule
    WAE &   11.30 &  54.61$_{\pm0.16}$ & 68.94 & 58.12$_{\pm0.73}$ \\
    SWAE &   13.68 &  42.96$_{\pm0.53}$ & 68.57 & 84.52$_{\pm0.44}$ \\
    IPRHCP-AE &  11.72 &  \textbf{40.03}$_{\pm0.13}$  & 69.40 & \textbf{56.00}$_{\pm0.08}$ \\
    PRHCP-AE &  \textbf{10.07} &  42.87$_{\pm0.46}$ & \textbf{66.65} & 67.82$_{\pm0.21}$ \\
    \bottomrule
  \end{tabular}
  \end{small}
\end{table}

\section{Conclusion}
In this work, we proposed a novel metric for distribution comparison, named Hilbert curve projection (HCP) distance.
Thanks to the locality-preserving property of the Hilbert curve projection, the HCP distance enjoys several advantages over the Wasserstein and SW distance.
Furthermore, we develop two variants of the HCP distance using (learnable) subspace projections to mitigate the curse-of-dimensionality.

\textbf{Limitations and future work.}
Currently, HCP distance still suffers from some limitations.
Like the Wasserstein distance, the HCP distance may not be robust to outliers.
To address this problem, we could follow the methods in \cite{balaji2020robust,mukherjee2021outlier,le2021robust}
by relaxing marginal constraints through penalty functions such as Kullback-Leibler divergence, total variation distance, and $\chi^2$ divergence.
Another possible solution is to consider partial OT methods instead of sorting in Step 3 of Algorithm 1.
Besides, HCP distance could not quantify the discrepancy between two measures with different masses.
We could follow the idea of (sliced) unbalanced optimal transport \citep{pham2020unbalanced,chizat2018scaling,sejourne2023unbalanced} by considering unbalanced OT methods instead of sorting in Step 3 of Algorithm 1. We left these directions for our future work.
In addition, we plan to apply these new metrics to more learning problems and extend them to Gromov-Wasserstein distance~\cite{memoli2011gromov}, multi-marginal optimal transport~\cite{pass2015multi,haasler2021multi}, and barycenter problems~\cite{agueh2011barycenters,peyre2016gromov}.
Additionally, we will explore the theoretical results for other formulations of the Hilbert curve, such as the adaptive Hilbert curve, which works well in practice.
And there is much literature on Hilbert sort, such as parallel Hilbert sort~\cite{luitjens2007parallel} and online Hilbert sort~\cite{imamura2016fast,kamel1993hilbert}, which may be extended.

\section*{Acknowledgment}
All authors contributed equally and the order of authors' names is alphabetical.
The authors would like to acknowledge the support from National Natural Science Foundation of China Grant No.62106271, No.12101606, No.12001042, and Renmin University of China research fund program for young scholars, and Beijing Institute of Technology research fund program for young scholars.


%

\appendices
\section{The Proofs of Theorems and Corollaries}
\subsection{Preliminaries}
The Hilbert curve enjoys three properties, see~\cite{zumbusch2012parallel,he2016extensible} for details.
\begin{enumerate}[noitemsep,topsep=0pt]
    \item Let $\lambda_{d}$ be the $d$-dimensional Lebesgue measure. 
    For any measurable set $\Omega \subseteq[0,1]$, one has $\lambda_{1}(\Omega)=\lambda_{d}(H(\Omega))$.
    \item If a random variable $Z$ follows the uniform distribution on $[0,1]$, then $H(Z)$ follows the uniform distribution on $[0,1]^{d}$. 
    This is to say, one has $\int_{[0,1]^{d}} f(z) \mathrm{d} z=\int_{0}^{1} f[H(z)] \mathrm{d} z$.
    \item For any $x, y \in[0,1]$, one has $\|H(x)-H(y)\|_2 \leq 2 \sqrt{d+3}|x-y|^{1 / d}.$  
\end{enumerate}

The H{\"o}lder continuous property also holds for $H_\mu$.
Considering that $H_\mu$ is a linear stretching of $H$ and using Cauchy–Schwarz inequality, we have $\|H_\mu(x)-H_\mu(y)\|_2 \leq C_\mu\|H(x)-H(y)\|_2 \leq 2C_\mu \sqrt{d+3}|x-y|^{1 / d}$  for any $x, y \in[0,1]$.
And by equivalence of vector norms, we have $\|H_\mu(x)-H_\mu(y)\|_p \leq C_p\|H_\mu(x)-H_\nu(y)\|_2 \leq 2C_\mu C_p \sqrt{d+3}|x-y|^{1 / d}$, where $C_p,C_\mu$ are two constants.

Before presenting the main statistical theoretical results of HCP distance, we first provide an essential lemma.
\begin{lemma}
If a random variable $Z$ follows the probability distribution function $g_\mu$, then $H_\mu(Z)$ follows the probability measure $\mu$.
\label{lemma}
\end{lemma}
\begin{proof}[\textbf{Proof of Lemma~\ref{lemma}}]

By the definition of $g_\mu$, we can easily know that
$Pr(Z\in A) = \inf\{\mu(A'):A'\text{ is a Borel set in } \RR^d, H_\mu(A\backslash \mathcal{N})\subset A'\}$
for any Borel measurable set $A\subset \mathbb{R}$.
And for any measurable set $B\subset \mathbb{R}^d$, $Pr(H_\mu(Z)\in B)= Pr(Z \in H_\mu^{-1}(B)\backslash \mathcal{N})=\inf\{\mu(A'):A'\text{ is a Borel set in } \RR^d, H_\mu(H_\mu^{-1}(B)\backslash \mathcal{N})\subset A'\}=\inf\{\mu(A'):A'\text{ is a Borel set in } \RR^d, B\subset A'\}=\mu(B)$.
\end{proof}

\subsection{The proofs related to HCP distance}
It is clear that HCP distance finds a transport plan between two probability measures by Hilbert sorting.
Thus, it obviously gives an upper bound of Wasserstein distance.
Below, we clarify it theoretically and prove HCP distance is a well-defined distance.

\begin{proof}[\textbf{Proof of Theorem~\ref{thm:HCP}}]
Here, we first borrow the Corollary~\ref{coro:conv1} proven below. 
It tells us that 
$\mathbb{E}\frac{1}{n}\sum_{i=1}^n \|x_{{(i)}^*}-y_{{(i)}^*}\|_p^p\to \text{HCP}^p_p(\mu, \nu)$, as $n \to \infty$.
Additionally, we know that $\mathbb{E}\inf_{\sigma}\frac{1}{n}\sum_{i=1}^n \|x_{i}-y_{\sigma{(i)}}\|_p^p\to \text{W}^p_p(\mu, \nu)$, as $n \to \infty$.
Clearly, $\inf_{\sigma}\frac{1}{n}\sum_{i=1}^n \|x_{i}-y_{\sigma{(i)}}\|_p^p \leq \frac{1}{n}\sum_{i=1}^n \|x_{{(i)}^*}-y_{{(i)}^*}\|_p^p$.
Hence, we have $W_p(\mu,\nu)\leq \text{HCP}_p(\mu,\nu)$.

By the definition of HCP, it's clear that $\text{HCP}_p(\mu,\nu)\geq 0$ and $\text{HCP}_p(\mu,\nu)=HCP_p(\nu,\mu)$.
Because Wasserstein distance is a well-define metric, we can easily get $\text{HCP}_p(\mu,\nu)=0$ if and only if $\mu=\nu$.
Next, we prove the triangle inequality.
\begin{align*}
 &\text{HCP}_p(\mu, \nu) \\
 =& \left(\int_0^1{\|H_\mu(g_\mu^{-1}(t))-H_\nu(g_\nu^{-1}(t))\|}_p^p\mathrm{d}t\right)^{1/p}\\
 =& \left(\int_0^1\theta_{\mu,\nu}(t)^p\mbox{d}t\right)^{1/p}\\
 \leq& \left(\int_0^1\theta_{\mu,\tau}(t)^p\mbox{d}t\right)^{1/p} + \left(\int_0^1|\theta_{\mu,\nu}(t)-\theta_{\mu,\tau}(t)|^p\mbox{d}t\right)^{1/p}\\
 \leq& \left(\int_0^1\theta_{\mu,\tau}(t)^p\mbox{d}t\right)^{1/p} + \left(\int_0^1\theta_{\nu,\tau}(t)^p\mbox{d}t\right)^{1/p}\\
 =&\text{HCP}_p(\mu, \tau)+\text{HCP}_p(\nu, \tau),
\end{align*}
where first inequality uses Minkowski inequality and the second uses triangle inequality.
\end{proof}

\begin{proof}[\textbf{Proof of Theorem~\ref{thm:topo}}]
    From the proof in Theorem~\ref{thm:conv1}, we have for some $\alpha>0$,
    \begin{align*}
        \text{HCP}_p(\mu_n,\mu)
        \leq \Bigl(\int_{0}^{1}|g_{\mu_n}^{-1}(t)-g_\mu^{-1}(t)|dt\Bigr)^{\alpha}
        =W_1(\widetilde{\mu_n},\widetilde{\mu})^\alpha ,      
    \end{align*}
    where the probability measure corresponding to $g_{\mu_n}$ is $\widetilde{\mu_n}$ and the probability measure corresponding to $g_{\mu}$ is $\widetilde{\mu}$.
    
    Since $TV(\mu_n,\mu)\to 0$, we have $\forall \epsilon$, there exists a $N_0$, $\sup_A|\mu_n(A)-\mu(A)|<\epsilon$ when $n>N_0$.
    For any measurable set $B\subset [0,1]$, we have
    $|\widetilde{\mu_n}(B)-\widetilde{\mu}(B)|<\epsilon$ when $n>N_0$.
    Hence, we know $W_1(\widetilde{\mu_n},\widetilde{\mu})\to 0$, and the proof is done. 
\end{proof}

\begin{proof}[\textbf{Proof of Theorem~\ref{thm:conv1}}]

    Let $\{y_i\}_{i=1}^n$ be an i.i.d sample, which is generated from probability distribution function $g_\mu$.
    By Lemma~\ref{lemma}, we could replace $\{x_i\}_{i=1}^n$ with $\{H_\mu(y_i)\}_{i=1}^n$, and thus,replace $\hat{g}_{\mu_n}(t)$ with $g_n(t)=\frac{1}{n}\sum_{i=1}^n\mathbf{1}_{\{y_i\leq t\}}$.
    Hence, to study the convergence of $\overline{\text{HCP}}_p(\mu,\mu_n)=(\int_{0}^{1}\|H_\mu(g_{\mu}^{-1}(t))-H_\mu(\hat{g}_{\mu_n}^{-1}(t))\|_p^p\mathrm{d}t)^{\frac{1}{p}}$, we only need to study $(\int_{0}^{1}\|H_\mu(g_{\mu}^{-1}(t))-H_\mu({g}_{n}^{-1}(t))\|_p^p\mathrm{d}t)^{\frac{1}{p}}$.
    
    We know that
    \begin{align*}
        &\left(\int_{0}^{1}\|H_\mu(g_{\mu}^{-1}(t))-H_\mu({g}_{n}^{-1}(t))\|_p^p\mathrm{d}t\right)^{\frac{1}{p}}\\
        \lesssim & \left(\int_{0}^{1}|g_n^{-1}(t)-g_\mu^{-1}(t)|^{\frac{p}{d}}\mathrm{d}t\right)^{\frac{1}{p}},
    \end{align*}
    and we have when $p\geq d$
    \begin{align*}
        \left(\int_{0}^{1}|g_n^{-1}(t)-g_\mu^{-1}(t)|^{\frac{p}{d}}\mathrm{d}t\right)^{\frac{1}{p}}
        & \leq \left(\int_{0}^{1}|g_n^{-1}(t)-g_\mu^{-1}(t)|\mathrm{d}t\right)^{\frac{1}{p}}.
    \end{align*}
    When $p<d$, by Jenson's inequality, we have 
    \begin{align*}
        \left(\int_{0}^{1}|g_n^{-1}(t)-g_\mu^{-1}(t)|^{\frac{p}{d}}\mathrm{d}t\right)^{\frac{1}{p}}
        & \leq \left(\int_{0}^{1}|g_n^{-1}(t)-g_\mu^{-1}(t)|\mathrm{d}t\right)^{\frac{1}{d}}.
    \end{align*}
    However, $\int_{0}^{1}|g_n^{-1}(t)-g_\mu^{-1}(t)|\mathrm{d}t$ is the 1-Wasserstein distance.
    Considering $g_\mu$ has bounded support, we know that $\int_{0}^{1}|g_n^{-1}(t)-g_\mu^{-1}(t)|\mathrm{d}t\to 0$ almost surely.
    Thus, we know that $\overline{\text{HCP}}_p(\mu,\mu_n)\to 0$ almost surely.
    
    Furthermore, we know that
    \begin{eqnarray*}
    \begin{aligned}
        \mathbb{E}\overline{\text{HCP}}_p(\mu,\mu_n)
        =&\mathbb{E}\left(\int_{0}^{1}\|H_\mu({g}_{n}^{-1}(t))-H_\mu(g_\mu^{-1}(t))\|_p^pdt\right)^{\frac{1}{p}}\\
        \lesssim& \mathbb{E}\left(\int_{0}^{1}|g_n^{-1}(t)-g_\mu^{-1}(t)|^{\frac{p}{d}}dt\right)^{\frac{1}{p}}\\
        \leq& \left(\mathbb{E}\int_{0}^{1}|g_n^{-1}(t)-g_\mu^{-1}(t)|^{\frac{p}{d}}dt\right)^{\frac{1}{p}}.
    \end{aligned}
    \end{eqnarray*}
    If $p\geq d$, we know that 
    \begin{eqnarray*}
    \begin{aligned}
        \mathbb{E}\overline{\text{HCP}}_p(\mu,\mu_n)
        \lesssim&\left(\mathbb{E}\int_{0}^{1}|g_n^{-1}(t)-g_\mu^{-1}(t)|^{\frac{p}{d}}dt\right)^{\frac{1}{p}}\\
        \lesssim&\left(\mathbb{E}\int_{0}^{1}|g_n^{-1}(t)-g_\mu^{-1}(t)|dt\right)^{\frac{1}{p}}\\
        \lesssim& O(n^{-\frac{1}{2p}}).
    \end{aligned}
    \end{eqnarray*}
    The last equality is by the upper bound of 1-Wasserstein distance~\cite{panaretos2019statistical}.

    If $p< d$, we know that 
    \begin{align*}
        \mathbb{E}\overline{\text{HCP}}_p(\mu,\mu_n)
        \lesssim& \left(\mathbb{E}\int_{0}^{1}|g_n^{-1}(t)-g_\mu^{-1}(t)|^{\frac{p}{d}}dt\right)^{\frac{1}{p}}\\
        \lesssim&  \left(\mathbb{E}(\int_{0}^{1}|g_n^{-1}(t)-g_\mu^{-1}(t)|dt)^{\frac{p}{d}}\right)^{\frac{1}{p}}\\
        \lesssim&  \left(\mathbb{E}(\int_{0}^{1}|g_n^{-1}(t)-g_\mu^{-1}(t)|dt)\right)^{\frac{1}{d}}\\
        \lesssim& O(n^{-\frac{1}{2d}}).
    \end{align*}
    The second and third inequalities use Jensen's inequality and the last inequality is by the upper bound of 1-Wasserstein distance~\cite{panaretos2019statistical}.
\end{proof}

\begin{proof}[\textbf{Proof of Corollary~\ref{coro:conv1}}]
By definition, we know that 
$\overline{\text{HCP}}_p(\mu_n,\nu_n)=\left(\frac{1}{n}\sum_{i=1}^n \|x_{{(i)}^*}-y_{{(i)}^*})\|_p^p\right)^{\frac{1}{p}}$.
By triangle inequality, we have
\begin{eqnarray*}
\begin{aligned}
    &\overline{\text{HCP}}_p(\mu_n,\nu_n)- \text{HCP}_p(\mu, \nu) \geq -\overline{\text{HCP}}_p(\mu,\mu_n)-\overline{\text{HCP}}_p(\nu,\nu_n),\\
    &\overline{\text{HCP}}_p(\mu_n,\nu_n)- \text{HCP}_p(\mu, \nu)\leq\overline{\text{HCP}}_p(\mu,\mu_n)+\overline{\text{HCP}}_p(\nu,\nu_n).
\end{aligned}
\end{eqnarray*}
By Theorem~\ref{thm:conv1}, we can easily conclude the almost surely converge property.

By Theorem~\ref{thm:conv1}, we have  
$$\overline{\text{HCP}}_p(\mu,\mu_n)
\lesssim O(n^{-\frac{1}{2\max\{p,d\}}}),$$
and 
$$\overline{\text{HCP}}_p(\nu,\nu_n)
\lesssim O(n^{-\frac{1}{2\max\{p,d\}}}).$$

Thus, we have
$$\overline{\text{HCP}}_p(\mu_n,\nu_n)- \text{HCP}_p(\mu, \nu)\lesssim O(n^{-\frac{1}{2\max\{p,d\}}}),$$
and
$$\overline{\text{HCP}}_p(\mu_n,\nu_n)- \text{HCP}_p(\mu, \nu)\gtrsim -O(n^{-\frac{1}{2\max\{p,d\}}}).$$
Hence, we have finished the proof.
\end{proof}
Corollary~\ref{coro:conv1} can be easily extended to samples with different weights, which are omitted here.

\begin{proof}[\textbf{Proof of Corollary~\ref{rmk:coro1}}]
It can be similarly concluded like Corollary~\ref{coro:conv1}.    
\end{proof}

\begin{proof}[\textbf{Proof of Corollary~\ref{rmk:conv1}}]
By Section 3.3 in~\cite{panaretos2019statistical}, we know that $\mathbb{E}W_p(\mu_n,\mu)\geq O(n^{-\frac{1}{2p}})$.
From Theorem~\ref{thm:conv1}, we have proved it.    
\end{proof}

\begin{proof}[\textbf{Proof of Corollary~\ref{thm:conv2}}]
    Without loss of generality, we assume $s_1,s_2,\ldots,s_K$ are sorted in Hilbert order and $s_i\neq s_j$ for any $i\neq j$.
    Since there are finite supports, we easily know that $Pr(A^c)\leq\sum_{i=1}^KPr(\{x_j\neq s_i,\forall j \})= \sum_{i=1}^K(p_i)^n$ where $A=\{\widetilde{\Omega}_\mu=\widetilde{\Omega}_{\mu_n}\}$.
    Denote $\hat{p}_i=\frac{\sum_{j=1}^n\mathbf{1}_{\{s_i\}}(x_j)}{n}$ where $\mathbf{1}$ is the indicator function.
    Following the notation in Theorem~\ref{thm:conv1}, we have
    \begin{eqnarray*}
    \begin{aligned}
        \mathbb{E}\text{HCP}_p(\mu,\mu_n)
        =&\mathbb{E}[\text{HCP}_p(\mu,\mu_n)\mathbf{1}_A+\text{HCP}_p(\mu,\mu_n)\mathbf{1}_{A^c}]\\
        \lesssim&\left(\mathbb{E}\int_{0}^{1}|g_n^{-1}(t)-g_\mu^{-1}(t)|^{\frac{p}{d}}\mathrm{d}t\right)^{\frac{1}{p}} + Pr(A^c)\\
        \lesssim&\Biggl(\mathbb{E}\text{W}_1\Biggl(\sum_{i=1}^K
        p_i\delta_{\{i\}},\sum_{i=1}^K \hat{p}_i\delta_{\{i\}}\Biggr)\Biggr)^{\frac{1}{p}}+ \sum_{i=1}^K(p_i)^n\\
        \lesssim& O(n^{-1/2p}).
    \end{aligned}  
    \end{eqnarray*}
     The second inequality is directly from the discrete distribution and the last inequality is from~\cite{panaretos2019statistical}.

    We have proved the upper bound.
    The lower bound can be directly concluded by $W_p(\mu_n,\mu)\gtrsim n^{-1/2p}$; see~\cite{panaretos2019statistical}.
\end{proof}

\subsection{The proofs related to IPRHCP and PRHCP}

\begin{proof}[\textbf{Proof of Theorem~\ref{thm:IPRHCP}}]

By definition and Theorem~\ref{thm:HCP}
\begin{eqnarray*}
\begin{aligned}
 \text{IPRHCP}_{p,q}(\mu, \nu)&=\Bigl(\int_{\mathbb{S}_{d, q}} \text{HCP}_{p}^{p}\left({P_\E}_{\#} \mu, {P_\E}_{\#} \nu\right) \mathrm{d} \sigma(\E)\Bigr)^{1 / p}\\
 &\geq \Bigl(\int_{\mathbb{S}_{d, q}} {W}_{p}^{p}\left({P_\E}_{\#} \mu, {P_\E}_{\#} \nu\right) \mathrm{d} \sigma(\E)\Bigr)^{1 / p}\\
 &=\text{IPRW}_{p,q}(\mu,\nu).
\end{aligned}
\end{eqnarray*}
By definition of IPRHCP, it is clear that $\text{IPRHCP}_{p,q}(\mu,\nu)\geq 0$ and $\text{IPRHCP}_{p,q}(\mu,\nu)=\text{IPRHCP}_{p,q}(\nu,\mu)$.
Because $\text{IPRW}_{p,q}$ is a  well-defined metric, we can easily get $\text{IPRHCP}_{p,q}(\mu,\nu)=0$ if and only if $\mu=\nu$.
Next, we prove the triangle inequality as follows,
\begin{eqnarray*}
\begin{aligned}
 \text{IPRHCP}_{p,q}(\mu, \nu)
 =&\Bigl(\int_{\mathbb{S}_{d, q}} \text{HCP}_{p}^{p}\left({P_\E}_{\#} \mu, {P_\E}_{\#} \nu\right) \mathrm{d} \sigma(\E)\Bigr)^{\frac{1}{p}}\\
 \leq& \Bigl(\int_{\mathbb{S}_{d, q}} [\text{HCP}_{p}({P_\E}_{\#} \mu, {P_\E}_{\#} \tau)+\\
 &\text{HCP}_{p}({P_\E}_{\#} \nu, {P_\E}_{\#} \tau) ]^p \mathrm{d}\sigma(\E)\Bigr)^{\frac{1}{p}}\\
 \leq& \text{IPRHCP}_{p,q}(\mu,\tau)+\text{IPRHCP}_{p,q}(\nu,\tau),
\end{aligned}
\end{eqnarray*}
where the first inequality uses triangle inequality from Theorem~\ref{thm:HCP} and the second uses Minkowski inequality.
\end{proof}

\begin{proof}[\textbf{Proof of Theorem~\ref{thm:iprhcp_ineq}}]
First, we introduce an inequality between W and SW distance, which has been proved in~\cite{bonnotte2013unidimensional}.
When $c(x, y) = \|x-y\|_p^p$ for $p\geq 2$, the following upper bound hold for the SW distance:
\begin{eqnarray*}
\begin{aligned}
\text{SW}_{p}^{p}\left(\mu, \nu\right) \leq \alpha_{d, p} W_{p}^{p}\left(\mu, \nu\right),
\end{aligned}
\end{eqnarray*}
where $\alpha_{d, p}=\frac{1}{d} \int_{\mathbb{S}_{d,1}}\|\theta\|_{p}^{p} d \theta \leq 1$.

Thus, we have:
\begin{align*}
&\text{IPRHCP}_{p,q}^p(\mu, \nu)\\
=&\int_{\mathbb{S}_{d, q}} \text{HCP}_{p}^{p}\left({P_\E}_{\#} \mu, {P_\E}_{\#} \nu\right) \mathrm{d} \sigma(\E)\\
 \geq& \int_{\mathbb{S}_{d, q}} {W}_{p}^{p}\left({P_\E}_{\#} \mu, {P_\E}_{\#} \nu\right) \mathrm{d} \sigma(\E)\\
 \geq& \int_{\mathbb{S}_{d, q}} \frac{1}{\alpha_{q,p}}\text{SW}_{p}^{p}\left({P_\E}_{\#} \mu, {P_\E}_{\#} \nu\right) \mathrm{d} \sigma(\E)\\
 =& \frac{1}{\alpha_{q,p}}\int_{\mathbb{S}_{d, q}} \int_{\mathbb{S}_{q,1}}W_p^p
 \left({P_\mathbf{v}}_{\#}[{P_\E}_{\#} \mu], {P_\mathbf{v}}_{\#}[{P_\E}_{\#} \nu]\right)\mathrm{d}\sigma(\mathbf{v}) \mathrm{d} \sigma(\E)\\
 =&\frac{1}{\alpha_{q,p}} \int_{\mathbb{S}_{d,1}}W_p^p
 \left({P_\mathbf{v}}_{\#}\mu, {P_\mathbf{v}}_{\#}\nu\right)\mathrm{d}\sigma(\mathbf{v}) \\
 =&\frac{1}{\alpha_{q,p}}\text{SW}_p^p(\mu,\nu),
\end{align*}
where the first inequality uses Theorem~\ref{thm:HCP}, the second inequality uses the above inequality and the last several equalities use the property of SW.
\end{proof}

\begin{proof}[\textbf{Proof of Corollary~\ref{rmk:iprhcp_ineq}}]

Here, we take $q=2$ as an example.
It can be easily extended to the case $q>2$.
From the construction of 2-$d$ Hilbert Curve, we know that if $x_1<x_2$, then $\min\{H^{-1}((x_1,x_1))\}<\min\{H^{-1}((x_2,x_2))\}$.
By $\E^{\top}\E=\mathbf{J}_{2}$, we know $\E=(\mathbf{v},\mathbf{v})$ with $\|\mathbf{v}\|_2=1$. 
Hence, we can easily conclude that 
\begin{equation*}
\text{HCP}_{p}^{p}\left({P_\E}_{\#} \mu, {P_\E}_{\#} \nu\right)=2{W}_{p}^{p}\left({P_{\mathbf{v}}}_{\#} \mu, {P_\mathbf{v}}_{\#} \nu\right).
\end{equation*}
Thus,
\begin{align*}
\text{IPRHCP}_{p,q}(\mu, \nu)&=\Bigl(\int \text{HCP}_{p}^{p}\left({P_\E}_{\#} \mu, {P_\E}_{\#} \nu\right) \mathrm{d} \sigma(\E)\Bigr)^{1 / p}\\
&=\Bigl(\int_{\mathbb{S}_{d,1}} 2{W}_{p}^{p}\left({P_{\mathbf{v}}}_{\#} \mu, {P_\mathbf{v}}_{\#} \nu\right)\mathrm{d} \sigma(\mathbf{v})\Bigr)^{1 / p}\\
&=2^{1/p}\text{SW}_p(\mu,\nu).
\end{align*}
\end{proof}

\begin{proof}[\textbf{Proof of Theorem~\ref{thm:iprhcp}}]

From the proof of Theorem~\ref{thm:conv1}, we know that for any given subspace $\E$,
\begin{align*}
    &\mathbb{E}\frac{1}{n}\sum_{i=1}^n \int_{\frac{i-1}{n}}^{\frac{i}{n}}\|x_{\E,{(i)}^*}-H_{{P_\E}_{\#} \mu}(g_{{P_\E}_{\#} \mu}^{-1}(t))\|_p^p \mathrm{d}t\\
    \leq & C_1\mathbb{E}\int_{0}^{1}|g_{n,\E}^{-1}(t)-g_{{P_\E}_{\#} \mu}^{-1}(t)|^{\frac{p}{q}}\mathrm{d}t,\\
\end{align*}
where $g_{n,\E}^{-1}(t)$, $g_{{P_\E}_{\#} \mu}^{-1}(t)$ are defined similar to Theorem~\ref{thm:conv1} and where $C_1$ is a constant, not depending on $\E$.

Considering $\mu$ has bounded support, from~\cite{bobkov2019one}, we know that for any given $\E$,
\begin{align*}
    \mathbb{E}\left(\int_{0}^{1}|g_{n,\E}^{-1}(t)-g_{{P_\E}_{\#} \mu}^{-1}(t)|\mathrm{d}t\right)\leq \frac{C_2}{\sqrt{n}},
\end{align*}
where $C_2$ is a constant, not depending on $\E$.

When $p\geq q$, we have
\begin{align*}
    &\mathbb{E}\int_{0}^{1}|g_{n,\E}^{-1}(t)-g_{{P_\E}_{\#} \mu}^{-1}(t)|^{\frac{p}{q}}\mathrm{d}t\\
    \leq& \mathbb{E}\int_{0}^{1}|g_{n,\E}^{-1}(t)-g_{{P_\E}_{\#} \mu}^{-1}(t)|\mathrm{d}t \leq \frac{C_2}{\sqrt{n}}.
\end{align*}
When $p<q$, by Jenson's inequality,
\begin{align*}
    &\mathbb{E}\int_{0}^{1}|g_{n,\E}^{-1}(t)-g_{{P_\E}_{\#} \mu}^{-1}(t)|^{\frac{p}{q}}\mathrm{d}t\\
    \leq& \left(\mathbb{E}\int_{0}^{1}|g_{n,\E}^{-1}(t)-g_{{P_\E}_{\#} \mu}^{-1}(t)|\mathrm{d}t\right)^{\frac{p}{q}}
    \leq C_2^{p/q}n^{p/2q}.
\end{align*}

When $p\geq q$, we have
\begin{eqnarray*}
\begin{aligned}
    &\mathbb{E}\Bigl(\int_{\E\in\mathbb{S}_{d, q}} \frac{1}{n}\sum_{i=1}^n \int_{\frac{i-1}{n}}^{\frac{i}{n}}\|x_{\E,{(i)}^*}-H_{{P_\E}_{\#} \mu}(g_{{P_\E}_{\#} \mu}^{-1}(t))\|_p^p \mathrm{d}t \mathrm{d} \sigma(\E)\Bigr)^{\frac{1}{p}}\\
    \leq&\Bigl(\int_{\E\in\mathbb{S}_{d, q}}\mathbb{E} \frac{1}{n}\sum_{i=1}^n \int_{\frac{i-1}{n}}^{\frac{i}{n}}\|x_{\E,{(i)}^*}-H_{{P_\E}_{\#} \mu}(g_{{P_\E}_{\#} \mu}^{-1}(t))\|_p^p \mathrm{d}t \mathrm{d} \sigma(\E)\Bigr)^{\frac{1}{p}}\\
    \leq& Ln^{-\frac{1}{2p}}.
\end{aligned}   
\end{eqnarray*}
When $p< q$, we have
\begin{eqnarray*}
\begin{aligned}
&\mathbb{E}\Bigl(\int_{\E\in\mathbb{S}_{d, q}} \frac{1}{n}\sum_{i=1}^n \int_{\frac{i-1}{n}}^{\frac{i}{n}}\|x_{\E,{(i)}^*}-H_{{P_\E}_{\#} \mu}(g_{{P_\E}_{\#} \mu}^{-1}(t))\|_p^p \mathrm{d}t \mathrm{d} \sigma(\E)\Bigr)^{\frac{1}{p}}\\
&\leq Ln^{-\frac{1}{2q}},
\end{aligned} 
\end{eqnarray*}
where $L$ is a constant. 
Following the proof in Corollary~\ref{coro:conv1}, we have finished the proof. 
\end{proof}

\begin{proof}[\textbf{Proof of Theorem~\ref{theorem:PRHCP}}]
By the definition of PRHCP and Theorem~\ref{thm:HCP}, we have
\begin{align*}
 \text{PRHCP}_{p,q}(\mu, \nu)&=\sup_{\E\in \mathbb{S}_{d, q}} \text{HCP}_{p}\left({P_\E}_{\#} \mu, {P_\E}_{\#} \nu\right) \\
 &\geq \sup_{\E\in \mathbb{S}_{d, q}} W_{p}\left({P_\E}_{\#} \mu, {P_\E}_{\#} \nu\right) \\
 &=\text{PRW}_{p,q}(\mu,\nu).
\end{align*} 
By definition of PRHCP, it is clear that $\text{PRHCP}_{p,q}(\mu,\nu)\geq 0$ and $\text{PRHCP}_{p,q}(\mu,\nu)=\text{PRHCP}_{p,q}(\nu,\mu)$.
Because $\text{PRW}_{p,q}$ is a  well-defined metric, we can easily get $\text{PRHCP}_{p,q}(\mu,\nu)=0$ if and only if $\mu=\nu$.
Next, we prove the triangle inequality as follows,
\begin{align*}
 &\text{PRHCP}_{p,q}(\mu, \nu)\\
 =&\sup_{\E\in \mathbb{S}_{d, q}} \text{HCP}_{p}\left({P_\E}_{\#} \mu, {P_\E}_{\#} \nu\right) \\
 \leq& \sup_{\E\in \mathbb{S}_{d, q}} \text{HCP}_{p}\left({P_\E}_{\#} \mu, {P_\E}_{\#} \nu\right)+
  \text{HCP}_{p}\left({P_\E}_{\#} \nu, {P_\E}_{\#} \tau\right)\\
  \leq& \sup_{\E\in \mathbb{S}_{d, q}} \text{HCP}_{p}\left({P_\E}_{\#} \mu, {P_\E}_{\#} \nu\right)+
  \sup_{\E\in \mathbb{S}_{d, q}} \text{HCP}_{p}\left({P_\E}_{\#} \nu, {P_\E}_{\#} \tau\right)\\
 \leq& \text{PRHCP}_{p,q}(\mu,\tau)+\text{PRHCP}_{p,q}(\nu,\tau),
\end{align*}
where the first inequality uses triangle inequality from Theorem~\ref{thm:HCP} and the second uses Minkowski inequality.
\end{proof}

\subsection{Statistical convergence analysis for $k$-order approximation Hilbert curve}
The bound of convergence rate in our manuscript is regardless of $k$ since we used the Hilbert curve rather than the $k$-order approximation Hilbert curve in the statistical convergence analysis.
In practice, we always approximate the Hilbert curve using $k$-order Hilbert curve.
Therefore,the bound should depend on the parameter $k$ in practice.
We now provide a more precise analysis in the following.

As shown in Section 3.4, following the definition in \cite{he2016extensible}, the $k$-order Hilbert curves, i.e.,
$\widehat{H}_{k,\mu}$
partitions both $[0,1]$ and $\widetilde\Omega_\mu$ into $2^{kd}$ equal blocks, denoted by $\{c'_{j,\mu}\}_{j=1}^{2^{dk}}$ and $\{c_{j,\mu}\}_{j=1}^{2^{dk}}$, respectively, and construct a bijection between these blocks (this bijection is along Hilbert curve).
Following the strategy used in the manuscript, we study a modified empirical $k$-order Hilbert curve projection distance instead. 
Specifically, following the definitions of $g_\mu$, we first define the cumulative distribution function and its inverse for the empirical measure $\mu_n$ corresponding to the $k$-order Hilbert curve:
\begin{eqnarray*}\label{def:g}
\begin{aligned}
    &\hat{g}_{\mu_n,k}(t)=\sideset{}{_{s\in\mathcal{K}_k,~s\geq t}}\inf\mu_n\Bigl(\widehat{H}_{k,\mu}([0,s])\Bigr),\\
    &\hat{g}_{\mu_n,k}^{-1}(t)=\sideset{}{_{s\in[0,1],~\hat{g}_{\mu_n,k}(s)>t}}\inf s.
\end{aligned}
\end{eqnarray*}
where $\mathcal{K}_k=\{\frac{m_1}{2^{dk}}:m_1\in\mathbb{N},m_1\leq2^{dk}\}$.
Accordingly, we define the modified empirical $k$-order Hilbert curve projection distance as:
\begin{eqnarray*}\label{def:k}
\begin{aligned}
    \overline{\text{HCP}}_{p,k}(\mu, \mu_n) = \Bigl(\int_0^1{\|H_\mu(g_\mu^{-1}(t))-\widehat{H}_{k,\mu}'(\hat{g}_{\mu_n,k}^{-1}(t))\|}_p^p\mathrm{d}t\Bigr)^{\frac{1}{p}}.
\end{aligned}
\end{eqnarray*}
where $\widehat{H}_{k,\mu}'(x) $ maps $x$ in block $c'_{j,\mu}$ to the center of the block $c_{j,\mu}$.
The following theorem provides an upper bound for the convergence rate. 

\begin{mytheorem}\label{thm:conv11}
Let $\{x_i\}_{i=1}^n$ be an i.i.d. sample that is generated from the probability measure $\mu\in\mathscr{P}_\infty(\RR^d)$.
The empirical measure is defined by $\mu_n=\frac{1}{n}\sum_{i=1}^n\delta_{x_i}$.
Then, we have
\begin{eqnarray*}
\begin{aligned}
    \mathbb{E}\overline{\text{HCP}}_{p,k}(\mu, \mu_n)&\lesssim \left(\max_{0\leq i\leq 2^{dk}-1} \mu\left(H_\mu([\frac{i}{2^{dk}},\frac{i+1}{2^{dk}}])\right)\right)^{\frac{1}{\max\{d,p\}}}\\
    &+ O(n^{-\frac{1}{2\max\{p,d\}}})
     + \frac{d^{1/p}}{2^k}.
\end{aligned}
\end{eqnarray*}
\end{mytheorem}

\begin{proof}
    We know that
\begin{align*}
    \overline{\text{HCP}}_{p,k}(\mu, \mu_n) &= \Bigl(\int_0^1{\|H_\mu(g_\mu^{-1}(t))-\widehat{H}_{k,\mu}'(\hat{g}_{\mu_n,k}^{-1}(t))\|}_p^p\mathrm{d}t\Bigr)^{\frac{1}{p}}\\
    & \leq \Bigl(\int_0^1{\|H_\mu(g_\mu^{-1}(t))-H_{\mu}(\hat{g}_{\mu_n,k}^{-1}(t))\|}_p^p\mathrm{d}t\Bigr)^{\frac{1}{p}}\\
    & + \Bigl(\int_0^1{\|H_{\mu}(\hat{g}_{\mu_n,k}^{-1}(t))-\widehat{H}_{k,\mu}'(\hat{g}_{\mu_n,k}^{-1}(t))\|}_p^p\mathrm{d}t\Bigr)^{\frac{1}{p}}\\
    & \lesssim \Bigl(\int_0^1{|g_\mu^{-1}(t)-\hat{g}_{\mu_n,k}^{-1}(t)|}^{\frac{p}{d}}\mathrm{d}t\Bigr)^{\frac{1}{p}}\\
    & + \Bigl(\int_0^1{\|H_{\mu}(\hat{g}_{\mu_n,k}^{-1}(t))-\widehat{H}_{k,\mu}'(\hat{g}_{\mu_n,k}^{-1}(t))\|}_p^p\mathrm{d}t\Bigr)^{\frac{1}{p}}
\end{align*}
where the first inequality uses Minkowski inequality and triangle inequality (similar to the proof of Theorem 1) and the second inequality uses locality-preserving property.
Since $H_{\mu}(t)$ and $\widehat{H}_{k,\mu}'(t)$ lie in the same block, it is easy to show that $\|H_{\mu}(t)-\widehat{H}_{k,\mu}'(t)\|_p^p\leq \frac{d}{2^{pk}}$ for any given $t\in[0,1]$. 

We have when $p\geq d$
    \begin{align*}
        \left(\int_{0}^{1}|g_\mu^{-1}(t)-\hat{g}_{\mu_n,k}^{-1}(t)|^{\frac{p}{d}}\mathrm{d}t\right)^{\frac{1}{p}}
        & \leq \left(\int_{0}^{1}|g_\mu^{-1}(t)-\hat{g}_{\mu_n,k}^{-1}(t)|\mathrm{d}t\right)^{\frac{1}{p}}.
    \end{align*}
    When $p<d$, by Jenson's inequality, we have 
    \begin{align*}
        \left(\int_{0}^{1}|g_\mu^{-1}(t)-\hat{g}_{\mu_n,k}^{-1}(t)|^{\frac{p}{d}}\mathrm{d}t\right)^{\frac{1}{p}}
        & \leq \left(\int_{0}^{1}|g_\mu^{-1}(t)-\hat{g}_{\mu_n,k}^{-1}(t)|\mathrm{d}t\right)^{\frac{1}{d}}.
    \end{align*} 
And we have $|g_\mu^{-1}(t)-\hat{g}_{\mu_n,k}^{-1}(t)|\leq |g_\mu^{-1}(t)-g_{\mu,k}^{-1}(t)|+|g_{\mu,k}^{-1}(t)-\hat{g}_{\mu_n,k}^{-1}(t)|$, 
where 
\begin{eqnarray*}
\begin{aligned}
    &g_{\mu,k}(t)=\sideset{}{_{s\in\mathcal{K}_k,~s\geq t}}\inf\mu\Bigl(\widehat{H}_{k,\mu}([0,s])\Bigr),\\
    &g_{\mu,k}^{-1}(t)=\sideset{}{_{s\in[0,1],~g_{\mu,k}(s)>t}}\inf s.
\end{aligned}
\end{eqnarray*}
Thus, we know that
\begin{align*}
    &\mathbb{E}\overline{\text{HCP}}_{p,k}(\mu, \mu_n)\\
    \lesssim& \left(\int_{0}^{1}|g_\mu^{-1}(t)-g_{\mu,k}^{-1}(t)|\mathrm{d}t\right)^{\frac{1}{\max\{d,p\}}}\\
     +& \left(\mathbb{E}\int_{0}^{1}|g_{\mu,k}^{-1}(t)-\hat{g}_{\mu_n,k}^{-1}(t)|\mathrm{d}t\right)^{\frac{1}{\max\{d,p\}}} + \frac{d^{1/p}}{2^k}\\
     \lesssim & \left(\int_{0}^{1}|g_\mu^{-1}(t)-g_{\mu,k}^{-1}(t)|\mathrm{d}t\right)^{\frac{1}{\max\{d,p\}}} + O(n^{-\frac{1}{2\max\{p,d\}}})
     + \frac{d^{1/p}}{2^k}\\
      = &\left(\int_{0}^{1}|g_\mu(t)-g_{\mu,k}(t)|\mathrm{d}t\right)^{\frac{1}{\max\{d,p\}}} + O(n^{-\frac{1}{2\max\{p,d\}}}) + \frac{d^{1/p}}{2^k}\\
     \leq & \left(\max_{i} \mu\left(H_\mu([\frac{i}{2^{dk}},\frac{i+1}{2^{dk}}])\right)\right)^{\frac{1}{\max\{d,p\}}} + O(n^{-\frac{1}{2\max\{p,d\}}})
     + \frac{d^{1/p}}{2^k}
\end{align*}
where the second inequality is from \cite{panaretos2019statistical} (similar
to the proof of Theorem 3), the third equality is from \cite{peyre2019computational} and the last inequality is from the fact that $|g_\mu(t)-g_{\mu,k}(t)|\leq\max_{i} \mu\left(H_\mu([\frac{i}{2^{dk}},\frac{i+1}{2^{dk}}])\right)$.
\end{proof}
\begin{myremark}
    If $\mu$ is absolutely continuous with bounded density function $f$, then we have 
    \begin{eqnarray*}
    \begin{aligned}
    \mathbb{E}\overline{\text{HCP}}_{p,k}(\mu, \mu_n)\lesssim \left(\frac{C}{2^{dk}}\right)^{\frac{1}{\max\{d,p\}}} + O(n^{-\frac{1}{2\max\{p,d\}}})
     + \frac{d^{1/p}}{2^k}.
\end{aligned}
\end{eqnarray*}
for some constant $C>0$.
\end{myremark}
The above theoretical analysis shows that we could take $k=O(\log(n))$ in practice.



\ifCLASSOPTIONcaptionsoff
  \newpage
\fi

\bibliographystyle{unsrt}
\bibliography{main}

\begin{thebibliography}{10}

\bibitem{kusner2015word}
Matt Kusner, Yu~Sun, Nicholas Kolkin, and Kilian Weinberger.
\newblock From word embeddings to document distances.
\newblock In {\em International conference on machine learning}, pages 957--966. PMLR, 2015.

\bibitem{huang2016supervised}
Gao Huang, Chuan Guo, Matt~J Kusner, Yu~Sun, Fei Sha, and Kilian~Q Weinberger.
\newblock Supervised word mover's distance.
\newblock {\em Advances in neural information processing systems}, 29, 2016.

\bibitem{rakotomamonjy2018distance}
Alain Rakotomamonjy, Abraham Traor{\'e}, Maxime Berar, R{\'e}mi Flamary, and Nicolas Courty.
\newblock Distance measure machines.
\newblock {\em arXiv preprint arXiv:1803.00250}, 2018.

\bibitem{goodfellow2014generative}
Ian Goodfellow, Jean Pouget-Abadie, Mehdi Mirza, Bing Xu, David Warde-Farley, Sherjil Ozair, Aaron Courville, and Yoshua Bengio.
\newblock Generative adversarial nets.
\newblock In {\em Advances in neural information processing systems}, pages 2672--2680, 2014.

\bibitem{kingma2013auto}
Diederik~P Kingma and Max Welling.
\newblock Auto-encoding variational bayes.
\newblock {\em International Conference on Learning Representations}, 2014.

\bibitem{arjovsky2017wasserstein}
Martin Arjovsky, Soumith Chintala, and L{\'e}on Bottou.
\newblock {W}asserstein generative adversarial networks.
\newblock In {\em International Conference on Machine Learning}, pages 214--223, 2017.

\bibitem{gretton2012kernel}
Arthur Gretton, Karsten~M Borgwardt, Malte~J Rasch, Bernhard Sch{\"o}lkopf, and Alexander Smola.
\newblock A kernel two-sample test.
\newblock {\em The Journal of Machine Learning Research}, 13(1):723--773, 2012.

\bibitem{villani2009optimal}
C{\'e}dric Villani.
\newblock {\em Optimal transport: old and new}, volume 338.
\newblock Springer, 2009.

\bibitem{tolstikhin2018wasserstein}
I~Tolstikhin, O~Bousquet, S~Gelly, and B~Sch{\"o}lkopf.
\newblock {W}asserstein auto-encoders.
\newblock In {\em 6th International Conference on Learning Representations}, 2018.

\bibitem{brenier1997homogenized}
Yann Brenier.
\newblock A homogenized model for vortex sheets.
\newblock {\em Archive for Rational Mechanics and Analysis}, 138(4):319--353, 1997.

\bibitem{benamou2002monge}
J-D Benamou, Yann Brenier, and Kevin Guittet.
\newblock The {M}onge--{K}antorovitch mass transfer and its computational fluid mechanics formulation.
\newblock {\em International Journal for Numerical methods in fluids}, 40(1-2):21--30, 2002.

\bibitem{rubner1997earth}
Yossi Rubner, Leonidas~J Guibas, and Carlo Tomasi.
\newblock The earth mover’s distance, multi-dimensional scaling, and color-based image retrieval.
\newblock In {\em Proceedings of the ARPA image understanding workshop}, volume 661, page 668, 1997.

\bibitem{pele2009fast}
Ofir Pele and Michael Werman.
\newblock Fast and robust earth mover's distances.
\newblock In {\em 2009 IEEE 12th International Conference on Computer Vision}, pages 460--467. IEEE, 2009.

\bibitem{cuturi2013sinkhorn}
Marco Cuturi.
\newblock Sinkhorn distances: Lightspeed computation of optimal transport.
\newblock {\em Advances in Neural Information Processing Systems}, 26:2292--2300, 2013.

\bibitem{bonneel2015sliced}
Nicolas Bonneel, Julien Rabin, Gabriel Peyr{\'e}, and Hanspeter Pfister.
\newblock Sliced and {R}adon {W}asserstein barycenters of measures.
\newblock {\em Journal of Mathematical Imaging and Vision}, 51(1):22--45, 2015.

\bibitem{kolouri2019generalized}
Soheil Kolouri, Kimia Nadjahi, Umut Simsekli, Roland Badeau, and K~Gustavo.
\newblock Generalized sliced {W}asserstein distances.
\newblock {\em Advances in Neural Information Processing Systems}, 2019.

\bibitem{le2019tree}
Tam Le, Makoto Yamada, Kenji Fukumizu, and Marco Cuturi.
\newblock Tree-sliced variants of {W}asserstein distances.
\newblock In {\em Advances in Neural Information Processing Systems}, 2019.

\bibitem{bader2012space}
Michael Bader.
\newblock {\em Space-filling curves: an introduction with applications in scientific computing}.
\newblock Springer, 2012.

\bibitem{abel1990comparative}
David~J Abel and David~M Mark.
\newblock A comparative analysis of some two-dimensional orderings.
\newblock {\em International Journal of Geographical Information Systems}, 4(1):21--31, 1990.

\bibitem{moon2001analysis}
Bongki Moon, Hosagrahar~V Jagadish, Christos Faloutsos, and Joel~H. Saltz.
\newblock Analysis of the clustering properties of the {H}ilbert space-filling curve.
\newblock {\em IEEE Transactions on knowledge and data engineering}, 13(1):124--141, 2001.

\bibitem{deshpande2019max}
Ishan Deshpande, Yuan-Ting Hu, Ruoyu Sun, Ayis Pyrros, Nasir Siddiqui, Sanmi Koyejo, Zhizhen Zhao, David Forsyth, and Alexander~G Schwing.
\newblock Max-sliced {W}asserstein distance and its use for {GAN}s.
\newblock In {\em Proceedings of the IEEE/CVF Conference on Computer Vision and Pattern Recognition}, pages 10648--10656, 2019.

\bibitem{kolouri2018sliced}
Soheil Kolouri, Phillip~E Pope, Charles~E Martin, and Gustavo~K Rohde.
\newblock Sliced {W}asserstein auto-encoders.
\newblock In {\em International Conference on Learning Representations}, 2018.

\bibitem{altschuler2017near}
Jason Altschuler, Jonathan Niles-Weed, and Philippe Rigollet.
\newblock Near-linear time approximation algorithms for optimal transport via {S}inkhorn iteration.
\newblock {\em Advances in neural information processing systems}, 30, 2017.

\bibitem{altschuler2019massively}
Jason Altschuler, Francis Bach, Alessandro Rudi, and Jonathan Niles-Weed.
\newblock Massively scalable {S}inkhorn distances via the {N}ystr{\"o}m method.
\newblock {\em Advances in neural information processing systems}, 32, 2019.

\bibitem{dvurechensky2017adaptive}
Pavel Dvurechensky, Alexander Gasnikov, Sergey Omelchenko, and Alexander Tiurin.
\newblock Adaptive similar triangles method: {A} stable alternative to {S}inkhorn's algorithm for regularized optimal transport.
\newblock {\em arXiv preprint arXiv:1706.07622}, 2017.

\bibitem{thibault2021overrelaxed}
Alexis Thibault, L{\'e}na{\"\i}c Chizat, Charles Dossal, and Nicolas Papadakis.
\newblock Overrelaxed {S}inkhorn--{K}nopp algorithm for regularized optimal transport.
\newblock {\em Algorithms}, 14(5):143, 2021.

\bibitem{li2023importance}
Mengyu Li, Jun Yu, Tao Li, and Cheng Meng.
\newblock Importance sparsification for {S}inkhorn algorithm.
\newblock {\em Journal of Machine Learning Research}, 24:1--44, 2023.

\bibitem{liao2022fast}
Qichen Liao, Jing Chen, Zihao Wang, Bo~Bai, Shi Jin, and Hao Wu.
\newblock Fast {S}inkhorn {I}: An ${O(N)}$ algorithm for the {W}asserstein-1 metric.
\newblock {\em Communications in Mathematical Sciences}, 20(7):2053--2067, 2022.

\bibitem{guo2020fast}
Wenshuo Guo, Nhat Ho, and Michael Jordan.
\newblock Fast algorithms for computational optimal transport and {W}asserstein barycenter.
\newblock In {\em International Conference on Artificial Intelligence and Statistics}, pages 2088--2097. PMLR, 2020.

\bibitem{dvurechensky2018computational}
Pavel Dvurechensky, Alexander Gasnikov, and Alexey Kroshnin.
\newblock Computational optimal transport: {C}omplexity by accelerated gradient descent is better than by {S}inkhorn’s algorithm.
\newblock In {\em International conference on machine learning}, pages 1367--1376. PMLR, 2018.

\bibitem{genevay2016stochastic}
Aude Genevay, Marco Cuturi, Gabriel Peyr{\'e}, and Francis Bach.
\newblock Stochastic optimization for large-scale optimal transport.
\newblock {\em Advances in neural information processing systems}, 29, 2016.

\bibitem{xie2020fast}
Yujia Xie, Xiangfeng Wang, Ruijia Wang, and Hongyuan Zha.
\newblock A fast proximal point method for computing exact {W}asserstein distance.
\newblock In {\em Uncertainty in artificial intelligence}, pages 433--453. PMLR, 2020.

\bibitem{wang2014bregman}
Huahua Wang and Arindam Banerjee.
\newblock Bregman alternating direction method of multipliers.
\newblock {\em Advances in Neural Information Processing Systems}, 27, 2014.

\bibitem{ye2017fast}
Jianbo Ye, Panruo Wu, James~Z Wang, and Jia Li.
\newblock Fast discrete distribution clustering using {W}asserstein barycenter with sparse support.
\newblock {\em IEEE Transactions on Signal Processing}, 65(9):2317--2332, 2017.

\bibitem{xu2022representing}
Hongteng Xu, Jiachang Liu, Dixin Luo, and Lawrence Carin.
\newblock Representing graphs via {G}romov-{W}asserstein factorization.
\newblock {\em IEEE Transactions on Pattern Analysis and Machine Intelligence}, 2022.

\bibitem{nguyen2020distributional}
Khai Nguyen, Nhat Ho, Tung Pham, and Hung Bui.
\newblock Distributional sliced-{W}asserstein and applications to generative modeling.
\newblock In {\em International Conference on Learning Representations}, 2020.

\bibitem{rowland2019orthogonal}
Mark Rowland, Jiri Hron, Yunhao Tang, Krzysztof Choromanski, Tamas Sarlos, and Adrian Weller.
\newblock Orthogonal estimation of {W}asserstein distances.
\newblock In {\em The 22nd International Conference on Artificial Intelligence and Statistics}, pages 186--195. PMLR, 2019.

\bibitem{paty2019subspace}
Fran{\c{c}}ois-Pierre Paty and Marco Cuturi.
\newblock Subspace robust {W}asserstein distances.
\newblock In {\em International conference on machine learning}, pages 5072--5081. PMLR, 2019.

\bibitem{lin2020projection}
Tianyi Lin, Chenyou Fan, Nhat Ho, Marco Cuturi, and Michael Jordan.
\newblock Projection robust {W}asserstein distance and {R}iemannian optimization.
\newblock {\em Advances in neural information processing systems}, 33:9383--9397, 2020.

\bibitem{lin2021projection}
Tianyi Lin, Zeyu Zheng, Elynn Chen, Marco Cuturi, and Michael~I Jordan.
\newblock On projection robust optimal transport: Sample complexity and model misspecification.
\newblock In {\em International Conference on Artificial Intelligence and Statistics}, pages 262--270. PMLR, 2021.

\bibitem{nguyen2022revisiting}
Khai Nguyen and Nhat Ho.
\newblock Revisiting sliced {W}asserstein on images: From vectorization to convolution.
\newblock {\em arXiv preprint arXiv:2204.01188}, 2022.

\bibitem{nguyen2022amortized}
Khai Nguyen and Nhat Ho.
\newblock Amortized projection optimization for sliced {W}asserstein generative models.
\newblock {\em arXiv preprint arXiv:2203.13417}, 2022.

\bibitem{genevay2018learning}
Aude Genevay, Gabriel Peyr{\'e}, and Marco Cuturi.
\newblock Learning generative models with {S}inkhorn divergences.
\newblock In {\em International Conference on Artificial Intelligence and Statistics}, pages 1608--1617. PMLR, 2018.

\bibitem{wu2019sliced}
Jiqing Wu, Zhiwu Huang, Dinesh Acharya, Wen Li, Janine Thoma, Danda~Pani Paudel, and Luc~Van Gool.
\newblock Sliced {W}asserstein generative models.
\newblock In {\em Proceedings of the IEEE/CVF Conference on Computer Vision and Pattern Recognition}, pages 3713--3722, 2019.

\bibitem{xu2020learning}
Hongteng Xu, Dixin Luo, Ricardo Henao, Svati Shah, and Lawrence Carin.
\newblock Learning autoencoders with relational regularization.
\newblock In {\em International Conference on Machine Learning}, pages 10576--10586. PMLR, 2020.

\bibitem{frogner2015learning}
Charlie Frogner, Chiyuan Zhang, Hossein Mobahi, Mauricio Araya, and Tomaso~A Poggio.
\newblock Learning with a {W}asserstein loss.
\newblock {\em Advances in neural information processing systems}, 28, 2015.

\bibitem{togninalli2019wasserstein}
Matteo Togninalli, Elisabetta Ghisu, Felipe Llinares-L{\'o}pez, Bastian Rieck, and Karsten Borgwardt.
\newblock {W}asserstein {W}eisfeiler-{L}ehman graph kernels.
\newblock {\em Advances in Neural Information Processing Systems}, 32, 2019.

\bibitem{li2023efficient}
Mengyu Li, Jun Yu, Hongteng Xu, and Cheng Meng.
\newblock Efficient approximation of {G}romov-{W}asserstein distance using importance sparsification.
\newblock {\em Journal of Computational and Graphical Statistics}, pages 1--12, 2023.

\bibitem{meng2020sufficient}
Cheng Meng, Jun Yu, Jingyi Zhang, Ping Ma, and Wenxuan Zhong.
\newblock Sufficient dimension reduction for classification using principal optimal transport direction.
\newblock {\em Advances in Neural Information Processing Systems}, 33:4015--4028, 2020.

\bibitem{flamary2018wasserstein}
R{\'e}mi Flamary, Marco Cuturi, Nicolas Courty, and Alain Rakotomamonjy.
\newblock {W}asserstein discriminant analysis.
\newblock {\em Machine Learning}, 107(12):1923--1945, 2018.

\bibitem{courty2017joint}
Nicolas Courty, R{\'e}mi Flamary, Amaury Habrard, and Alain Rakotomamonjy.
\newblock Joint distribution optimal transportation for domain adaptation.
\newblock {\em Advances in Neural Information Processing Systems}, 30, 2017.

\bibitem{xu2019gromov}
Hongteng Xu, Dixin Luo, Hongyuan Zha, and Lawrence~Carin Duke.
\newblock Gromov-{W}asserstein learning for graph matching and node embedding.
\newblock In {\em International conference on machine learning}, pages 6932--6941. PMLR, 2019.

\bibitem{xu2019scalable}
Hongteng Xu, Dixin Luo, and Lawrence Carin.
\newblock Scalable {G}romov-{W}asserstein learning for graph partitioning and matching.
\newblock {\em Advances in neural information processing systems}, 32, 2019.

\bibitem{rabin2014adaptive}
Julien Rabin, Sira Ferradans, and Nicolas Papadakis.
\newblock Adaptive color transfer with relaxed optimal transport.
\newblock In {\em 2014 IEEE international conference on image processing (ICIP)}, pages 4852--4856. IEEE, 2014.

\bibitem{meng2019large}
Cheng Meng, Yuan Ke, Jingyi Zhang, Mengrui Zhang, Wenxuan Zhong, and Ping Ma.
\newblock Large-scale optimal transport map estimation using projection pursuit.
\newblock {\em Advances in Neural Information Processing Systems}, 32, 2019.

\bibitem{yurochkin2019hierarchical}
Mikhail Yurochkin, Sebastian Claici, Edward Chien, Farzaneh Mirzazadeh, and Justin~M Solomon.
\newblock Hierarchical optimal transport for document representation.
\newblock {\em Advances in Neural Information Processing Systems}, 32, 2019.

\bibitem{he2016extensible}
Zhijian He and Art~B Owen.
\newblock Extensible grids: uniform sampling on a space filling curve.
\newblock {\em Journal of the Royal Statistical Society: Series B (Statistical Methodology)}, 78(4):917--931, 2016.

\bibitem{zumbusch2012parallel}
Gerhard Zumbusch.
\newblock {\em Parallel multilevel methods: adaptive mesh refinement and loadbalancing}.
\newblock Springer Science \& Business Media, 2012.

\bibitem{fatras2020learning}
Kilian Fatras, Younes Zine, R{\'e}mi Flamary, Remi Gribonval, and Nicolas Courty.
\newblock Learning with minibatch {W}asserstein: asymptotic and gradient properties.
\newblock In {\em International Conference on Artificial Intelligence and Statistics}, pages 2131--2141. PMLR, 2020.

\bibitem{piccoli2014generalized}
Benedetto Piccoli and Francesco Rossi.
\newblock Generalized {W}asserstein distance and its application to transport equations with source.
\newblock {\em Archive for Rational Mechanics and Analysis}, 211:335--358, 2014.

\bibitem{weed2019sharp}
Jonathan Weed and Francis Bach.
\newblock Sharp asymptotic and finite-sample rates of convergence of empirical measures in {W}asserstein distance.
\newblock {\em Bernoulli}, 25(4A):2620--2648, 2019.

\bibitem{xi2022distributional}
Jiaqi Xi and Jonathan Niles-Weed.
\newblock Distributional convergence of the sliced {W}asserstein process.
\newblock {\em Advances in Neural Information Processing Systems}, 35:13961--13973, 2022.

\bibitem{panaretos2019statistical}
Victor~M Panaretos and Yoav Zemel.
\newblock Statistical aspects of {W}asserstein distances.
\newblock {\em Annual review of statistics and its application}, 6:405--431, 2019.

\bibitem{peyre2019computational}
Gabriel Peyr{\'e} and Marco Cuturi.
\newblock Computational optimal transport: With applications to data science.
\newblock {\em Foundations and Trends{\textregistered} in Machine Learning}, 11(5-6):355--607, 2019.

\bibitem{butz1969convergence}
Arthur~R Butz.
\newblock Convergence with {H}ilbert's space-filling curve.
\newblock {\em Journal of Computer and System Sciences}, 3(2):128--146, 1969.

\bibitem{butz1971alternative}
Arthur~R Butz.
\newblock Alternative algorithm for {H}ilbert's space-filling curve.
\newblock {\em IEEE Transactions on Computers}, 100(4):424--426, 1971.

\bibitem{skilling2004programming}
John Skilling.
\newblock Programming the {H}ilbert curve.
\newblock In {\em AIP Conference Proceedings}, volume 707, pages 381--387. American Institute of Physics, 2004.

\bibitem{tanaka2001study}
A~Tanaka.
\newblock {\em Study on a fast ordering of high dimensional data to spatial index}.
\newblock PhD thesis, Master thesis, Kyushu Institute of Technology, 2001.

\bibitem{imamura2016fast}
Yasunobu Imamura, Takeshi Shinohara, Kouichi Hirata, and Tetsuji Kuboyama.
\newblock Fast {H}ilbert sort algorithm without using {H}ilbert indices.
\newblock In {\em International Conference on Similarity Search and Applications}, pages 259--267. Springer, 2016.

\bibitem{fabri2009cgal}
Andreas Fabri and Sylvain Pion.
\newblock {CGAL}: The computational geometry algorithms library.
\newblock In {\em Proceedings of the 17th ACM SIGSPATIAL international conference on advances in geographic information systems}, pages 538--539, 2009.

\bibitem{hamilton2008compact}
Chris~H Hamilton and Andrew Rau-Chaplin.
\newblock Compact {H}ilbert indices: Space-filling curves for domains with unequal side lengths.
\newblock {\em Information Processing Letters}, 105(5):155--163, 2008.

\bibitem{bernton2019approximate}
Espen Bernton, Pierre~E Jacob, Mathieu Gerber, and Christian~P Robert.
\newblock Approximate {B}ayesian computation with the {W}asserstein distance.
\newblock {\em Journal of the Royal Statistical Society: Series B (Statistical Methodology)}, 81(2):235--269, 2019.

\bibitem{wu20153d}
Zhirong Wu, Shuran Song, Aditya Khosla, Fisher Yu, Linguang Zhang, Xiaoou Tang, and Jianxiong Xiao.
\newblock 3d shapenets: A deep representation for volumetric shapes.
\newblock In {\em Proceedings of the IEEE conference on computer vision and pattern recognition}, pages 1912--1920, 2015.

\bibitem{qi2017pointnet}
Charles~R Qi, Hao Su, Kaichun Mo, and Leonidas~J Guibas.
\newblock Pointnet: Deep learning on point sets for 3d classification and segmentation.
\newblock In {\em Proceedings of the IEEE conference on computer vision and pattern recognition}, pages 652--660, 2017.

\bibitem{mikolov2013distributed}
Tomas Mikolov, Ilya Sutskever, Kai Chen, Greg~S Corrado, and Jeff Dean.
\newblock Distributed representations of words and phrases and their compositionality.
\newblock {\em Advances in neural information processing systems}, 26, 2013.

\bibitem{lecun1998gradient}
Yann LeCun, L{\'e}on Bottou, Yoshua Bengio, and Patrick Haffner.
\newblock Gradient-based learning applied to document recognition.
\newblock {\em Proceedings of the IEEE}, 86(11):2278--2324, 1998.

\bibitem{radford2015unsupervised}
Alec Radford, Luke Metz, and Soumith Chintala.
\newblock Unsupervised representation learning with deep convolutional generative adversarial networks.
\newblock {\em International Conference on Learning Representations}, 2016.

\bibitem{kingma2014adam}
Diederik~P Kingma and Jimmy Ba.
\newblock Adam: A method for stochastic optimization.
\newblock {\em International Conference on Learning Representations}, 2015.

\bibitem{heusel2017gans}
Martin Heusel, Hubert Ramsauer, Thomas Unterthiner, Bernhard Nessler, and Sepp Hochreiter.
\newblock {GAN}s trained by a two time-scale update rule converge to a local {N}ash equilibrium.
\newblock {\em Advances in neural information processing systems}, 30, 2017.

\bibitem{balaji2020robust}
Yogesh Balaji, Rama Chellappa, and Soheil Feizi.
\newblock Robust optimal transport with applications in generative modeling and domain adaptation.
\newblock {\em Advances in Neural Information Processing Systems}, 33:12934--12944, 2020.

\bibitem{mukherjee2021outlier}
Debarghya Mukherjee, Aritra Guha, Justin~M Solomon, Yuekai Sun, and Mikhail Yurochkin.
\newblock Outlier-robust optimal transport.
\newblock In {\em International Conference on Machine Learning}, pages 7850--7860. PMLR, 2021.

\bibitem{le2021robust}
Khang Le, Huy Nguyen, Quang~M Nguyen, Tung Pham, Hung Bui, and Nhat Ho.
\newblock On robust optimal transport: Computational complexity and barycenter computation.
\newblock {\em Advances in Neural Information Processing Systems}, 34:21947--21959, 2021.

\bibitem{pham2020unbalanced}
Khiem Pham, Khang Le, Nhat Ho, Tung Pham, and Hung Bui.
\newblock On unbalanced optimal transport: An analysis of {S}inkhorn algorithm.
\newblock In {\em International Conference on Machine Learning}, pages 7673--7682. PMLR, 2020.

\bibitem{chizat2018scaling}
Lenaic Chizat, Gabriel Peyr{\'e}, Bernhard Schmitzer, and Fran{\c{c}}ois-Xavier Vialard.
\newblock Scaling algorithms for unbalanced optimal transport problems.
\newblock {\em Mathematics of Computation}, 87(314):2563--2609, 2018.

\bibitem{sejourne2023unbalanced}
Thibault S{\'e}journ{\'e}, Cl{\'e}ment Bonet, Kilian Fatras, Kimia Nadjahi, and Nicolas Courty.
\newblock Unbalanced optimal transport meets sliced-{W}asserstein.
\newblock {\em arXiv preprint arXiv:2306.07176}, 2023.

\bibitem{memoli2011gromov}
Facundo M{\'e}moli.
\newblock Gromov--{W}asserstein distances and the metric approach to object matching.
\newblock {\em Foundations of computational mathematics}, 11(4):417--487, 2011.

\bibitem{pass2015multi}
Brendan Pass.
\newblock Multi-marginal optimal transport: theory and applications.
\newblock {\em ESAIM: Mathematical Modelling and Numerical Analysis-Mod{\'e}lisation Math{\'e}matique et Analyse Num{\'e}rique}, 49(6):1771--1790, 2015.

\bibitem{haasler2021multi}
Isabel Haasler, Rahul Singh, Qinsheng Zhang, Johan Karlsson, and Yongxin Chen.
\newblock Multi-marginal optimal transport and probabilistic graphical models.
\newblock {\em IEEE Transactions on Information Theory}, 67(7):4647--4668, 2021.

\bibitem{agueh2011barycenters}
Martial Agueh and Guillaume Carlier.
\newblock Barycenters in the {W}asserstein space.
\newblock {\em SIAM Journal on Mathematical Analysis}, 43(2):904--924, 2011.

\bibitem{peyre2016gromov}
Gabriel Peyr{\'e}, Marco Cuturi, and Justin Solomon.
\newblock Gromov-{W}asserstein averaging of kernel and distance matrices.
\newblock In {\em International Conference on Machine Learning}, pages 2664--2672. PMLR, 2016.

\bibitem{luitjens2007parallel}
Justin Luitjens, Martin Berzins, and Tom Henderson.
\newblock Parallel space-filling curve generation through sorting.
\newblock {\em Concurrency and Computation: Practice and Experience}, 19(10):1387--1402, 2007.

\bibitem{kamel1993hilbert}
Ibrahim Kamel and Christos Faloutsos.
\newblock Hilbert {R}-tree: An improved {R}-tree using fractals.
\newblock Technical report, 1993.

\bibitem{bonnotte2013unidimensional}
Nicolas Bonnotte.
\newblock {\em Unidimensional and evolution methods for optimal transportation}.
\newblock PhD thesis, Paris 11, 2013.

\bibitem{bobkov2019one}
Sergey Bobkov and Michel Ledoux.
\newblock {\em One-dimensional empirical measures, order statistics, and Kantorovich transport distances}, volume 261.
\newblock American Mathematical Society, 2019.

\end{thebibliography}

\begin{IEEEbiography}
[{\includegraphics[width=1in,height=1.25in,clip,keepaspectratio]{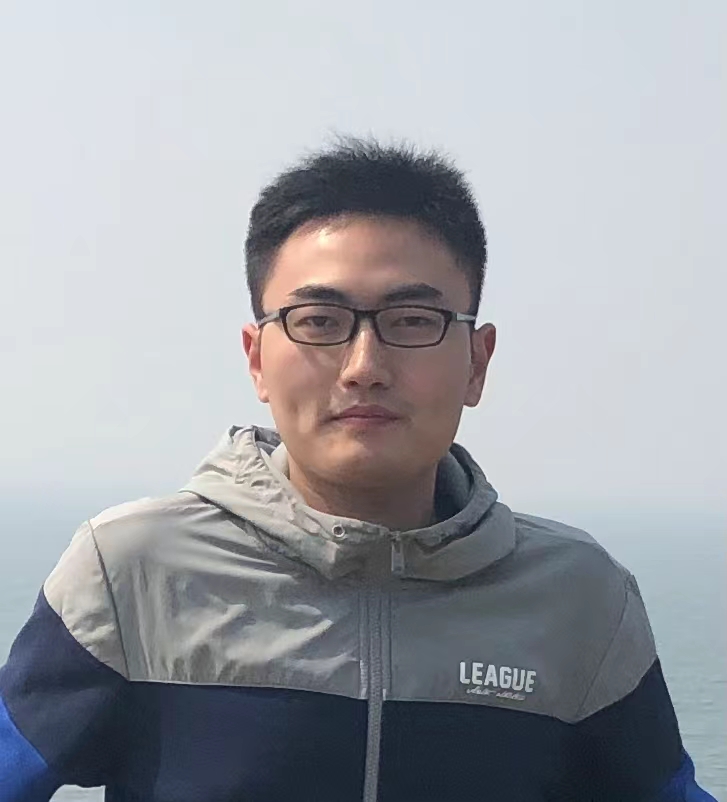}}]{Tao Li}
is a Ph.D. student in the Institute of Statistics and Big Data,
Renmin University of China.
He received his BS degree in mathematics from Nanjing University in 2019.
His research interests include optimal transport problems, generative model, sufficient dimension reduction and variable selection.
\end{IEEEbiography}

\begin{IEEEbiography}
[{\includegraphics[width=1in,height=1in,clip,keepaspectratio]{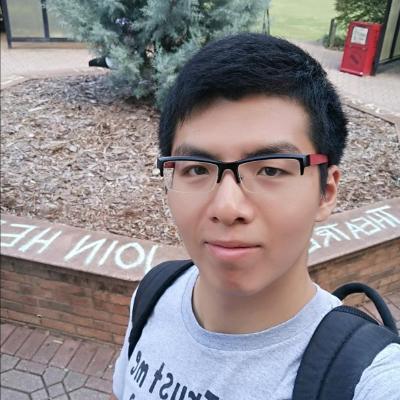}}]{Cheng Meng}
is an assistant professor (Tenure-Track) in the Institute of Statistics and Big Data, Renmin University of China. 
He received his Ph.D. from the Department of Statistics at University of Georgia in 2020. 
His research interests include numerical linear algebra, optimal transport problems, sufficient dimension reduction, nonparametric statistics, and machine learning.
\end{IEEEbiography}

\begin{IEEEbiography}
[{\includegraphics[width=1in,height=1.25in,clip,keepaspectratio]{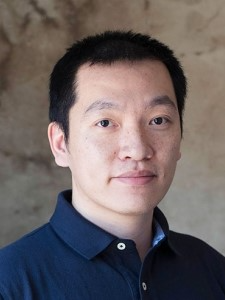}}]{Hongteng Xu}
is an associate professor (Tenure-Track) in the Gaoling School of Artificial Intelligence, Renmin University of China. 
From 2018 to 2020, he was a senior research scientist in Infinia ML Inc. 
In the same time period, he is a visiting faculty member in the Department of Electrical and Computer Engineering, Duke University. 
He received his Ph.D. from the School of Electrical and Computer Engineering at Georgia Institute of Technology (Georgia Tech) in 2017. 
His research interests include machine learning and its applications, especially optimal transport theory, sequential data modeling and analysis, deep learning techniques, and their applications in computer vision and data mining.
\end{IEEEbiography}

\begin{IEEEbiography}
[{\includegraphics[width=1in,height=1.25in,clip,keepaspectratio]{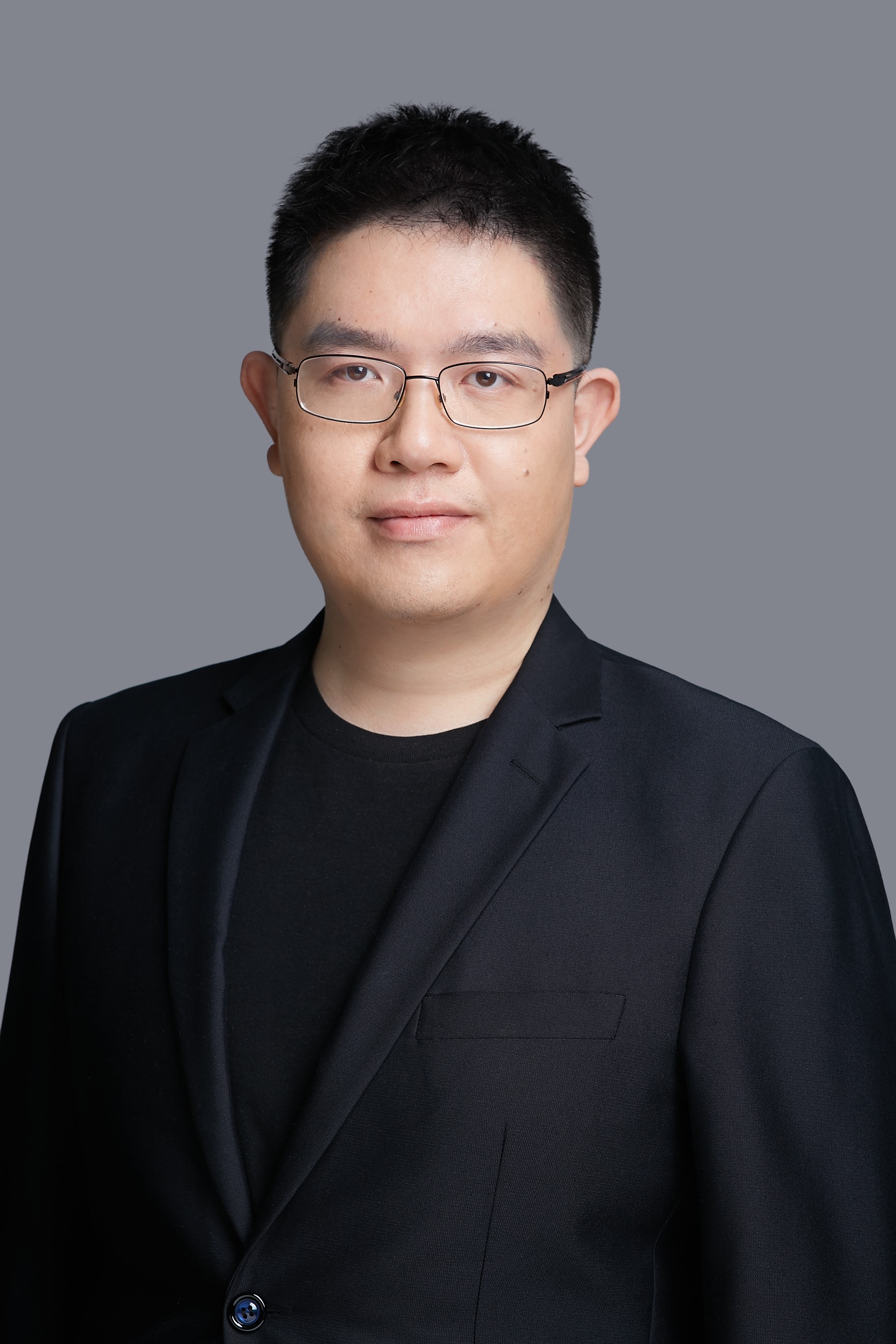}}]{Jun Yu}
is an associate professor in the school of mathematics and statistics, Beijing Key Laboratory on MCAACI, and key laboratory of mathematical theory and computation in information security, Beijing institute of technology.
He received the Ph.D degree in statistics from Peking University, China in 2019.
His research interests include the design of experiments, statistical sketching and sampling methods for large-scale data, and applied statistics in solving scientific and engineering problems.
\end{IEEEbiography}

\end{document}